# Semisupervised Neural Proto-Language Reconstruction

**Liang Lu**[1] **Peirong Xie**[2] **David R. Mortensen**[1]
[1]Carnegie Mellon University [2]University of Southern California
lianglu@cs.cmu.edu louisxie@usc.edu dmortens@cs.cmu.edu

## Abstract

Existing work implementing comparative reconstruction of ancestral languages (proto-languages) has usually required full supervision. However, historical reconstruction models are only of practical value if they can be trained with a limited amount of labeled data. We propose a semisupervised historical reconstruction task in which the model is trained on only a small amount of labeled data (cognate sets with protoforms) and a large amount of unlabeled data (cognate sets without protoforms). We propose a neural architecture for comparative reconstruction (DPD-BiReconstructor) incorporating an essential insight from linguists' comparative method: that reconstructed words should not only be reconstructable from their daughter words, but also deterministically transformable back into their daughter words. We show that this architecture is able to leverage unlabeled cognate sets to outperform strong semisupervised baselines on this novel task[1].

## 1 Introduction

In the 19th century, European philologists made a discovery that would change the direction of the human sciences: they discovered that languages change in systematic ways and that, by leveraging these systematic patterns, it was possible to reproducibly reconstruct ancestors of families of languages (proto-languages) even when no record of those languages survived. This technique, called the comparative method, provided an unprecedented window into the human past—its cultures, its migrations, and its encounters between populations.

The assumption that historical changes in pronunciation ("sound changes") are regular, known as 'the regularity principle' or 'the Neogrammarian hypothesis', is fundamental to the comparative method (Campbell, 2021). As the 19th century Neogrammarians Hermann Osthoff and Karl Brugmann put it:

> "Every sound change, in so far as it proceeds mechanically, is completed in accordance with laws admitting of no exceptions; i.e. the direction in which the change takes place is always the same for all members of a language community, apart from the case of dialect division, and all words in which the sound subject to change occurs in the same conditions are affected by the change without exception." (*Morphologische Untersuchungen auf dem Gebiete der indogermanischen Sprachen i*, Brugmann and Osthoff, 1878, p. xiii, translated and quoted in Szemerényi, 1996)

| | 'grandchild' | 'bone' | 'breast' | 'laugh' |
|---|---|---|---|---|
| Kachai | ðɐ | rɐ | nɐ | ni |
| Huishu | ruk | ruk | nuk | nuk |
| Ukhrul | ru | ru | nu | nu |
| Reference | *du | *ru | *nu | *nɨ |
| Label | Y | Y | N | N |
| D2P | *du | *ru | *nu | *nu |
| DPD | *du | *ru | *nu | *nU |

Table 1: Hypothetical illustration of the shortcomings of daughter-to-proto (D2P) models drawn from Tangkhulic Languages. "U" represents a sound other than "u" predicted by a hypothetical daughter-to-proto-to-daughter (DPD) model. "Y" and "N" indicate whether the reference protoform is labeled.

The comparative method, however, is challenging for humans to apply. This is true largely because it involves dealing with large volumes of data and modeling numerous interactions between competing patterns. One must balance the need for phonetic similarity between reconstructed words and their descendants (reflexes) with the need to be able to deterministically derive reflexes from reconstructed words with a single set of sound changes. It imposes a heavy cognitive load. For this reason, researchers have long aspired to implement the comparative method computationally.

[1]The code is available at https://github.com/cmu-llab/dpd.

With some exceptions (He et al., 2023; Akavarapu and Bhattacharya, 2023), recent attempts at automatic reconstruction models mostly take the form of neural sequence-to-sequence transduction models similar to those used for machine translation, trained with full supervision on a dataset where every cognate set is paired with a gold reconstruction (Meloni et al., 2021; Chang et al., 2022; Fourrier, 2022; Cui et al., 2022; Kim et al., 2023).

The development of supervised reconstruction systems has given the field insights into how reconstruction of proto-languages (unattested ancestor languages) can be modelled computationally. However, in a realistic scenario, these models only become usable once the hardest part of reconstruction has been done (since they rely on the linguist having already identified enough of the sound changes in the data to reconstruct a substantial part of the lexicon). Computational models of comparative reconstruction are most useful if they can be deployed without training data, or if only a small volume of labeled data is needed to prime the comparative pump.

We introduce a semisupervised protoform (reconstructed parent word) reconstruction task wherein the reconstruction model has access at training time to both a small number of cognate sets (sets of daughter words—reflexes—of a single parent) labeled with a protoform and a large number of unlabeled cognate sets, mirroring the situation of historical linguists early in their reconstruction of a proto-language. Though similar to semisupervised machine translation, the semisupervised reconstruction formulation entails the absence of target-side monolingual data. Most semisupervised machine translation techniques rely on monolingual data in the target language, such as back-translation (Edunov et al., 2018; Sennrich et al., 2016) and pre-trained target-side language models (Skorokhodov et al., 2018; Gülçehre et al., 2015). In contrast, in this task, models have access only to cognate sets (no monolingual text), meaning the structure of the problem is quite different.

In this paper, we propose to incorporate the comparative method into a semisupervised reconstruction model via end-to-end multi-task learning. Our proposed model, named DPD-BiReconstructor, learns to improve its reconstructions by performing reflex predictions on an intermediate representation of its predicted reconstructions. Reflex prediction losses are propagated into the reconstruction network, allowing the model to train on cognate sets without protoform labels. A hypothetical example from three Tangkhulic languages is shown in Table 1. In this example, the phonetic information in the 'grandchild' and 'bone' sets is insufficient to reconstruct 'laugh' with a distinct vowel, so daughter-to-proto (D2P) models will typically reconstruct the two words identically. Models incorporating reflex prediction, however, are able—in principle—learn to reconstruct words like 'laugh' as distinct from words like 'breast'. Experiments show that DPD is an attractive approach for semisupervised reconstruction, and a combination of DPD with existing semisupervised strategies performs significantly better than baseline strategies in almost all situations. Additionally, analyses show some indication that DPD-based models could help improve supervised reconstruction.

## 2 Methods

### 2.1 Model

We propose a multi-task reconstruction strategy that learns to recover reflexes from its own reconstructions, effectively utilizing unlabeled cognate sets. Our model comprises a reconstruction sub-network (D2P for daughter-to-protoform) and a reflex-prediction sub-network (P2D for protoform-to-daughter) with shared phoneme embeddings. On labeled data, the model learns sound changes from accurate reconstructions to reflexes, in addition to learning reconstruction. In the absence of labeled protoforms, the reflex prediction sub-network acts as a weak supervision by informing the reconstruction sub-network on whether correct reflexes can be derived from proposed reconstructions. This workflow directly mirrors the comparative method's constraint that reconstruction must yield protoforms such that reflexes can be derived from them through regular sound changes. We refer to our training strategy and its architectural realization as Daughters-to-Protoform-to-Daughters Bidirectional Reconstructor (DPD).

For D2P to learn from P2D, we propagate gradients from P2D into D2P. It is not feasible, however, to simply feed token predictions from D2P into P2D, as token outputs are not differentiable.

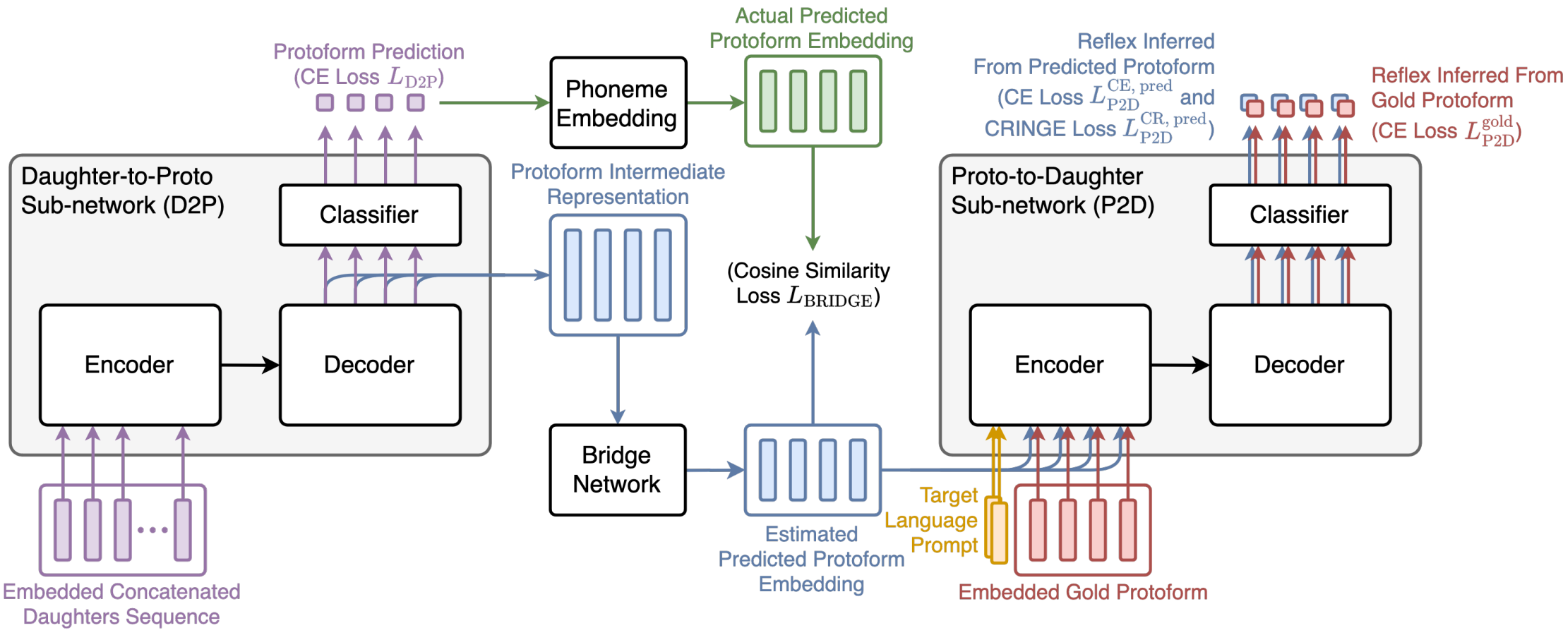

Figure 1: The DPD architecture, with a reconstruction sub-network (D2P), a reflex prediction sub-network (P2D), and a dense bridge network. The bridge network connects the final-layer decoder state from D2P to the encoder input of P2D. For labeled data, supervised cross-entropy (CE) is computed for D2P's protoform prediction $L_{\text{D2P}}$ and P2D's reflex prediction from the gold protoform $L_{\text{P2D}}^{\text{gold}}$. On both labeled and unlabeled data, reflex prediction losses (based on the predicted protoform) consisting of a CE loss $L_{\text{P2D}}^{\text{CE, pred}}$ and a CRINGE loss $L_{\text{P2D}}^{\text{CR, pred}}$ (for incorrectly predicted protoforms) are propagated into both sub-networks. An additional cosine similarity loss $L_{\text{BRIDGE}}$ is used to train the bridge network. The overall loss is calculated as a weighted sum of all the losses, with the weights being tunable hyperparameters. Reflexes in the same cognate are concatenated together (with separaters) into one sequence and embedded with both phoneme and language embedding as in Meloni et al. (2021) and Kim et al. (2023) (and an additional positional embedding for Transformer). Phoneme embedding is shared among D2P and P2D, whereas language embedding is only used in D2P.

Instead, we represent reconstruction outputs as a continuous latent space, inspired by end-to-end spoken language understanding systems in which a semantic understanding sub-network receives a latent representation of text transcriptions from a speech-recognition sub-network (Saxon et al., 2021; Arora et al., 2022). In particular, D2P's final-layer decoder output is connected to P2D's encoder via a trainable dense network, referred to as the bridge network (See Figure 1).

There is no easy way to transform gold protoforms used to train P2D so that they match the bridge network's output latent representation. To encourage a consistent input representation for P2D, we add a cosine similarity loss between the bridge network's output and the actual phoneme embeddings of D2P's predictions, effectively training the bridge network to serve as an alternative embedding for reconstructed protoforms.

As a final training objective, we discourage P2D from producing correct reflexes given incorrect protoforms so that D2P can be better informed when producing incorrect reconstructions. This is achieved via a CRINGE loss on the P2D outputs, designed to penalize negative examples (Adolphs et al., 2022). For negative examples, the CRINGE loss contrasts each negative token with an estimated positive token (by sampling a non-negative token in the model's top-$k$ predictions) and applies a contrastive learning objective (cross-entropy loss) to lower the probability of the negative token. When training the P2D sub-network, a correctly predicted reflex is considered a negative example if the input was an incorrect protoform.

The DPD strategy does not specify the architecture of the D2P and P2D sub-networks. In this paper, we adapt existing encoder-decoder architectures proposed for neural reconstruction, including GRU sub-networks based on Meloni et al. (2021)[2] and Transformer sub-networks based on Kim et al. (2023).

### 2.2 Semisupervised Strategies

Aside from our DPD strategy, a naïve approach is to discard unlabeled data and perform supervised training. We take this, the **supervised-only strategy**, as our first baseline. To compare with established semisupervised machine learning techniques, we implement two more strategies: Bootstrapping and Π-models, representing

[2]We use Chang et al. (2022)'s PyTorch reimplementation obtain from `https://github.com/cmu-llab/middle-chinese-reconstruction`, with minor modifications to support multi-layer.

the **proxy-labelling** and **consistency regularization** approaches respectively.

Bootstrapping adds the model's most confident predictions on unlabeled data to the train set as pseudo-labels (Lee, 2013). In our Bootstrapping setup, the model's most confident (i.e. probable) protoform reconstructions for unlabeled cognate sets, filtered by a minimum confidence threshold and capped at a maximum number of top predictions per epoch, are added as pseudo-labels to the training set at the end of each epoch starting from a set number of warmup epochs (See Appendix C for Bootstrapping hyperparameters).

Π-model optimizes the model's consistency by creating two stochastically augmented inputs from the same training example, feeding both of augmented inputs into the model, and minimizing the mean square difference between the two outputs (Laine and Aila, 2017). For continuous inputs, stochastic augmentation could be simple noise. Stochastic changes to phonemes, however, would defy protoform reconstruction's goal of finding regular sound changes. Instead, we implement stochastic cognate set augmentation by randomly permuting the order of reflexes and randomly dropping daughter languages.

Observe that some of the above strategies can be combined: Bootstrapping can always be used on top of every other strategy, while our proposed DPD architecture can be merged with Π-model into a model that performs both reflex prediction and consistency regularization (See Appendix O for detail). In total, we test 8 strategies: supervised only (SUPV), Bootstrapping (BST), Π-model (ΠM), Π-model with Bootstrapping (ΠM-BST), DPD-BiReconstructor (DPD), DPD with Bootstrapping (DPD-BST), DPD merged with Π-model (DPD-ΠM), and DPD-ΠM with Bootstrapping (DPD-ΠM-BST). Among the 8 strategies, we consider single baseline strategies (SUPV, BST, ΠM) to be weak baselines, and the combination of non-DPD semisupervised techniques (ΠM-BST) to be the strong baseline. Combined with 2 encoder-decoder architectures, GRU and Transformer, we experiment with 16 strategy-architecture combinations which we identify by an architecture prefix (GRU- or Trans-) followed by the strategy name.

### 2.3 Experiments

**Datasets:** We test both Romance and Sinitic languages, represented by the phonetic version[3] of Meloni et al. (2021)'s Romance dataset (Rom-phon) for Latin reconstruction and Chang et al. (2022)'s WikiHan dataset for Middle Chinese reconstruction[4]. We simulate the semisupervised reconstruction scenario by hiding a random subset of protoform labels from the fully labeled train set. We refer to the percentage of labels retained after label removal as the **labeling setting**.

**Cross-strategy comparisons:** The primary interest of our experiments is how reconstruction performance differs between strategies with a fixed percentage of labels. We fix the labeled percentage at 10%, which entails approximately 516 and 870 labeled cognate sets for WikiHan and Rom-phon respectively. Due to randomness in semisupervised dataset generation, we repeat the experiment on four randomly generated semisupervised trained sets for each of WikiHan and Rom-phon, labeled group 1 to group 4. Each of the 16 strategy-architecture combinations is tested 10 times in each group. We then compare the performance of strategies within the same architecture.

**Cross-labeling setting comparisons:** We test all strategies on datasets with 5%, 10%, 20%, and 30% of labels to study the relationship between the labeling setting and the model's performance[5]. With these labeling settings, the numbers of labeled cognate sets ranging from 181 to 1,084 for WikiHan and 304 to 1,821 for Rom-phon[6]. Randomly selecting labels for each of the labeling settings—especially on already small datasets—introduces variations on the learnable information contained in the labels and could introduce noise. To mitigate this, we enforce a monotonic subset selection constraint: given the complete train label set $L$ and a semisupervised label set $L_{p_i} \subseteq L$ retaining $p_i$ per cent of the labels, semisupervised sets $L_{p_1}, L_{p_2}$ of increasing percentages ($p_1 \leq p_2$) must satisfy $L_{p_1} \subseteq L_{p_2}$ (See Figure 2). The 10% labeled dataset we use in cross-labeling setting comparison corresponds to group 1 in cross-strategy comparison (See Appendix B for detail).

[3]We focus on the phonetic version due to a large number of semisupervised strategies and since DPD is primarily motivated by phonetic reconstruction.

[4]Rom-phon is not licenced for redistribution; WikiHan is licenced under Creative Commons CC0. For more dataset details, see Appendix A.

[5]It is often the case that, in manual reconstruction projects, the majority of sound changes can be discovered from a minority of the available cognate sets. We hope to capture this observation with our choice of labeling settings.

[6]See Table 6 for detail.

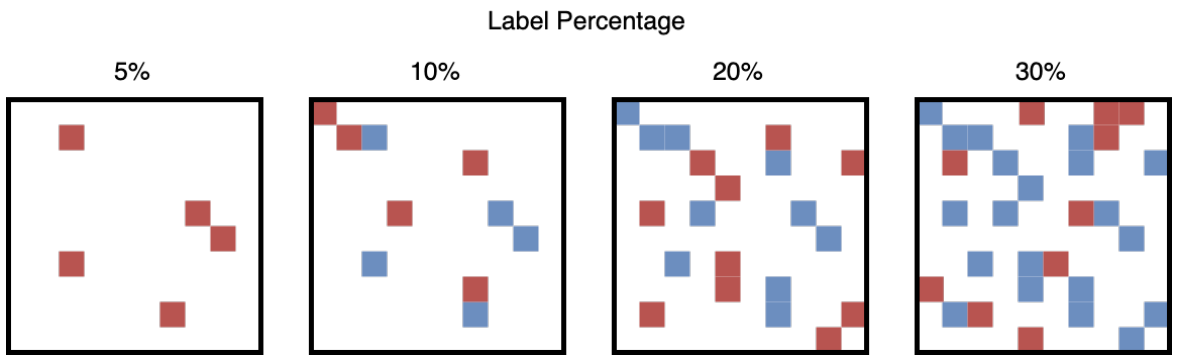

Figure 2: An illustration of the monotonic constraint for creating semisupervised datasets of varied labeling settings. In a hypothetical dataset with 100 cognate sets, represented as a $10 \times 10$ grid, shaded cells indicate cognate sets with associated labels (i.e. gold protoform). "■" indicates a protoform label not present in a previous subset. Observe that as the percentage of labels increases, no label is removed.

**Hyperparameters:** We use Bayesian search to tune hyperparameters for all 16 strategy-architectures combinations on a fixed 10% labeled semisupervised dataset[7]. For each strategy-architecture combination, for 100 iterations, we select the hyperparameter leading to the best validation phoneme edit distance. See Appendix C for details on hyperparameters.

**Evaluation metrics:** Evaluation metrics we use follow directly from supervised reconstruction literature, including token edit distance (TED), a count of the number of insertion, deletion, or substitution operations between the prediction and the target (Levenshtein, 1966); token error rate (TER), a length-normalized token edit distance (Cui et al., 2022); accuracy (ACC); feature error rate (FER), a measure of phonetic distance by PanPhon (Mortensen et al., 2016); and B-Cubed F Score (BCFS), a measure of structural similarity between the prediction and the target (Amigó et al., 2009; List, 2019).

**Statistical tests:** For each combination of dataset group, labeling setting, and architecture, we test all strategies against each other for differences. We use both the Wilcoxon Rank-Sum test (Wilcoxon, 1992) and Bootstrap test for mean difference (Sivaganesan, 1994) to test for significance in performance difference. We use a $\alpha = 0.01$ significance threshold and consider results to be significant if indicated by both tests.

[7]Due to a large number of strategy-architecture combinations, tuning the model on every labeling settings is costly. We fix the dataset generation seed to 0 for tuning.

## 3 Results

**Cross-strategy comparisons:** Table 2 shows the performance of all strategy-architecture combinations on four 10% labeled datasets. For both architectures on WikiHan, DPD-ΠM-BST performs the best and significantly better than all baselines on all metrics. Transformer trained with DPD attains similar performance to DPD-ΠM-BST, outperforming all baseline strategies. GRU trained with DPD performs similarly to ΠM-BST, both of which perform better than weak baselines in a majority of situations. On Rom-phon, Transformer performed the best when trained with DPD-ΠM-BST while GRU performed the best when trained with DPD-BST, both of which are significantly better than all baselines across all metrics.

Interestingly, while Kim et al. (2023) finds that a supervised Transformer model outperforms Meloni et al. (2021)'s GRU model on Rom-phon but not WikiHan, we observe the opposite in a 10% labeled semisupervised reconstruction setup, with Transformer outperforming GRU on WikiHan but not Rom-phon under most strategies. This appears to contradict Kim et al. (2023)'s hypothesis that a Transformer reconstruction model requires more data compared to an RNN.

Consistent with our hypothesis that access to different subsets of labels affects learning outcomes, we observe high variations in performance between dataset groups. Figures 6 and 7 visualize the performance of strategy-architecture combination by group, revealing that most strategies tend to do better on some dataset seeds.

**Performance for varied labeling settings:** Despite only being tuned with 10% of labels, DPD-based strategies generalized to other labeling settings, often outperforming strong baseline strategies on all metrics on at least one of the architectures (See Appendix F for detail). We see a non-linear scaling between performance and the percentage of labels and notice higher performance variations between strategies for lower percentages of labels. At a 5% labeling setting, for example, GRU-DPD-ΠM-BST almost doubles the accuracy of GRU-SUPV on WikiHan. At a 30% labeling setting, some strategies attain accuracy close to existing fully supervised reconstruction models, with Trans-DPD-ΠM 5.14 percentage points behind Meloni et al. (2021)'s supervised GRU on WikiHan (Chang et al., 2022) and GRU-DPD-BST 7.30 percentage points behind Kim et al. (2023)'s

| Architecture | Strategy | ACC% ↑ | TED ↓ | TER ↓ | FER ↓ | BCFS ↑ |
|---|---|---|---|---|---|---|
| Transformer | DPD-ΠM-BST (ours) | **40.50%** ❶❷❸❹ | **1.0075** ❶❷❸❹ | **0.2360** ❶❷❸❹ | **0.0970** ❶❷❸❹ | **0.6707** ❶❷❸❹ |
| | DPD-BST (ours) | 39.06% ❶❷❸❹ | 1.0367 ❶❷❸❹ | 0.2428 ❶❷❸❹ | 0.0997 ❶❷❸❹ | 0.6630 ❶❷❸❹ |
| | DPD-ΠM (ours) | 37.72% ❶❷❸❹ | 1.0791 ①❷❸ | 0.2528 ①❷❸ | 0.1022 ①❷❸❹ | 0.6472 ❷❸ |
| | DPD (ours) | 39.50% ❶❷❸❹ | 1.0356 ❶❷❸❹ | 0.2426 ❶❷❸❹ | 0.0993 ❶❷❸❹ | 0.6564 ❶❷❸❹ |
| | ΠM-BST | 34.21% | 1.1489 | 0.2691 | 0.1106 | 0.6371 |
| | BST (Lee, 2013) | 34.78% | 1.1455 | 0.2683 | 0.1109 | 0.6334 |
| | ΠM (Laine and Aila, 2017) | 34.30% | 1.1699 | 0.2740 | 0.1122 | 0.6209 |
| | SUPV | 33.25% | 1.1891 | 0.2785 | 0.1140 | 0.6138 |
| GRU | DPD-ΠM-BST (ours) | **39.74%** ❶❷❸❹ | **1.0280** ❶❷❸❹ | **0.2408** ❶❷❸❹ | **0.0972** ❶❷❸❹ | **0.6683** ❶❷❸❹ |
| | DPD-BST (ours) | 35.89% ①❷③ | 1.1025 ①❷❸④ | 0.2582 ①❷❸④ | 0.1039 ①❷❸④ | 0.6493 ①❷❸④ |
| | DPD-ΠM (ours) | 37.90% ①❷❸❹ | 1.0697 ①❷❸❹ | 0.2506 ①❷❸❹ | 0.1006 ①❷❸❹ | 0.6517 ①❷❸❹ |
| | DPD (ours) | 34.51% ①③④ | 1.1538 ①③④ | 0.2703 ①③④ | 0.1091 ③④ | 0.6278 ①③④ |
| | ΠM-BST | 34.99% ①②③ | 1.1479 ①③ | 0.2689 ①③ | 0.1077 ①③ | 0.6354 ①②③ |
| | BST (Lee, 2013) | 28.18% | 1.3092 | 0.3067 | 0.1208 | 0.5939 |
| | ΠM (Laine and Aila, 2017) | 32.59% | 1.2047 | 0.2822 | 0.1137 | 0.6166 |
| | SUPV | 28.16% | 1.3257 | 0.3105 | 0.1234 | 0.5835 |

| Architecture | Strategy | ACC% ↑ | TED ↓ | TER ↓ | FER ↓ | BCFS ↑ |
|---|---|---|---|---|---|---|
| Transformer | DPD-ΠM-BST (ours) | **34.63%** ❶❷❸❹ | **1.3115** ❶❷❸❹ | **0.1463** ❶❷❸❹ | **0.0588** ❶❷❸❹ | **0.7850** ❶❷❸❹ |
| | DPD-BST (ours) | 33.51% ❶❷③❹ | 1.3605 ❶❷③❹ | 0.1517 ❶❷③❹ | 0.0599 ❶❷❸❹ | 0.7763 ❶❷③❹ |
| | DPD-ΠM (ours) | 29.24% | 1.5888 | 0.1772 | 0.0732 | 0.7423 |
| | DPD (ours) | 31.94% ①②③④ | 1.5111 | 0.1685 | 0.0678 ②③ | 0.7529 |
| | ΠM-BST | 32.10% ①②③④ | 1.4005 ①②③④ | 0.1562 ①②③④ | 0.0636 ①②③④ | 0.7716 ①②③④ |
| | BST (Lee, 2013) | 29.95% | 1.5066 | 0.1680 | 0.0704 | 0.7555 |
| | ΠM (Laine and Aila, 2017) | 26.97% | 1.7134 | 0.1911 | 0.0796 | 0.7239 |
| | SUPV | 26.99% | 1.7331 | 0.1933 | 0.0794 | 0.7218 |
| GRU | DPD-ΠM-BST (ours) | 36.78% ❶❷ | 1.2380 ❶❷❸❹ | 0.1381 ❶❷❸❹ | 0.0483 ❶❸❹ | 0.7980 ❶❷③❹ |
| | DPD-BST (ours) | **37.60%** ❶❷❸❹ | **1.2149** ❶❷❸❹ | **0.1355** ❶❷❸❹ | **0.0457** ❶❷❸❹ | **0.8014** ❶❷❸❹ |
| | DPD-ΠM (ours) | 31.51% | 1.4892 | 0.1661 | 0.0628 | 0.7586 |
| | DPD (ours) | 31.12% | 1.4837 | 0.1655 | 0.0608 | 0.7591 |
| | ΠM-BST | 35.50% | 1.2970 | 0.1447 | 0.0531 | 0.7909 ① |
| | BST (Lee, 2013) | 35.87% | 1.2893 | 0.1438 | 0.0509 | 0.7908 |
| | ΠM (Laine and Aila, 2017) | 29.40% | 1.5440 | 0.1722 | 0.0643 | 0.7517 |
| | SUPV | 30.69% | 1.5018 | 0.1675 | 0.0612 | 0.7558 |

Table 2: Performance of all strategies on 10% labeled WikiHan (top) and Rom-phon (bottom) for each architecture, averaged across all runs in four groups (10 runs per strategy-architecture combination per group). Bold: best-performing strategy for the corresponding architecture and dataset; ①: significantly better than all weak baselines (SUPV, BST, and ΠM) on group 1 with $p < 0.01$; ❶: significantly better than the ΠM-BST strong baseline and all weak baselines on group 1 with $p < 0.01$; ②, ③, ④, ❷, ❸, ❹: likewise for groups 2, 3, and 4.

supervised Transformer on Rom-phon (See Table 15 and Appendix J for detail).

## 4 Analysis

### 4.1 DPD Training

We investigate the interaction between D2P and P2D during training. Figure 3 shows the train and validation accuracy trajectories of reflexes reconstructed from the intermediate representation of protoform reconstructions. We perform a CRINGE loss ablation by repeating the same run but with CRINGE loss disabled (same hyperparameter and model parameter initialization). In both cases, reflexes are more accurately predicted from correct protoform reconstructions, corroborating our motivation that better reflex reconstruction should promote better protoform reconstruction. CRINGE loss appears to lead to slightly lower reflex accuracy given incorrect reconstruction at earlier epochs, albeit at the expense of slightly lower reflex accuracy given correct reconstruction. In practice, we find the CRINGE loss weight to be a relatively insignificant hyperparameter.

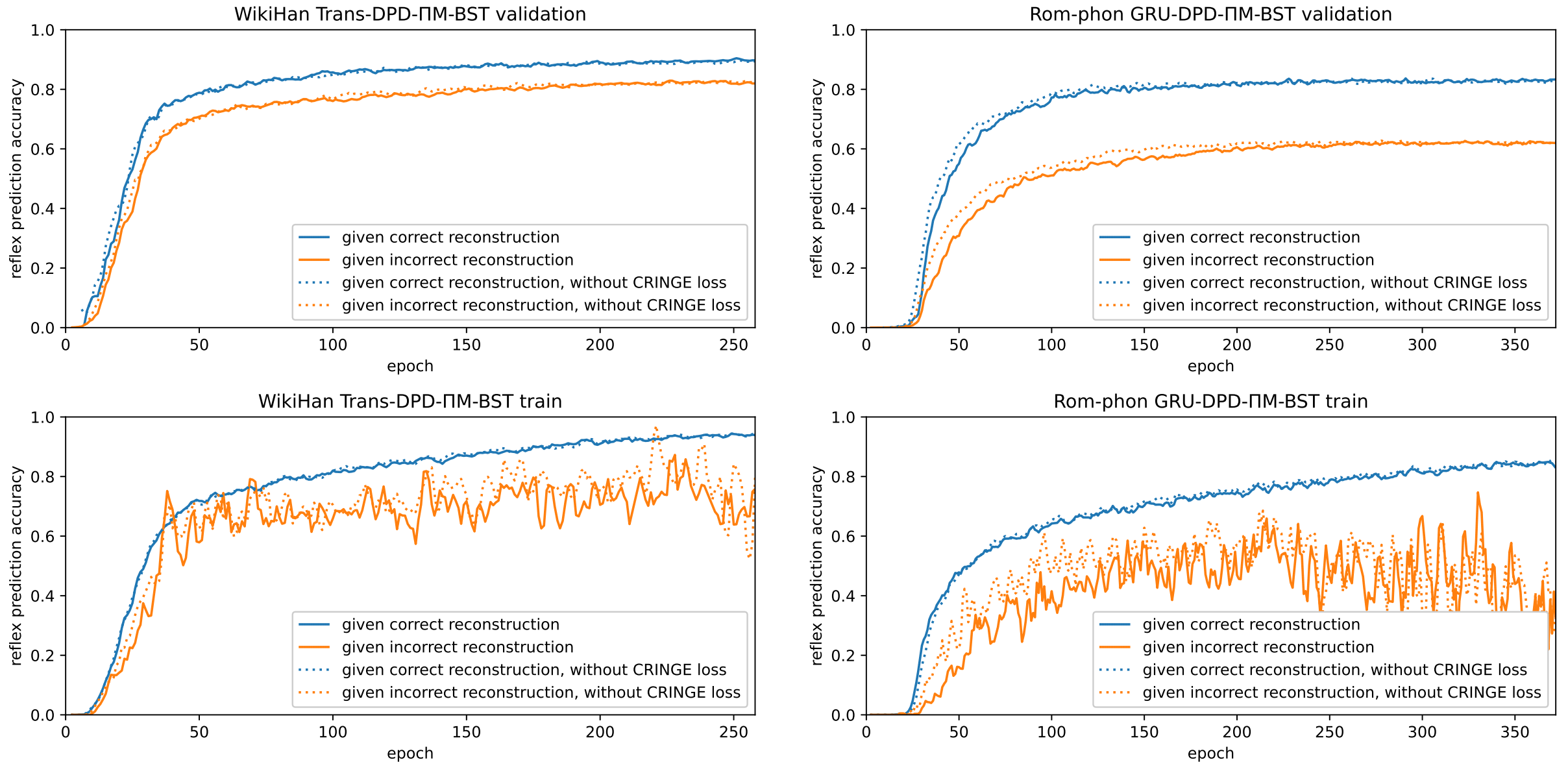

(a) 10% labeled WikiHan group 1

(b) 10% labeled Rom-phon group 1

Figure 3: Validation (top) and train (bottom) reflex reconstruction accuracy given correct versus incorrect protoform prediction during training for a randomly selected run within the most accurate strategy. A rolling average of window size 3 is used for smoothing.

### 4.2 Reflex Prediction Performance

While the DPD architecture is designed for reflex prediction-assisted reconstruction, we observe good reflex performance in some situations. Table 3 shows the strategy-architecture combinations with the most accurate P2D sub-network when evaluated on gold protoforms[8]. Compared to reconstruction, we obtain semisupervised reflex prediction performance that are much closer to supervised performance, even with small percentages of labels—consistent with the assumption that sound changes in the proto-to-daughters direction are easier to model.

It is worth noting, however, that reflex prediction performance based on gold protoforms depends on the weight of the corresponding loss. In fact, some of the best DPD models perform poorly when the P2D sub-network is evaluated on gold protoforms. We conclude that P2D's ability to assist D2P during training is not contingent on P2D's performance on gold protoforms as discrete input.

[8]It is arguably more interesting to evaluate reflex prediction performance based on the model's latent protoform representation. Unfortunately, obtaining latent representations of correct protoforms is not always possible.

### 4.3 Learned Phonetic Representations

Inspired by Meloni et al. (2021), we probe the model's learned embeddings for a hierarchical organization of phonemes using sklearn's Ward variance minimization algorithm (Ward, 1963). Figure 4 shows the results for two selected daughter languages on the most accurate model in each dataset (GRU-DPD-BST for Rom-phon and Trans-DPD-ΠM-BST for WikiHan) and the best model from their non-DPD counterpart (GRU-BST and Trans-ΠM-BST respectively).

For French, phoneme embeddings trained with DPD-BST reveals a clear division between vowels and consonants similar to Meloni et al. (2021)'s supervised reconstruction model. Except for [ø], nasal vowels are grouped together. Specific

| | | 5% | 10% | 20% | 30% | 100% |
|---|---|---|---|---|---|---|
| WikiHan | Top performer | GRU-DPD-ΠM-BST | GRU-DPD-ΠM-BST | Trans-DPD-ΠM-BST | GRU-DPD-BST | - |
| | ACC | 45.37% | 56.74% | 61.83% | 62.02% | 66.43% |
| Rom-phon | Top performer | Trans-DPD-ΠM-BST | Trans-DPD-ΠM-BST | Trans-DPD-BST | Trans-DPD-BST | - |
| | ACC | 53.51% | 57.95% | 60.08% | 61.13% | 63.85% |

Table 3: Strategy-architecture combinations with the highest reflex prediction accuracy in group 1 for each labeling setting, along with reference supervised (100% labeled) reflex prediction accuracy.

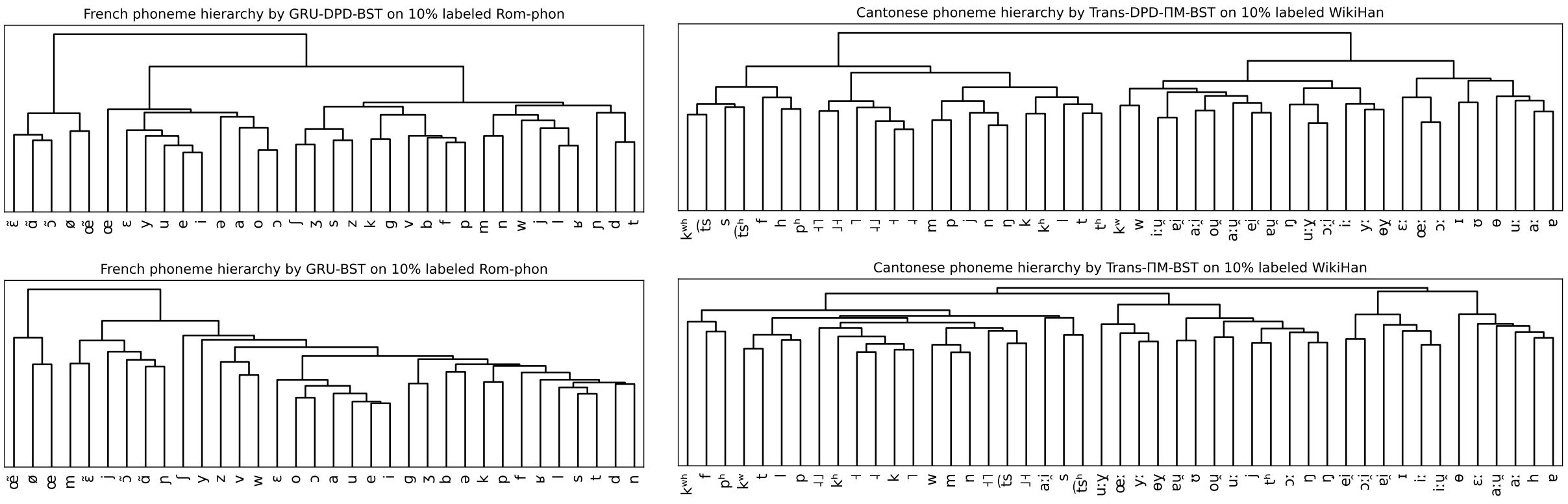

(a) French (from Rom-phon)

(b) Cantonese (from WikiHan)

Figure 4: Hierarchical clustering revealing learned phoneme organization for two selected daughter languages obtained from the best run (within group 1 of 10% labeling setting) in the best DPD-based strategy-architecture combination (top) and the best run from their non-DPD counterpart (bottom).

phoneme pairs with minimal difference in features are also placed together, such as the alveolar fricatives [s] and [z], post-alveolar fricatives [ʃ] and [ʒ], and velar plosives [k] and [g], and all of which only differ in voicing. In the embeddings trained with BST, some but not all vowels are clustered together, and sister groups (i.e. immediate relative in the tree) are less interpretable, such as [m] and [ɛ̃] as well as [b] and [ə]. For Cantonese, we see a similar pattern where vowels and consonants have a clearer division when trained with DPD-ΠM-BST. Additionally, tones are organized into the same cluster by DPD-ΠM-BST but not by ΠM-BST. In Appendix N, we extend our probe to set of all phonemes present in daughter languages[9] and find that the above observations generalize to phoneme organization beyond a single language, which Meloni et al. (2021) does not consider as part of their analysis.

We conclude from phonetic probing that DPD-based strategies are better at capturing linguistically meaningful representations of phonemes. It is possible that DPD-based strategies need good phonetic representations to perform well on multiple phonology-intensive tasks, which could in turn better inform protoform reconstruction.

### 4.4 Ablation on Unlabeled Data

To study whether the performance gains of semisupervised strategies are because of their effective use of unlabeled cognate sets, we perform ablation experiments at a 10% labeling setting but with all unlabeled training data removed[10], effectively creating a small supervised training set[11].

We find that, in the absence of unlabeled data, ΠM, DPD, and DPD-ΠM can sometimes perform significantly better than SUPV, but almost always perform significantly worse than when unlabeled data is used or when unlabeled data is used in conjunction with Bootstrapping (see Tables 13 and 14). This seems to suggest that ΠM, DPD, and DPD-ΠM learn effectively from both labeled and unlabeled data. It is possible that, on labeled data, the P2D sub-network in DPD can still inform the D2P sub-subnetwork, and the stochastic data augmentation in our implementation of Π-model can augment labeled training examples.

### 4.5 Applicability in Supervised Reconstruction

Seeing some indication that the semisupervised reconstruction strategies are applicable for supervised reconstruction on a small subset of the training set (see Section 4.4), we test whether their advantages generalize to supervised reconstruction on the full training set[12].

Table 15 compares the supervised reconstruction performance of ΠM, DPD, and DPD-ΠM with existing supervised reconstruction methods, including Meloni et al. (2021)'s GRU model, Kim et al. (2023)'s Transformer model, and Lu et al. (2024)'s state-of-the-art reranked reconstruction systems. We find that, with the right architecture,

[9]Phonemes present only in the proto-language are not included because non-DPD do not update their embeddings.

[10]We exclude strategies with Bootstrapping because it has no effect when there is no unlabeled data.

[11]See Appendix I for experimental details.

[12]See Appendix J for experimental details.

ΠM, DPD, and DPD-ΠM can often outperform SUPV. On average, Trans-DPD-ΠM performs the best on WikiHan for all metrics, and GRU-DPD performs the best on Rom-phon for all metrics except FER. On WikiHan, Trans-DPD-ΠM performs significantly better than Lu et al. (2024) on FER only. On Rom-phon, GRU-DPD and GRU-DPD-ΠM both perform significantly better than Lu et al. (2024) on ACC only. We conclude that, despite being motivated by semisupervised reconstruction, ΠM and DPD could be useful for supervised reconstruction. We leave it to future work to understand the role of data augmentation and the DPD architecture in supervised settings.

## 5 Related Work

**Computational Historical Linguistics:** Protoform reconstruction and reflex prediction are two central tasks in computational historical linguistics. Protoform reconstruction predicts the protoform given reflexes in a cognate set, while reflex prediction models the changes from the protoform to its reflexes. Computational reconstruction and reflex prediction methods vary and include rule-based systems (Heeringa and Joseph, 2007; Marr and Mortensen, 2020, 2023), probabilistic models operating on phylogenetic trees (Bouchard-Côté et al., 2007b,a, 2009, 2013), automated alignment systems (Bodt and List, 2022; List et al., 2022), and more recently, neural networks (Fourrier, 2022).

**Supervised Reconstruction and reflex prediction:** Ciobanu and Dinu (2018) and Ciobanu et al. (2020) formulate protoform reconstruction as a sequence labelling task and use conditional random field to reconstruct protoform phonemes at each position in the daughter sequences. Meloni et al. (2021) reformulates the reconstruction task as a sequence-to-sequence task by concatenating reflexes into a single input sequence separated by language tags and uses GRU to reconstruct Latin on a new Romance dataset. A group of subsequent researchers refined this task with additional datasets (Chang et al., 2022) and improved neural methods (Kim et al., 2023; Akavarapu and Bhattacharya, 2023; Lu et al., 2024).

Reflex prediction can be viewed as sequence-to-sequence transduction in the reverse direction. Cathcart and Rama (2020) propose an LSTM encoder-decoder model to infer Indo-Aryan reflexes from Old Indo-Aryan, aided by semantic embedding. Arora et al. (2023) replicates Cathcart and Rama (2020)'s experiments on South Asia languages with both GRU and Transformer models. As reflex prediction maps from proto-language to multiple daughter languages, a prompting token is often attached to the input to specify the target daughter language.

**Non-Supervised Computational Historical Linguistics:** Past work in which comparative reconstruction was performed without full supervision included Bouchard-Côté et al. (2009) and He et al. (2023). To the best of our knowledge, no research focuses specifically on semisupervised neural reconstruction.

**Semisupervised Learning:** Effective semisupervised learning should utilize unlabeled training data. One approach is proxy-labelling, by which synthesized labels are added to unlabeled training examples via heuristics (Ouali et al., 2020). Another approach is consistency regularisation, largely based on the smoothness assumption that is applicable regardless of whether labels are present: similar input data in high-density regions should have similar labels, whereas input data separated by low-density regions should not (Tsai and Lin, 2019; Ouali et al., 2020; Luo et al., 2018).

Our task is related to semisupervised machine translation (Edunov et al., 2018; Sennrich et al., 2016; Skorokhodov et al., 2018; Gülçehre et al., 2015; Cheng et al., 2016). However, it differs crucially in that the model has no access to monolingual text, only labeled and unlabeled cognate sets.

## 6 Conclusion

We introduce the task of semisupervised reconstruction, marking a step forward toward a practical computational reconstruction system that can assist early-stage proto-language reconstruction projects. We design the DPD-BiReconstructor architecture to implement historical linguists' comparative method, yielding performance that surpasses existing sequence-to-sequence reconstruction models and established semisupervised learning techniques, especially when protoform labels are scarce.

## Limitations

Due to a large number of possible strategies, we have limited our focus of semisupervised reconstruction experiments with DPD to 10% labeled

WikiHan and Rom-phon datasets. It is left to future work to expand the research on semisupervised reconstruction to other datasets.

Though DPD has a clear motivation and demonstrates superior empirical performance, interpretations of what sound changes DPD learns from the unlabeled cognate sets that enable its better reconstruction performance are less clear. Our theory is that neural networks within the system are sufficiently expressive to learn better reconstruction in a bidirectional manner, but we have not yet obtained evidence that the model's reasoning matches that of a linguist beyond having a better representation of phonemes and taking the step to derive the reflexes from the reconstruction. Nevertheless, we demonstrate that inferring reflexes from reconstructions is not just a powerful methodology for historical linguistics, but also for computational historical linguistics.

Although we observe less accurate reflexes being predicted from incorrect protoforms compared to correct protoforms, reflexes derived from incorrect protoforms are still highly accurate. Future work could explore ways to mitigate this issue and improve the reflex prediction sub-network's ability to discriminate between correct and incorrect protoforms.

It is also notable that our implementation of the Π-Model included only reflex permutation and daughter deletion as noising strategies. It is possible that other possible strategies may have strengthened this baseline.

The protoform reconstruction task is far from solved—with humans' success at the reconstruction of ancient languages, a truly intelligent reconstruction system should in the future be able to perform reconstruction without the help of labels.

## Ethics

Historical reconstruction involves very limited risks to humans. The risks that do exist are both individual and political. On the one hand, we cannot guarantee that all of the data used in this study were collected in an ethical fashion. However, we did not do any data collection and relied upon existing resources which, to the best of our knowledge, were collected under standard scholarly and scientific protocols. On the other hand, the results of historical reconstruction can be politically fraught. For example, historical reconstructions can be used to show, in some cases, that linguistic boundaries between people groups do not align with cultural and political loyalties. This can be disruptive and may even be associated with violence. Because our work does not concentrate directly on phylogeny—the main source of political complications in comparative reconstruction—we believe that the risks are minimal.

## Acknowledgements

This work is supported by Carnegie Mellon University's SURF grant. We would like to thank Chenxuan Cui, Kalvin Chang, and Graham Neubig for helpful ideas and discussions. We are grateful to our anonymous reviewers for many comments that helped improve the writing and informed additional experiments and analyses.

## A Dataset Details

Both WikiHan and Rom-phon are split by 70%, 10%, and 20% into train, validation, and test sets. We remove labels from the supervised train set to create semisupervised train sets. The validation and test sets are left unmodified. The splits for WikiHan (Chang et al., 2022) match the original work. The splits and preprocessing for Meloni et al. (2021) and matches Kim et al. (2023). Table 4 shows the number of cognate sets in each split. WikiHan includes 8 daughter languages: Cantonese, Gan, Hakka, Jin, Mandarin, Hokkien, Wu, and Xiang. Rom-phon includes 5 daughter languages: French, Italian, Spanish, Romanian, and Portuguese.

## B Dataset Groups

Semisupervised reconstruction datasets are generated pseudo-randomly based on a seed, which is itself chosen randomly. That is, a dataset seed deterministically generates a semisupervised dataset. To generate a semisupervised dataset, we initialize PyTorch's pseudo-random number generator with the dataset seed, assign a uniformly distributed number between 0 and 1 to each training example with `torch.rand` in the same order as they appear in the dataset, and keep the protoform label on cognate sets whose assigned number is above a threshold such that the desired percentage of labels remain. The dataset seed is independent of seeds used to initialize model parameters in the experiments. Table 5 details the dataset seed used to select the subset of training labels to include for each experiment setup, and Table 6 details the number of labels at each labeling setting. We use group 1 for comparisons between labeling settings, with the same dataset seed ensuring label sets are monotonic subsets with respect to labeling settings with increasing percentages of labels. For the 10% labeled setting, four distinct dataset seeds simulate variations in the training data.

| | WikiHan | Rom-phon |
|---|---|---|
| Train | 3,615 | 6,071 |
| Validation | 517 | 878 |
| Test | 1,033 | 1,754 |
| Total | 5,165 | 8,703 |

Table 4: Number of cognate sets in the train, validation, and test split of both datasets.

# C Hyperparameters

We tune hyperparameters using Bayesian search on WandB (Biewald, 2020) with 100 runs for each strategy-architecture combination. We validate the model every 3 epochs and use early stopping if no improvement is made after 24 epochs. The dataset used for tuning semisupervised models is a 10% labeled train set generated using 0 as the dataset seed, making it different from the semisupervised datasets used in the experiments. Hyperparameters for semisupervised models can be found in Tables 16 and 17. Additional hyperparameter tunings for analysis purposes (Sections I and J) follow the same procedure but remove unlabeled training data or use a different labeling setting.

# D Training

Parameter counts for the models can be found in Table 22. Hyperparameter tuning and experiments are performed on a mix of NVIDIA GeForce GTX 1080 Ti, NVIDIA GeForce RTX 2080 Ti, NVIDIA RTX 6000 Ada Generation, NVIDIA RTX A6000, Quadro RTX 8000, and Tesla V100-SXM2-32GB GPUs for a total of 411 GPU days.

# E Package Usage

Our model is implemented in PyTorch (See the code for details at `https://github.com/cmu-llab/dpd`). Sequence alignment is done using lingpy (List and Forkel, 2021) with default parameters. Hierarchical clustering is done using `AgglomerativeClustering` from sklearn with Ward linkage and distance threshold set to 0. Bootstrap tests are done using scipy with random state set to 0, a 99% confidence interval, and 9,999 resamples (default). Wilcoxon Rank-Sum tests are done using scipy at default parameter. Hierarchical clustering is visualized using scipy. Plots are created using Matplotlib.

# F Additional Performance Data

Tables 8 and 9 show the performance of each strategy on group 1 for the 5%, 20%, and 30% labeling settings, along with indicators of statistical significance.

# G Transductive Evaluation

Evaluation for semisupervised learning can be categorized as transductive and inductive (Ouali et al., 2020). Inductive evaluation tests the model's performance on unseen data in a test set, whereas transductive evaluation tests the model's ability to predict labels for the unlabeled data in the train set. In the context of semisupervised protoform reconstruction, transductive evaluation corresponds to the early stage of a real-world reconstruction project where cognate sets in higher abundance than known protoforms. In this section, we report the transductive performance of all the strategies.

We find that transductive performance differences between strategies are largely similar to industive performance on the test set. Significance as to whether our DPD strategies perform better than baseline is also similar to that of inductive evaluation on the test set. For detailed transductive performance and their statistical significance, see Tables 10 (10% labeled WikiHan and Rom-phon), 11 (5%, 20%, and 30% labeled WikiHan), and 12 (5%, 20%, and 30% labeled Rom-phon).

| | | 5% | 10% | 20% | 30% |
|---|---|---|---|---|---|
| WikiHan | Group 1 | 2706283079 | 2706283079 | 2706283079 | 2706283079 |
| | Group 2 | | 2188396888 | | |
| | Group 3 | | 2718489156 | | |
| | Group 4 | | 1416758132 | | |
| Rom-phon | Group 1 | 1893608599 | 1893608599 | 1893608599 | 1893608599 |
| | Group 2 | | 1517201602 | | |
| | Group 3 | | 2341117665 | | |
| | Group 4 | | 3045950670 | | |

Table 5: Dataset seeds used to create semisupervised train sets for each group and labeling setting. Using the same seed on Group 1 guarantees the monotonic increasing subset selection constraint for comparison between different labeling settings.

| | WikiHan | Rom-phon |
|---|---|---|
| 5% | 181 | 304 |
| 10% | 362 | 607 |
| 20% | 723 | 1,214 |
| 30% | 1,084 | 1,821 |
| 100% | 3,615 | 6,071 |

Table 6: Number of labeled training examples (i.e. cognate sets with an associated gold protoform) in the train set for each labeling setting and dataset, as well as the total number of cognate sets for reference (100%).

We observed no clear pattern as to which strategies perform better on transductive evaluation compared to inductive evaluation. On 10% labeled WikiHan, average transductive accuracies are all 0.0-1.14% higher than their test accuracies. On 10% labeled Rom-phon, transductive accuracies for all strategies are 1.20-1.90% worse than their test accuracies. Given that differences between test and transductive performance are relatively consistent across strategies, including for supervised strategies in which the unlabeled portion of the train set effectively acts as another test set, our hypothesis is that the fixed train-test split played a role in the evaluation.

## H Aligned Error Analysis

As an additional analysis, we align the protoform predictions and their targets on the test set using lingpy (List and Forkel, 2021) and identify errors made by reconstruction models. Consistent with Meloni et al. (2021)'s (supervised reconstruction) error analysis on Rom-phon, tense-lax errors occur most frequently for both the best DPD strategy and its non-DPD counterpart, with [i]/[ɪ], [e]/[ɛ], [o]/[ɔ], and [u]/[ʊ] being the top-four exchange errors[13]. The top errors in WikiHan include inserting or deleting [j] and [w], vowel height exchange errors between pairs such as [e]/[i], [o]/[u], and [æ]/[a], tone errors, along with [a]/[o] errors. Exchange errors make up 68% and 63% of all errors for WikiHan and Rom-phon respectively.

We compare the average number of exchange errors between the best DPD-based strategy and their non-DPD counterpart. Figure 5 shows the most prominent exchange and non-exchange (i.e. insertion or deletion) error differences between the strategies. On average, GRU-DPD-BST makes 130 fewer mistakes than GRU-BST among 1731 test entries on Rom-phon, with a majority (72%) being insertion or deletion error reductions. Trans-DPD-ΠM-BST makes 144 fewer mistakes than Trans-ΠM-BST on WikiHan among 1033 test entries, with a majority (74%) being exchange error reductions. Some error differences between DPD and its non-DPD counterpart appear highly dependent on the dataset, such as GRU-DPD-BST making on average about 13 less [o]/[ɔ] errors on group 4 but more such errors on all other groups. Differences between groups could indicate that the distribution of additional unlabeled cognate sets in the training data plays an important role in the error patterns of DPD-BST strategies.

On Rom-phon, DPD is better at deciding whether to insert [m] in the reconstruction. This is the exact type of problem DPD is designed to handle: in situations where it is not apparent from the reflexes whether a phoneme absent in the reflexes should be added, it is often the case that one of the decisions will lead to lost information in the reconstruction. In DPD, the reflex prediction subnetwork should be able to detect such information

[13]The [i]/[ɪ] exchange error, for example, refers to [i] being predicted in place of [ɪ] or vice versa.

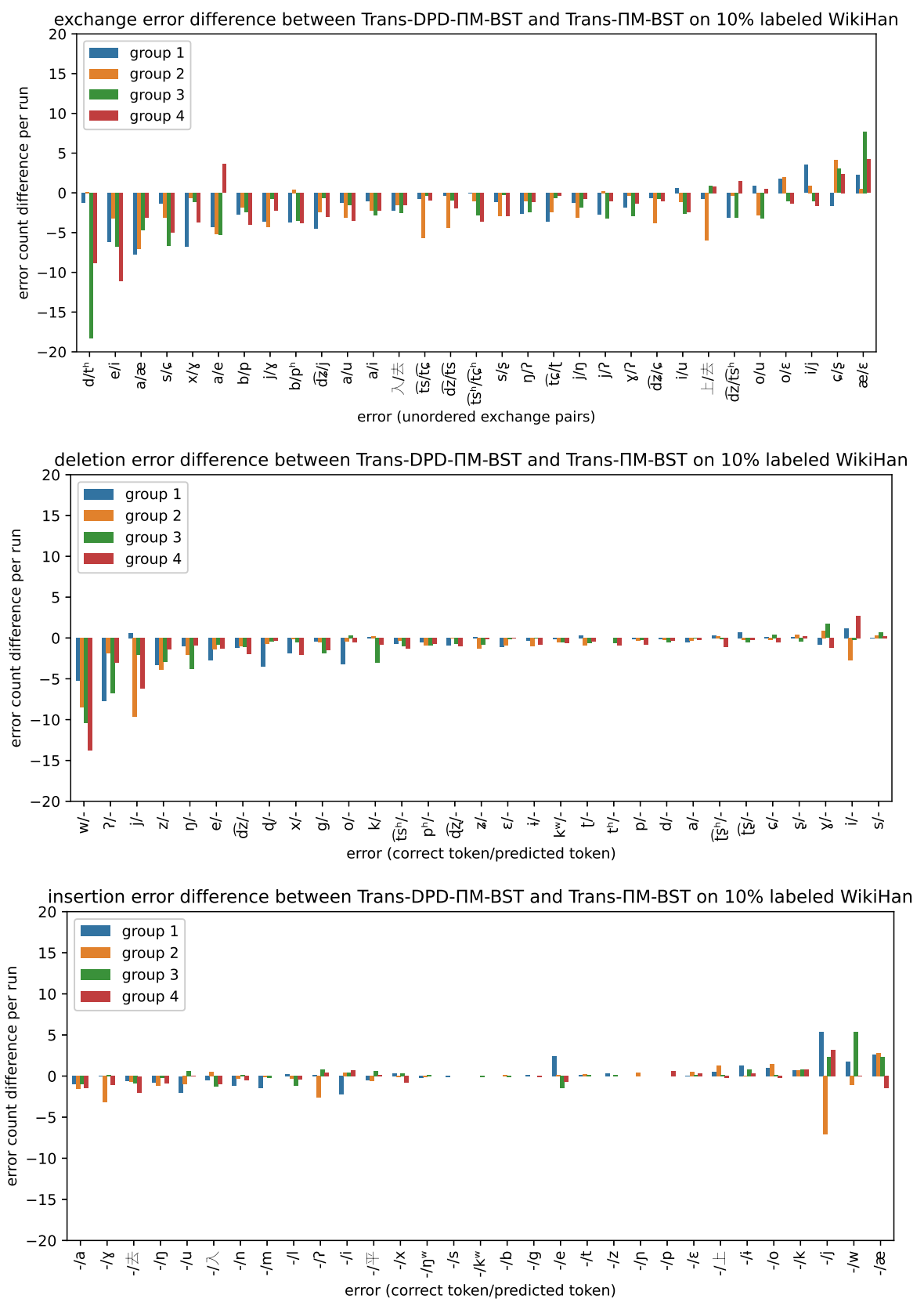

(a) Error count differences between Trans-DPD-ΠM-BST and Trans-ΠM-BST on WikiHan.

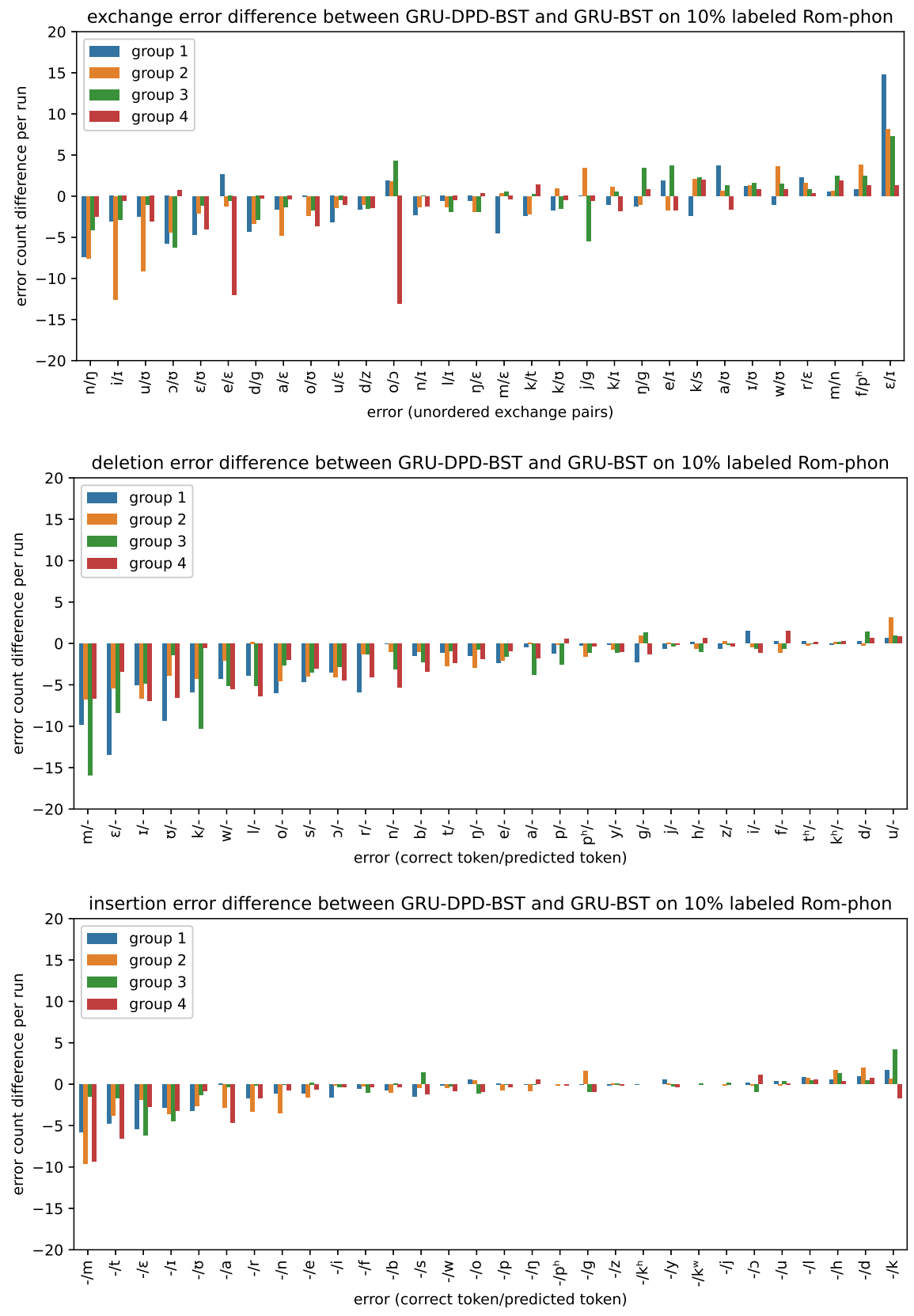

(b) Error count differences between GRU-DPD-BST and GRU-BST on Rom-phon.

Figure 5: Most notable error count differences per run between the best-performing strategy-architecture combination and their non-DPD counterpart, averaged across 10 runs for each group on 10% labeled datasets. A negative count difference indicates fewer mistakes made by a DPD-based strategy. Exchange error pairs (top) are in no specific order. insertion or deletion errors (middle and bottom) are ordered and the x-labels indicate the target phoneme followed by the predicted phoneme, separated by a slash.

loss, which would otherwise lead to reflexes not being inferrable from the reconstruction.

We compare errors made by the best GRU-DPD-BST against GRU-BST (within group 1 at 10% labeling setting). Among test examples where only one of GRU-DPD-BST and GRU-BST is correct, we see 12[14] instances where GRU-DPD-BST produces the correct reconstruction but GRU-BST makes a mistake on [m] insertion or deletion. The opposite is true in only 2 instances[15]. Table 7 shows three test examples where GRU-DPD-BST performs better at deciding whether to add an [m] in the reconstruction as well as one example where it fails.

We hypothesize that since DPD makes better use of unlabeled data, an error is less likely to occur if there is an abundance of unlabeled examples involving information about the underlying sound change pattern governing the context of a possible error. However, without explicit rules being learned by the models, it is difficult to assess what additional information is available to DPD in the unlabeled portion of the dataset and how the model learns from it. In an attempt to test this hypothesis, we estimate the abundance of learning resources in the unlabeled dataset by counting the number of examples involving the same sound correspondences (using lingpy Meloni et al. (2021)'s multi-sequence alignment) as a heuristic. We find no indication that count differences in exchange, insertion, or deletion errors are correlated with a higher abundance of training examples with the same sound correspondence in the unlabeled portion of the train set. Previous work has observed that statistical baselines can perform better than neural models at identifying the sound correct correspondences Fourrier and Sagot (2022), implying that sound correspondence is not necessarily a

[14] excluding two false positives by lingpy
[15] excluding one false positive by lingpy

| | konsɛntirɛ | pɔpʊlarɛm | ɪmmatɛrɪalɛm |
|---|---|---|---|
| French | k ɔ̃ - s ɑ̃ - - t i ʁ - - | p o p y l ɛ ʁ - - | i m m a t e ʁ j ɛ l - - |
| Italian | k o n s e - n t i r e - | p o p o l a r e - | i m m a t e r i a l e - |
| Romanian | k o n s i - m t - sʲ - - | p o p u l a r - - | i - m a t e ɾ j a l - - |
| Spanish | k o n s e - n t i ɾ - - | p o p u l a ɾ - - | i n m a t e ɾ j a l - - |
| Portuguese | k u ŋ s e ɪ ŋ t i ɹ - - | p u p u l a ɹ - - | i - m ɐ t e ɾ i ɑ l - - |
| Latin | k o n s ɛ - n t i r ɛ - | p ɔ p ʊ l a r ɛ m | ɪ m m a t ɛ r ɪ a l ɛ m |
| GRU-DPD-BST prediction | k o n s ɛ - n t i r ɛ - | p ɔ p ʊ l a r ɛ m | ɪ m m a t ɛ r ɪ a l ɛ m |
| GRU-BST prediction | k ɔ n s ɛ - m t ɪ r ɛ m | p ɔ p ɔ l a r ɛ - | ɪ - m a t ɛ r ɪ a l ɛ m |

| | traditorɛm |
|---|---|
| French | t ʁ ɛ - - - t - ʁ - - |
| Italian | t r a - d i t o r e - |
| Romanian | t r ə - d ə t o r - - |
| Spanish | t ɾ a ɪ ð - - o ɾ - - |
| Portuguese | t ɹ a ɪ d - - o ɹ - - |
| Latin | t r a - d ɪ t o r ɛ m |
| GRU-DPD-BST prediction | t r a ɪ d - - ʊ r ɛ - |
| GRU-BST prediction | t r a - d ɪ t o r ɛ m |

Table 7: Instances where GRU-DPD-BST produces the correct protoform but GRU-BST makes a [m] insertion or deletion error (top) and an instance where the opposite is true (bottom). The success examples are *immaterialem* 'immaterial', *consentire* 'to agree', and *popularem* 'popular'. The failure example is *traditorem* 'traitor'. Words are aligned manually, and '-' indicates an empty position in the alignment. The positions where an [m] insertion or deletion error occur are shaded.

proxy for sound change rules. It is likely that our DPD models learnt to use unlabeled cognate sets via means other than just sound correspondences.

## I Details on Unlabeled Data Ablation Experiments

We perform ablations on unlabeled training at a 10% labeling setting using the group 1 dataset seed, effectively keeping the same subset of labeled data as semisupervised group 1 experiments. This is equivalent to supervised reconstruction but with only 362 and 607 training examples for WikiHan and Rom-phon respectively.

We perform additional hyperparameter tuning for non-trivial semisupervised strategies[16] to account for a large difference in dataset size. The hyperparameter we use are reported in Tables 18 and 19. We then perform 10 runs (random seed) for each strategy other than SUPV.

Table 13 compares the performance of various strategies after ablation, and Table 14 compares the performance with base strategies including ΠM, DPD, and DPD-ΠM under the configuration of whether to exclude unlabeled data, include unlabeled data, or include and pseudo-label unlabeled data (+BST, include unlabelled).

In many situations, the non-trivial semisupervised strategies still perform significantly better than SUPV despite the lack of unlabeled data. On WikiHan, DPD-ΠM performs the best on all metrics except Trans-DPD-ΠM on BCFS. On Rom-Phon, ΠM performs the best on all metrics. For strategies involving DPD but not ΠM, performance is always higher when unlabeled cognate sets are used. Interestingly, strategies involving ΠM sometimes perform worse when using unlabeled cognate sets on Rom-phon. In almost all situations, configurations that include and pseudo-label unlabeled cognate sets perform significantly better than when unlabeled data is excluded.

## J Details on 100% Labeled Supervised Reconstruction Experiments

Similar to the setup for unlabeled data ablation at 10% labeling setting (see Section I), we tune additional hyperparameters for ΠM, DPD, and DPD-ΠM to account for a difference in data. The hyperparameters we obtain are reported in Tables 20 and 21.

The SUPV strategy is equivalent to existing reconstruction models: GRU-SUPV is equivalent to

[16]For SUPV, including unlabeled data has no effect, so we reuse hyperparameters and runs from semisupervised reconstruction experiments.

Meloni et al. (2021) and Trans-SUPV is equivalent Kim et al. (2023). Existing hyperparameters and checkpoints exist for SUPV: for WikiHan, we obtain 10 checkpoints from Kim et al. (2023) and 10 additional checkpoints from Lu et al. (2024); for Rom-phon, we obtain 20 checkpoints from Lu et al. (2024) (for hyperparameters, refer to Kim et al. (2023) and Lu et al. (2024)). For all other strategies, we perform 10 runs (random seed).

We compare the results of ΠM, DPD, and DPD-ΠM (both GRU and Transformer) against existing supervised reconstruction systems in Table 15. We use Meloni et al. (2021) and Kim et al. (2023) as SUPV baselines, and Lu et al. (2024)'s state-of-the-art supervised reconstruction systems (both GRU-BS + GRU Reranker and GRU-BS + Trans. Reranker) as strong baselines. We obtain evaluation results for 20 runs per setup from Lu et al. (2024).

## K Sample Outputs at Different Labeling Settings

We compare the protoform predictions of the best-performing strategy-architecture combination (best at a 10% labeling setting) trained on different labeling settings (with group 1 dataset seed)[17]. We present sample predictions stratified by the following categories:

- The predictions are correct at all %
- The predictions are correct only above a certain % threshold
- The predictions are correct only below a certain % threshold
- The predictions are incorrect at all %
- All other patterns

Table 23 shows the distribution of test examples in these categories. Tables 25 and 24 show sample protoform predictions in each of these categories (proportionate sampling). For both datasets, correct only above a certain % threshold is the most common category when the predictions differ between labeling setting.

## L Sample Outputs from Different Strategies

We show sample predictions from the best performing runs in group 1. Tables 28 and 27 show predictions of the best-performing strategy-architecture combination and its non-DPD counterpart for each dataset, proportionately sampled by the confusion matrices in Table 26. Tables 30 and 29 compares predictions across all 16 strategy-architecture pairs for 4 randomly selected test examples among those with the most diverse (approximately in the upper quartile[18]) protoform predictions.

## M Performance Data Visualizations

Figures 6 and 7 show the distribution of semisupervised reconstruction performance by group on 10% labeled datasets. Figures 8 and 9 visualize the group 1 performance of each strategy-architecture combination on datasets with different percentages of label.

## N Additional Phonetic Probing Results

A complete set of hierarchical clusterings of learned phoneme embeddings can be found in Figures 10 (Rom-phon) and 11 (WikiHan). Similar to French and Cantonese, phoneme embeddings learned using a DPD-based strategy also appear more interpretable for other languages. For instance, Trans-DPD-ΠM-BST creates a big cluster of tones within all phonemes present in daughter languages in WikiHan, and GRU-DPD-BST creates a cluster encompassing most palatalized consonants in Romanian.

## O DPD-ΠM Implementation

In a hybrid model combining the DPD architecture with Π-model, we first augment the input cognate set to obtain two augmented inputs $A$ and $B$. Input $A$ is used to train DPD as described in Figure 1. Input $B$ is fed through D2P (and not P2D). A mean square difference loss then is used to minimize the difference between the D2P classifier logits given $A$ and $B$ as inputs.

## P Responsible Use of AI

GitHub Copilot has been used as a coding assistant for our model implementation and data analysis. All code generated by Github Copilot is checked manually.

[17] At a 100% labeling setting, BST has no effect, so Trans-DPD-ΠM-BST is equivalent to Trans-DPD-ΠM and GRU-DPD-BST is equivalent to GRU-DPD.

[18] The precise cutoff is top 25.65% for WikiHan and top 24.91% for Rom-phon, accounting for ties and integer divisions.

| **Architecture** | **Strategy** | **ACC% ↑** | **TED ↓** | **TER ↓** | **FER ↓** | **BCFS ↑** |
|---|---|---|---|---|---|---|
| Transformer | DPD-ΠM-BST (ours) | 21.85% ❶ | 1.5150 ❶ | 0.3549 ❶ | 0.1575 ❶ | 0.5485 ❶ |
| | DPD-BST (ours) | 23.41% ❶ | 1.4501 ❶ | 0.3397 ❶ | 0.1441 ❶ | 0.5676 ❶ |
| | DPD-ΠM (ours) | 23.90% ❶ | 1.4488 ❶ | 0.3394 ❶ | 0.1445 ❶ | 0.5622 ❶ |
| | DPD (ours) | **25.46%** ❶ | **1.3682** ❶ | **0.3205** ❶ | **0.1374** ❶ | **0.5747** ❶ |
| | ΠM-BST | 16.04% | 1.7841 | 0.4179 | 0.1805 | 0.4872 |
| | BST (Lee, 2013) | 16.47% | 1.7365 | 0.4068 | 0.1736 | 0.4897 |
| | ΠM (Laine and Aila, 2017) | 18.18% | 1.6594 | 0.3887 | 0.1672 | 0.5076 |
| | SUPV | 16.12% | 1.7178 | 0.4024 | 0.1703 | 0.4927 |
| GRU | DPD-ΠM-BST (ours) | **27.70%** ❶ | **1.2834** ❶ | **0.3006** ❶ | **0.1207** ❶ | **0.6144** ❶ |
| | DPD-BST (ours) | 22.85% ① | 1.4037 ❶ | 0.3288 ❶ | 0.1338 ❶ | 0.5865 ❶ |
| | DPD-ΠM (ours) | 25.33% ❶ | 1.3600 ❶ | 0.3186 ❶ | 0.1286 ❶ | 0.5901 ❶ |
| | DPD (ours) | 21.39% | 1.4842 | 0.3477 | 0.1419 | 0.5570 |
| | ΠM-BST | 21.52% | 1.4863 | 0.3481 | 0.1462 | 0.5666 |
| | BST (Lee, 2013) | 16.22% | 1.6901 | 0.3959 | 0.1656 | 0.5152 |
| | ΠM (Laine and Aila, 2017) | 20.57% | 1.5301 | 0.3584 | 0.1464 | 0.5503 |
| | SUPV | 15.99% | 1.6812 | 0.3938 | 0.1616 | 0.5140 |

| **Architecture** | **Strategy** | **ACC% ↑** | **TED ↓** | **TER ↓** | **FER ↓** | **BCFS ↑** |
|---|---|---|---|---|---|---|
| Transformer | DPD-ΠM-BST (ours) | **47.02%** ❶ | **0.8648** ❶ | **0.2026** ❶ | **0.0816** | **0.7049** ❶ |
| | DPD-BST (ours) | 45.90% | 0.8773 | 0.2055 | 0.0834 | 0.7027 |
| | DPD-ΠM (ours) | 44.72% | 0.9246 | 0.2166 | 0.0881 | 0.6884 |
| | DPD (ours) | 45.03% | 0.9087 | 0.2129 | 0.0896 | 0.6916 |
| | ΠM-BST | 45.48% | 0.8905 | 0.2086 | 0.0848 | 0.6979 |
| | BST (Lee, 2013) | 45.30% | 0.8887 | 0.2082 | 0.0840 | 0.6988 |
| | ΠM (Laine and Aila, 2017) | 42.26% | 0.9758 | 0.2286 | 0.0949 | 0.6739 |
| | SUPV | 42.62% | 0.9641 | 0.2258 | 0.0945 | 0.6745 |
| GRU | DPD-ΠM-BST (ours) | **46.17%** ❶ | 0.9069 ❶ | 0.2124 ❶ | **0.0851** ❶ | **0.6941** ❶ |
| | DPD-BST (ours) | 45.23% ① | **0.9045** ❶ | **0.2119** ❶ | 0.0861 ❶ | 0.6940 ❶ |
| | DPD-ΠM (ours) | 44.12% ① | 0.9554 ① | 0.2238 ① | 0.0909 ① | 0.6794 ① |
| | DPD (ours) | 43.11% ① | 0.9605 ① | 0.2250 ① | 0.0906 ① | 0.6764 ① |
| | ΠM-BST | 44.32% ① | 0.9475 ① | 0.2220 ① | 0.0900 ① | 0.6839 ① |
| | BST (Lee, 2013) | 40.47% | 1.0143 | 0.2376 | 0.0953 | 0.6614 |
| | ΠM (Laine and Aila, 2017) | 41.37% | 1.0056 | 0.2356 | 0.0957 | 0.6661 |
| | SUPV | 39.24% | 1.0448 | 0.2447 | 0.0992 | 0.6530 |

| **Architecture** | **Strategy** | **ACC% ↑** | **TED ↓** | **TER ↓** | **FER ↓** | **BCFS ↑** |
|---|---|---|---|---|---|---|
| Transformer | DPD-ΠM-BST (ours) | 49.93% | 0.8229 | 0.1928 | 0.0789 | 0.7163 |
| | DPD-BST (ours) | 49.99% | 0.8170 | 0.1914 | 0.0782 | 0.7177 |
| | DPD-ΠM (ours) | **50.44%** | **0.8090** | **0.1895** | **0.0778** | **0.7212** |
| | DPD (ours) | 48.66% | 0.8457 | 0.1981 | 0.0811 | 0.7078 |
| | ΠM-BST | 49.64% | 0.8268 | 0.1937 | 0.0790 | 0.7149 |
| | BST (Lee, 2013) | 49.50% | 0.8270 | 0.1937 | 0.0782 | 0.7134 |
| | ΠM (Laine and Aila, 2017) | 47.56% | 0.8700 | 0.2038 | 0.0834 | 0.7025 |
| | SUPV | 46.92% | 0.8842 | 0.2071 | 0.0856 | 0.6958 |
| GRU | DPD-ΠM-BST (ours) | 49.43% ① | 0.8389 ❶ | 0.1965 ❶ | **0.0789** ❶ | 0.7126 ❶ |
| | DPD-BST (ours) | 48.79% ① | **0.8342** ❶ | **0.1954** ❶ | 0.0797 ① | 0.7116 ❶ |
| | DPD-ΠM (ours) | **49.84%** ❶ | 0.8370 ❶ | 0.1961 ❶ | 0.0791 ❶ | **0.7136** ❶ |
| | DPD (ours) | 47.67% ① | 0.8696 | 0.2037 | 0.0824 | 0.7020 ① |
| | ΠM-BST | 47.73% | 0.8788 | 0.2059 | 0.0825 | 0.7005 |
| | BST (Lee, 2013) | 45.59% | 0.9066 | 0.2124 | 0.0853 | 0.6897 |
| | ΠM (Laine and Aila, 2017) | 46.28% | 0.9045 | 0.2119 | 0.0857 | 0.6924 |
| | SUPV | 44.82% | 0.9199 | 0.2155 | 0.0866 | 0.6867 |

Table 8: Performance of all strategies on 5% (top), 20% (middle), and 30% (bottom) labeled WikiHan for each architecture, averaged across 10 runs in group 1. Bold: best-performing strategy for the corresponding architecture; ①: significantly better than all weak baselines (SUPV, BST, and ΠM) with $p < 0.01$; ❶: significantly better than the ΠM-BST strong baseline and all weak baselines with $p < 0.01$.

| Architecture | Strategy | ACC% ↑ | TED ↓ | TER ↓ | FER ↓ | BCFS ↑ |
|---|---|---|---|---|---|---|
| Transformer | DPD-ΠM-BST (ours) | **27.38%** ❶ | **1.5508** ❶ | **0.1730** ❶ | **0.0717** ❶ | **0.7523** ❶ |
| | DPD-BST (ours) | 26.04% ❶ | 1.6169 ❶ | 0.1803 ❶ | 0.0751 ❶ | 0.7403 ❶ |
| | DPD-ΠM (ours) | 16.33% | 2.3046 | 0.2570 | 0.1149 | 0.6474 |
| | DPD (ours) | 23.52% ❶ | 1.8284 ❶ | 0.2039 ❶ | 0.0854 ❶ | 0.7110 ❶ |
| | ΠM-BST | 18.19% | 2.1148 | 0.2359 | 0.1127 | 0.6724 |
| | BST (Lee, 2013) | 16.65% | 2.2502 | 0.2510 | 0.1222 | 0.6560 |
| | ΠM (Laine and Aila, 2017) | 10.66% | 2.8262 | 0.3152 | 0.1468 | 0.5806 |
| | SUPV | 14.54% | 2.4150 | 0.2694 | 0.1188 | 0.6314 |
| GRU | DPD-ΠM-BST (ours) | 30.68% | 1.3867 ❶ | 0.1547 ❶ | 0.0562 ❶ | 0.7788 ❶ |
| | DPD-BST (ours) | **30.94%** | **1.3731** ❶ | **0.1531** ❶ | **0.0518** ❶ | **0.7803** ❶ |
| | DPD-ΠM (ours) | 25.48% | 1.6870 | 0.1882 | 0.0740 | 0.7328 |
| | DPD (ours) | 24.13% | 1.7210 | 0.1919 | 0.0733 | 0.7269 |
| | ΠM-BST | 28.96% | 1.4869 | 0.1658 | 0.0654 | 0.7673 |
| | BST (Lee, 2013) | 30.29% | 1.4572 | 0.1625 | 0.0605 | 0.7687 |
| | ΠM (Laine and Aila, 2017) | 22.81% | 1.8119 | 0.2021 | 0.0809 | 0.7141 |
| | SUPV | 24.11% | 1.7354 | 0.1936 | 0.0747 | 0.7255 |

| Architecture | Strategy | ACC% ↑ | TED ↓ | TER ↓ | FER ↓ | BCFS ↑ |
|---|---|---|---|---|---|---|
| Transformer | DPD-ΠM-BST (ours) | **40.60%** ① | **1.1887** ① | **0.1326** ① | 0.0531 ① | **0.8004** ① |
| | DPD-BST (ours) | 39.49% ① | 1.2052 ① | 0.1344 ① | **0.0515** ❶ | 0.7968 |
| | DPD-ΠM (ours) | 38.84% | 1.2421 | 0.1385 | 0.0556 | 0.7921 |
| | DPD (ours) | 37.79% | 1.3035 | 0.1454 | 0.0562 | 0.7824 |
| | ΠM-BST | 40.22% ① | 1.2023 ① | 0.1341 ① | 0.0536 | 0.7982 ① |
| | BST (Lee, 2013) | 38.18% | 1.2467 | 0.1390 | 0.0553 | 0.7919 |
| | ΠM (Laine and Aila, 2017) | 37.98% | 1.2797 | 0.1427 | 0.0585 | 0.7865 |
| | SUPV | 35.14% | 1.3910 | 0.1551 | 0.0609 | 0.7699 |
| GRU | DPD-ΠM-BST (ours) | **42.57%** ❶ | 1.1351 ❶ | 0.1266 ❶ | 0.0468 | 0.8080 ① |
| | DPD-BST (ours) | 42.13% ❶ | **1.1079** ❶ | **0.1236** ❶ | **0.0426** ❶ | **0.8134** ❶ |
| | DPD-ΠM (ours) | 38.51% | 1.2476 | 0.1392 | 0.0519 | 0.7922 |
| | DPD (ours) | 37.47% | 1.2702 | 0.1417 | 0.0510 | 0.7890 |
| | ΠM-BST | 41.01% | 1.1721 | 0.1307 | 0.0490 | 0.8045 |
| | BST (Lee, 2013) | 40.33% | 1.1779 | 0.1314 | 0.0474 | 0.8031 |
| | ΠM (Laine and Aila, 2017) | 36.71% | 1.2994 | 0.1449 | 0.0529 | 0.7860 |
| | SUPV | 37.25% | 1.2947 | 0.1444 | 0.0524 | 0.7834 |

| Architecture | Strategy | ACC% ↑ | TED ↓ | TER ↓ | FER ↓ | BCFS ↑ |
|---|---|---|---|---|---|---|
| Transformer | DPD-ΠM-BST (ours) | **45.05%** ① | **1.0812** ❶ | **0.1206** ❶ | **0.0475** | **0.8162** ❶ |
| | DPD-BST (ours) | 43.28% | 1.1281 | 0.1258 | 0.0482 | 0.8089 |
| | DPD-ΠM (ours) | 42.65% | 1.1182 ① | 0.1247 ① | 0.0501 | 0.8132 |
| | DPD (ours) | 43.07% | 1.1588 | 0.1293 | 0.0492 | 0.8034 |
| | ΠM-BST | 44.21% ① | 1.1196 | 0.1249 | 0.0499 | 0.8102 |
| | BST (Lee, 2013) | 42.76% | 1.1515 | 0.1284 | 0.0485 | 0.8045 |
| | ΠM (Laine and Aila, 2017) | 41.61% | 1.1447 | 0.1277 | 0.0519 | 0.8095 |
| | SUPV | 40.61% | 1.2381 | 0.1381 | 0.0526 | 0.7912 |
| GRU | DPD-ΠM-BST (ours) | 45.19% ① | 1.0966 | 0.1223 | 0.0457 | 0.8127 |
| | DPD-BST (ours) | **45.74%** ① | **1.0417** ❶ | **0.1162** ❶ | **0.0400** ❶ | **0.8214** ❶ |
| | DPD-ΠM (ours) | 43.31% | 1.1358 | 0.1267 | 0.0466 | 0.8089 |
| | DPD (ours) | 41.57% | 1.1681 | 0.1303 | 0.0473 | 0.8037 |
| | ΠM-BST | 44.70% | 1.1134 | 0.1242 | 0.0472 | 0.8117 |
| | BST (Lee, 2013) | 44.05% | 1.1081 | 0.1236 | 0.0446 | 0.8125 |
| | ΠM (Laine and Aila, 2017) | 41.11% | 1.1838 | 0.1320 | 0.0482 | 0.8023 |
| | SUPV | 40.94% | 1.1985 | 0.1337 | 0.0479 | 0.7983 |

Table 9: Performance of all strategies on 5% (top), 20% (middle), and 30% (bottom) labeled Rom-phon for each architecture, averaged across 10 runs in group 1. Bold: best-performing strategy for the corresponding architecture; ①: significantly better than all weak baselines (SUPV, BST, and ΠM) with $p < 0.01$; ❶: significantly better than the ΠM-BST strong baseline and all weak baselines with $p < 0.01$.

| **Architecture** | **Strategy** | **ACC% ↑** | **TED ↓** | **TER ↓** | **FER ↓** | **BCFS ↑** |
|---|---|---|---|---|---|---|
| Transformer | DPD-ΠM-BST (ours) | **41.15%** ❶❷❸❹ | **0.9955** ❶❷❸❹ | **0.2326** ❶❷❸❹ | **0.0920** ❶❷❸❹ | **0.6703** ❶❷❸❹ |
| | DPD-BST (ours) | 39.83% ❶❷❸❹ | 1.0212 ❶❷❸❹ | 0.2386 ❶❷❸❹ | 0.0944 ❶❷❸❹ | 0.6636 ❶❷❸❹ |
| | DPD-ΠM (ours) | 38.28% ①❷❸❹ | 1.0612 ①❷❸❹ | 0.2480 ①❷❸❹ | 0.0968 ①❷❸❹ | 0.6478 ❷❸❹ |
| | DPD (ours) | 39.61% ❶❷❸❹ | 1.0295 ❶❷❸❹ | 0.2406 ❶❷❸❹ | 0.0946 ❶❷❸❹ | 0.6547 ❶❷❸❹ |
| | ΠM-BST | 34.76% | 1.1447 | 0.2675 | 0.1078 | 0.6336 |
| | BST (Lee, 2013) | 34.93% | 1.1490 | 0.2685 | 0.1086 | 0.6281 |
| | ΠM (Laine and Aila, 2017) | 34.43% | 1.1721 | 0.2739 | 0.1087 | 0.6151 |
| | SUPV | 33.57% | 1.1920 | 0.2785 | 0.1110 | 0.6078 |
| GRU | DPD-ΠM-BST (ours) | **40.50%** ❶❷❸❹ | **0.9977** ❶❷❸❹ | **0.2331** ❶❷❸❹ | **0.0894** ❶❷❸❹ | **0.6716** ❶❷❸❹ |
| | DPD-BST (ours) | 36.64% ①❷③ | 1.0794 ①❷❸④ | 0.2522 ①❷❸④ | 0.0978 ①❷❸ | 0.6512 ①❷❸④ |
| | DPD-ΠM (ours) | 38.67% ①❷❸❹ | 1.0460 ①❷❸❹ | 0.2444 ①❷❸❹ | 0.0933 ①❷❸❹ | 0.6533 ①❷❸❹ |
| | DPD (ours) | 35.12% ① | 1.1339 ① | 0.2650 ① | 0.1030 | 0.6282 ① |
| | ΠM-BST | 35.28% ① | 1.1266 ①③ | 0.2632 ①③ | 0.1016 ① | 0.6360 ①②③ |
| | BST (Lee, 2013) | 29.24% | 1.2929 | 0.3021 | 0.1174 | 0.5931 |
| | ΠM (Laine and Aila, 2017) | 33.69% | 1.1774 | 0.2751 | 0.1063 | 0.6173 |
| | SUPV | 28.98% | 1.3066 | 0.3053 | 0.1185 | 0.5827 |

| **Architecture** | **Strategy** | **ACC% ↑** | **TED ↓** | **TER ↓** | **FER ↓** | **BCFS ↑** |
|---|---|---|---|---|---|---|
| Transformer | DPD-ΠM-BST (ours) | **32.94%** ❶❷❸❹ | **1.3682** ❶❷❸❹ | **0.1533** ❶❷❸❹ | **0.0597** ❶❷❸❹ | **0.7717** ❶❷❸❹ |
| | DPD-BST (ours) | 31.80% ❶❷③❹ | 1.4636 ❶❷③❹ | 0.1640 ❶❷③❹ | 0.0634 ❶❷❸❹ | 0.7553 ❶❷③❹ |
| | DPD-ΠM (ours) | 27.58% | 1.6277 | 0.1824 | 0.0732 ①② | 0.7328 |
| | DPD (ours) | 30.15% ❶②③ | 1.5413 ①②③④ | 0.1727 ①②③④ | 0.0678 ①②③④ | 0.7449 ①②③④ |
| | ΠM-BST | 30.36% ①②③④ | 1.5174 ①②③④ | 0.1700 ①②③④ | 0.0681 ①②③④ | 0.7487 ①②③④ |
| | BST (Lee, 2013) | 28.70% | 1.6216 | 0.1817 | 0.0754 | 0.7332 |
| | ΠM (Laine and Aila, 2017) | 25.18% | 1.7561 | 0.1968 | 0.0804 | 0.7144 |
| | SUPV | 25.37% | 1.7756 | 0.1990 | 0.0802 | 0.7122 |
| GRU | DPD-ΠM-BST (ours) | 35.34% ❶❹ | 1.2634 ❶❷❸❹ | 0.1416 ❶❷❸❹ | 0.0480 ❶❷❸❹ | 0.7899 ❶❷❸❹ |
| | DPD-BST (ours) | **35.83%** ❶❷❸❹ | **1.2402** ❶❷❸❹ | **0.1390** ❶❷❸❹ | **0.0458** ❶❷❸❹ | **0.7929** ❶❷❸❹ |
| | DPD-ΠM (ours) | 30.19% | 1.5103 | 0.1692 | 0.0619 | 0.7511 |
| | DPD (ours) | 29.83% | 1.5071 | 0.1689 | 0.0604 | 0.7512 |
| | ΠM-BST | 34.31% | 1.3134 ① | 0.1472 ① | 0.0523 | 0.7846 ①④ |
| | BST (Lee, 2013) | 34.22% | 1.3163 | 0.1475 | 0.0514 | 0.7819 |
| | ΠM (Laine and Aila, 2017) | 28.16% | 1.5648 | 0.1754 | 0.0636 | 0.7447 |
| | SUPV | 29.30% | 1.5292 | 0.1714 | 0.0610 | 0.7473 |

Table 10: Transductive evaluation of all strategies on 10% labeled WikiHan (top) and Rom-phon (bottom) for each architecture, averaged across all runs in four groups (10 runs per strategy-architecture combination per group). Bold: best-performing strategy for the corresponding architecture and dataset; ①: significantly better than all weak baselines (SUPV, BST, and ΠM) on group 1 with $p < 0.01$; ❶: significantly better than the ΠM-BST strong baseline and all weak baselines on group 1 with $p < 0.01$; ②, ③, ④, ❷, ❸, ❹: likewise for groups 2, 3, and 4.

| Architecture | Strategy | ACC% ↑ | TED ↓ | TER ↓ | FER ↓ | BCFS ↑ |
|---|---|---|---|---|---|---|
| Transformer | DPD-ΠM-BST (ours) | 21.40% ❶ | 1.5246 ❶ | 0.3564 ❶ | 0.1525 ❶ | 0.5403 ❶ |
| | DPD-BST (ours) | 23.74% ❶ | 1.4558 ❶ | 0.3403 ❶ | 0.1409 ❶ | 0.5608 ❶ |
| | DPD-ΠM (ours) | 23.98% ❶ | 1.4538 ❶ | 0.3398 ❶ | 0.1407 ❶ | 0.5564 ❶ |
| | DPD (ours) | **25.35%** ❶ | **1.3911** ❶ | **0.3252** ❶ | **0.1348** ❶ | **0.5636** ❶ |
| | ΠM-BST | 15.97% | 1.7986 | 0.4204 | 0.1786 | 0.4774 |
| | BST (Lee, 2013) | 16.43% | 1.7464 | 0.4082 | 0.1681 | 0.4805 |
| | ΠM (Laine and Aila, 2017) | 17.80% | 1.6913 | 0.3953 | 0.1665 | 0.4935 |
| | SUPV | 16.42% | 1.7418 | 0.4071 | 0.1681 | 0.4815 |
| GRU | DPD-ΠM-BST (ours) | **28.38%** ❶ | **1.2499** ❶ | **0.2922** ❶ | **0.1126** ❶ | **0.6175** ❶ |
| | DPD-BST (ours) | 24.32% ❶ | 1.3716 ❶ | 0.3206 ❶ | 0.1258 ❶ | 0.5876 ❶ |
| | DPD-ΠM (ours) | 25.73% ❶ | 1.3493 ❶ | 0.3154 ❶ | 0.1238 ❶ | 0.5887 ❶ |
| | DPD (ours) | 22.27% | 1.4420 | 0.3371 | 0.1329 ① | 0.5599 |
| | ΠM-BST | 22.06% | 1.4639 | 0.3422 | 0.1395 | 0.5641 ① |
| | BST (Lee, 2013) | 16.50% | 1.6554 | 0.3870 | 0.1596 | 0.5169 |
| | ΠM (Laine and Aila, 2017) | 20.89% | 1.5094 | 0.3528 | 0.1407 | 0.5486 |
| | SUPV | 16.72% | 1.6520 | 0.3861 | 0.1572 | 0.5120 |

| Architecture | Strategy | ACC% ↑ | TED ↓ | TER ↓ | FER ↓ | BCFS ↑ |
|---|---|---|---|---|---|---|
| Transformer | DPD-ΠM-BST (ours) | **48.17%** ❶ | **0.8513** ❶ | **0.1986** ❶ | **0.0749** ❶ | **0.7039** ❶ |
| | DPD-BST (ours) | 46.61% | 0.8666 ❶ | 0.2022 ❶ | 0.0768 | 0.7004 ① |
| | DPD-ΠM (ours) | 44.99% | 0.8983 | 0.2096 | 0.0798 | 0.6912 |
| | DPD (ours) | 44.88% | 0.9125 | 0.2129 | 0.0838 | 0.6852 |
| | ΠM-BST | 45.93% | 0.8904 | 0.2077 | 0.0788 | 0.6939 |
| | BST (Lee, 2013) | 45.43% | 0.8995 | 0.2099 | 0.0791 | 0.6904 |
| | ΠM (Laine and Aila, 2017) | 41.61% | 0.9829 | 0.2293 | 0.0898 | 0.6671 |
| | SUPV | 42.25% | 0.9698 | 0.2263 | 0.0888 | 0.6683 |
| GRU | DPD-ΠM-BST (ours) | **47.92%** ❶ | **0.8527** ❶ | **0.1990** ❶ | **0.0755** ❶ | **0.7031** ❶ |
| | DPD-BST (ours) | 46.07% ❶ | 0.8783 ❶ | 0.2049 ❶ | 0.0791 ❶ | 0.6966 ❶ |
| | DPD-ΠM (ours) | 44.82% ① | 0.9194 ① | 0.2145 ① | 0.0827 ① | 0.6839 ① |
| | DPD (ours) | 44.01% ① | 0.9309 ① | 0.2172 ① | 0.0855 ① | 0.6801 ① |
| | ΠM-BST | 44.73% ① | 0.9232 ① | 0.2154 ① | 0.0831 ① | 0.6836 ① |
| | BST (Lee, 2013) | 41.60% | 0.9888 | 0.2307 | 0.0908 | 0.6646 |
| | ΠM (Laine and Aila, 2017) | 42.34% | 0.9793 | 0.2285 | 0.0889 | 0.6666 |
| | SUPV | 40.12% | 1.0283 | 0.2399 | 0.0938 | 0.6518 |

| Architecture | Strategy | ACC% ↑ | TED ↓ | TER ↓ | FER ↓ | BCFS ↑ |
|---|---|---|---|---|---|---|
| Transformer | DPD-ΠM-BST (ours) | 50.21% | 0.8336 | 0.1941 | 0.0738 | 0.7087 |
| | DPD-BST (ours) | 50.42% ① | 0.8217 ❶ | 0.1913 ❶ | 0.0716 ❶ | 0.7121 ① |
| | DPD-ΠM (ours) | **50.60%** ① | **0.8105** ❶ | **0.1887** ❶ | **0.0716** ❶ | **0.7166** ❶ |
| | DPD (ours) | 48.51% | 0.8571 | 0.1995 | 0.0760 | 0.7013 |
| | ΠM-BST | 49.81% | 0.8389 | 0.1953 | 0.0744 | 0.7071 |
| | BST (Lee, 2013) | 49.40% | 0.8476 | 0.1973 | 0.0751 | 0.7046 |
| | ΠM (Laine and Aila, 2017) | 46.97% | 0.8905 | 0.2073 | 0.0804 | 0.6923 |
| | SUPV | 46.27% | 0.9019 | 0.2100 | 0.0805 | 0.6881 |
| GRU | DPD-ΠM-BST (ours) | 49.55% ① | **0.8296** ❶ | **0.1931** ❶ | 0.0737 ❶ | **0.7119** ❶ |
| | DPD-BST (ours) | 49.01% ① | 0.8413 ① | 0.1959 ① | 0.0746 ① | 0.7066 ❶ |
| | DPD-ΠM (ours) | **49.55%** ❶ | 0.8310 ❶ | 0.1935 ❶ | **0.0732** ❶ | 0.7107 ❶ |
| | DPD (ours) | 47.95% | 0.8635 | 0.2010 | 0.0770 | 0.7009 |
| | ΠM-BST | 48.31% ① | 0.8653 ① | 0.2015 ① | 0.0761 ① | 0.6996 ① |
| | BST (Lee, 2013) | 46.35% | 0.9051 | 0.2107 | 0.0812 | 0.6879 |
| | ΠM (Laine and Aila, 2017) | 46.00% | 0.9041 | 0.2105 | 0.0803 | 0.6893 |
| | SUPV | 44.53% | 0.9415 | 0.2192 | 0.0847 | 0.6787 |

Table 11: Transductive evaluation of all strategies on 5% (top), 20% (middle), and 30% (bottom) labeled WikiHan for each architecture, averaged across 10 runs in group 1. Bold: best-performing strategy for the corresponding architecture; ①: significantly better than all weak baselines (SUPV, BST, and ΠM) with $p < 0.01$; ❶: significantly better than the ΠM-BST strong baseline and all weak baselines with $p < 0.01$.

| Architecture | Strategy | ACC% ↑ | TED ↓ | TER ↓ | FER ↓ | BCFS ↑ |
|---|---|---|---|---|---|---|
| Transformer | DPD-ΠM-BST (ours) | **25.73%** ❶ | **1.6015** ❶ | **0.1797** ❶ | **0.0735** ❶ | **0.7399** ❶ |
| | DPD-BST (ours) | 24.62% ❶ | 1.7219 ❶ | 0.1932 ❶ | 0.0797 ❶ | 0.7199 ❶ |
| | DPD-ΠM (ours) | 15.46% | 2.3351 | 0.2620 | 0.1148 ① | 0.6387 |
| | DPD (ours) | 22.38% ❶ | 1.8658 ❶ | 0.2093 ❶ | 0.0861 ❶ | 0.7002 ❶ |
| | ΠM-BST | 17.29% | 2.2002 | 0.2468 | 0.1160 | 0.6544 |
| | BST (Lee, 2013) | 15.85% | 2.3260 | 0.2610 | 0.1248 | 0.6394 |
| | ΠM (Laine and Aila, 2017) | 9.67% | 2.8616 | 0.3210 | 0.1461 | 0.5706 |
| | SUPV | 14.06% | 2.4557 | 0.2755 | 0.1203 | 0.6219 |
| GRU | DPD-ΠM-BST (ours) | 29.62% ❶ | **1.3946** ❶ | **0.1565** ❶ | 0.0553 ❶ | **0.7724** ❶ |
| | DPD-BST (ours) | **29.67%** ❶ | 1.3977 ❶ | 0.1568 ❶ | **0.0519** ❶ | 0.7712 ❶ |
| | DPD-ΠM (ours) | 24.86% | 1.7148 | 0.1924 | 0.0731 | 0.7230 |
| | DPD (ours) | 24.02% | 1.7394 | 0.1951 | 0.0734 | 0.7185 |
| | ΠM-BST | 28.42% | 1.5011 | 0.1684 | 0.0648 | 0.7601 |
| | BST (Lee, 2013) | 28.49% | 1.5016 | 0.1685 | 0.0624 | 0.7568 |
| | ΠM (Laine and Aila, 2017) | 22.36% | 1.8316 | 0.2055 | 0.0807 | 0.7054 |
| | SUPV | 23.49% | 1.7649 | 0.1980 | 0.0751 | 0.7146 |

| Architecture | Strategy | ACC% ↑ | TED ↓ | TER ↓ | FER ↓ | BCFS ↑ |
|---|---|---|---|---|---|---|
| Transformer | DPD-ΠM-BST (ours) | **38.96%** ❶ | **1.2502** ❶ | **0.1404** ❶ | **0.0539** ❶ | **0.7860** ❶ |
| | DPD-BST (ours) | 37.75% ① | 1.3183 | 0.1480 | 0.0551 ① | 0.7740 |
| | DPD-ΠM (ours) | 36.67% | 1.2899 ① | 0.1448 ① | 0.0552 ① | 0.7804 ① |
| | DPD (ours) | 36.11% | 1.3521 | 0.1518 | 0.0569 | 0.7706 |
| | ΠM-BST | 38.04% ① | 1.2899 ① | 0.1448 ① | 0.0557 ① | 0.7796 ① |
| | BST (Lee, 2013) | 35.99% | 1.3854 | 0.1555 | 0.0609 | 0.7647 |
| | ΠM (Laine and Aila, 2017) | 35.57% | 1.3284 | 0.1491 | 0.0580 | 0.7748 |
| | SUPV | 33.61% | 1.4521 | 0.1630 | 0.0621 | 0.7569 |
| GRU | DPD-ΠM-BST (ours) | **41.19%** ❶ | 1.1635 ❶ | 0.1306 ❶ | 0.0462 ❶ | 0.7991 ① |
| | DPD-BST (ours) | 41.00% ❶ | **1.1345** ❶ | **0.1274** ❶ | **0.0425** ❶ | **0.8041** ❶ |
| | DPD-ΠM (ours) | 36.77% | 1.2877 | 0.1446 | 0.0513 | 0.7811 |
| | DPD (ours) | 35.94% | 1.3036 | 0.1464 | 0.0511 | 0.7788 |
| | ΠM-BST | 40.07% | 1.1982 | 0.1345 | 0.0482 | 0.7960 |
| | BST (Lee, 2013) | 39.21% | 1.2014 | 0.1349 | 0.0472 | 0.7945 |
| | ΠM (Laine and Aila, 2017) | 35.17% | 1.3368 | 0.1501 | 0.0530 | 0.7751 |
| | SUPV | 35.60% | 1.3277 | 0.1491 | 0.0524 | 0.7736 |

| Architecture | Strategy | ACC% ↑ | TED ↓ | TER ↓ | FER ↓ | BCFS ↑ |
|---|---|---|---|---|---|---|
| Transformer | DPD-ΠM-BST (ours) | **41.96%** ① | **1.1563** ❶ | **0.1302** ❶ | **0.0485** ❶ | **0.7998** ❶ |
| | DPD-BST (ours) | 40.53% ① | 1.2425 | 0.1399 | 0.0513 | 0.7853 |
| | DPD-ΠM (ours) | 39.29% | 1.1843 ① | 0.1333 ① | 0.0503 ① | 0.7974 ❶ |
| | DPD (ours) | 39.57% | 1.2368 | 0.1392 | 0.0509 | 0.7867 |
| | ΠM-BST | 41.02% ① | 1.1950 | 0.1345 | 0.0510 | 0.7929 |
| | BST (Lee, 2013) | 39.94% | 1.2865 | 0.1448 | 0.0542 | 0.7773 |
| | ΠM (Laine and Aila, 2017) | 37.92% | 1.2134 | 0.1366 | 0.0523 | 0.7933 |
| | SUPV | 37.44% | 1.3133 | 0.1478 | 0.0545 | 0.7755 |
| GRU | DPD-ΠM-BST (ours) | 43.29% ① | 1.1312 | 0.1273 | 0.0447 | 0.8023 |
| | DPD-BST (ours) | **43.92%** ❶ | **1.0702** ❶ | **0.1205** ❶ | **0.0402** ❶ | **0.8119** ❶ |
| | DPD-ΠM (ours) | 40.62% | 1.1738 | 0.1321 | 0.0461 | 0.7974 |
| | DPD (ours) | 38.84% | 1.2149 | 0.1368 | 0.0470 | 0.7903 |
| | ΠM-BST | 42.93% ① | 1.1497 | 0.1294 | 0.0464 | 0.8006 |
| | BST (Lee, 2013) | 41.81% | 1.1402 | 0.1284 | 0.0439 | 0.8016 |
| | ΠM (Laine and Aila, 2017) | 39.08% | 1.2188 | 0.1372 | 0.0475 | 0.7911 |
| | SUPV | 38.77% | 1.2268 | 0.1381 | 0.0471 | 0.7878 |

Table 12: Transductive evaluation of all strategies on 5% (top), 20% (middle), and 30% (bottom) labeled Rom-phon for each architecture, averaged across 10 runs in group 1. Bold: best-performing strategy for the corresponding architecture; ①: significantly better than all weak baselines (SUPV, BST, and ΠM) with $p < 0.01$; ❶: significantly better than the ΠM-BST strong baseline and all weak baselines with $p < 0.01$.

| **Architecture** | **Strategy** | **ACC% ↑** | **TED ↓** | **TER ↓** | **FER ↓** | **BCFS ↑** |
|---|---|---|---|---|---|---|
| Transformer | SUPV | 33.25% | 1.1891 | 0.2785 | 0.1140 | 0.6138 |
| | ΠM (Laine and Aila, 2017) | 32.21% | 1.2141 | 0.2844 | 0.1140 | 0.6077 |
| | DPD (ours) | 33.57% | 1.1621 | 0.2722 | 0.1108 | 0.6246 ❶ |
| | DPD-ΠM (ours) | **34.93%** ❶ | **1.1307** ❶ | **0.2649** ❶ | **0.1097** | **0.6344** ❶ |
| GRU | SUPV | 28.16% | 1.3257 | 0.3105 | 0.1234 | 0.5835 |
| | ΠM (Laine and Aila, 2017) | 29.46% ① | 1.2475 ① | 0.2922 ① | 0.1157 ① | **0.6124** ① |
| | DPD (ours) | 26.57% | 1.3424 | 0.3144 | 0.1259 | 0.5830 |
| | DPD-ΠM (ours) | **30.27%** ① | **1.2393** ① | **0.2903** ① | **0.1156** ① | 0.6067 ① |

| **Architecture** | **Strategy** | **ACC% ↑** | **TED ↓** | **TER ↓** | **FER ↓** | **BCFS ↑** |
|---|---|---|---|---|---|---|
| Transformer | SUPV | 26.99% | 1.7331 | 0.1933 | 0.0794 | 0.7218 |
| | ΠM (Laine and Aila, 2017) | **30.01%** ① | **1.5261** ① | **0.1702** ① | **0.0699** ① | **0.7536** ① |
| | DPD (ours) | 28.22% | 1.6713 | 0.1864 | 0.0745 | 0.7308 |
| | DPD-ΠM (ours) | 27.42% | 1.5860 ① | 0.1769 ① | 0.0741 | 0.7465 ① |
| GRU | SUPV | 30.69% | 1.5018 | 0.1675 | 0.0612 | 0.7558 |
| | ΠM (Laine and Aila, 2017) | **32.38%** ① | **1.4232** ① | **0.1587** ① | **0.0591** | **0.7718** ① |
| | DPD (ours) | 30.68% | 1.5010 | 0.1674 | 0.0629 | 0.7588 |
| | DPD-ΠM (ours) | 32.35% ① | 1.4294 ① | 0.1594 ① | 0.0596 | 0.7702 ① |

Table 13: Performance when unlabeled cognate sets are excluded for 10% labeled group 1 WikiHan (top) and 10% labeled group 1 Rom-phon (bottom), averaged across 10 runs. Bold: best-performing strategy for the corresponding architecture; ①: significantly better than SUPV ($p < 0.01$); ❶: significantly better than both SUPV and ΠM ($p < 0.01$).

| Architecture | Base Strategy | Configuration | ACC% ↑ | TED ↓ | TER ↓ | FER ↓ | BCFS ↑ |
|---|---|---|---|---|---|---|---|
| Transformer | SUPV | exclude/include unlabeled | 32.41% | 1.1922 | 0.2793 | 0.1145 | 0.6122 |
| | | +BST, include unlabeled | **34.87%** ❶ | **1.1277** ❶ | **0.2641** ❶ | **0.1088** ❶ | **0.6388** ❶ |
| | ΠM | exclude unlabeled | 32.21% | 1.2141 | 0.2844 | 0.1140 | 0.6077 |
| | | include unlabeled | **34.50%** ❶ | 1.1547 ❶ | 0.2705 ❶ | 0.1102 | 0.6256 ❶ |
| | | +BST, include unlabeled | 34.41% | **1.1348** ❶ | **0.2658** ❶ | **0.1067** ❶ | **0.6418** ❶ |
| | DPD | exclude unlabeled | 33.57% | 1.1621 | 0.2722 | 0.1108 | 0.6246 |
| | | include unlabeled | 39.56% ❶ | 1.0153 ❶ | 0.2378 ❶ | 0.0972 ❶ | 0.6628 ❶ |
| | | +BST, include unlabeled | **39.61%** ❶ | **1.0051** ❶ | **0.2354** ❶ | **0.0948** ❶ | **0.6722** ❶ |
| | DPD-ΠM | exclude unlabeled | 34.93% | 1.1307 | 0.2649 | 0.1097 | 0.6344 |
| | | include unlabeled | 37.16% ❶ | 1.0819 ❶ | 0.2534 ❶ | 0.1027 ❶ | 0.6469 ❶ |
| | | +BST, include unlabeled | **39.93%** ❶ | **0.9997** ❶ | **0.2342** ❶ | **0.0959** ❶ | **0.6747** ❶ |
| GRU | SUPV | exclude/include unlabeled | 27.42% | 1.3288 | 0.3112 | 0.1238 | 0.5834 |
| | | +BST, include unlabeled | **29.44%** | **1.2600** ❶ | **0.2951** ❶ | **0.1167** | **0.6071** ❶ |
| | ΠM | exclude unlabeled | 29.46% | 1.2475 | 0.2922 | 0.1157 | 0.6124 |
| | | include unlabeled | 31.32% | 1.2195 | 0.2856 | 0.1129 | 0.6145 |
| | | +BST, include unlabeled | **35.46%** ❶ | **1.1168** ❶ | **0.2616** ❶ | **0.1026** ❶ | **0.6452** ❶ |
| | DPD | exclude unlabeled | 26.57% | 1.3424 | 0.3144 | 0.1259 | 0.5830 |
| | | include unlabeled | 33.62% ❶ | 1.1666 ❶ | 0.2733 ❶ | 0.1102 ❶ | 0.6265 ❶ |
| | | +BST, include unlabeled | **35.59%** ❶ | **1.1035** ❶ | **0.2585** ❶ | **0.1031** ❶ | **0.6499** ❶ |
| | DPD-ΠM | exclude unlabeled | 30.27% | 1.2393 | 0.2903 | 0.1156 | 0.6067 |
| | | include unlabeled | 36.19% ❶ | 1.0992 ❶ | 0.2575 ❶ | 0.1007 ❶ | 0.6440 ❶ |
| | | +BST, include unlabeled | **39.92%** ❶ | **1.0093** ❶ | **0.2364** ❶ | **0.0952** ❶ | **0.6735** ❶ |

| Architecture | Base Strategy | Configuration | ACC% ↑ | TED ↓ | TER ↓ | FER ↓ | BCFS ↑ |
|---|---|---|---|---|---|---|---|
| Transformer | SUPV | exclude/include unlabeled | 27.66% | 1.6753 | 0.1869 | 0.0758 | 0.7303 |
| | | +BST, include unlabeled | **30.56%** ❶ | **1.4712** ❶ | **0.1641** ❶ | **0.0681** ❶ | **0.7610** ❶ |
| | ΠM | exclude unlabeled | 30.01% | 1.5261 | 0.1702 | 0.0699 | 0.7536 |
| | | include unlabeled | 27.69% | 1.6483 | 0.1838 | 0.0751 | 0.7332 |
| | | +BST, include unlabeled | **31.90%** ❶ | **1.3935** ❶ | **0.1554** ❶ | **0.0636** ❶ | **0.7740** ❶ |
| | DPD | exclude unlabeled | 28.22% | 1.6713 | 0.1864 | 0.0745 | 0.7308 |
| | | include unlabeled | 31.81% ❶ | 1.5031 ❶ | 0.1677 ❶ | 0.0668 ❶ | 0.7543 ❶ |
| | | +BST, include unlabeled | **33.96%** ❶ | **1.3332** ❶ | **0.1487** ❶ | **0.0591** ❶ | **0.7812** ❶ |
| | DPD-ΠM | exclude unlabeled | 27.42% | 1.5860 | 0.1769 | 0.0741 | 0.7465 |
| | | include unlabeled | 30.09% ❶ | 1.5505 | 0.1729 | 0.0708 ❶ | 0.7482 |
| | | +BST, include unlabeled | **34.33%** ❶ | **1.3121** ❶ | **0.1463** ❶ | **0.0592** ❶ | **0.7856** ❶ |
| GRU | SUPV | exclude/include unlabeled | 30.01% | 1.5156 | 0.1690 | 0.0607 | 0.7547 |
| | | +BST, include unlabeled | **35.21%** ❶ | **1.3284** ❶ | **0.1482** ❶ | **0.0530** ❶ | **0.7857** ❶ |
| | ΠM | exclude unlabeled | 32.38% | 1.4232 | 0.1587 | 0.0591 | 0.7718 |
| | | include unlabeled | 29.42% | 1.5511 | 0.1730 | 0.0643 | 0.7529 |
| | | +BST, include unlabeled | **35.38%** ❶ | **1.3072** ❶ | **0.1458** ❶ | **0.0537** ❶ | **0.7896** ❶ |
| | DPD | exclude unlabeled | 30.68% | 1.5010 | 0.1674 | 0.0629 | 0.7588 |
| | | include unlabeled | 30.74% | 1.4816 | 0.1652 | 0.0591 ❶ | 0.7589 |
| | | +BST, include unlabeled | **36.45%** ❶ | **1.2469** ❶ | **0.1391** ❶ | **0.0473** ❶ | **0.7977** ❶ |
| | DPD-ΠM | exclude unlabeled | 32.35% | 1.4294 | 0.1594 | 0.0596 | 0.7702 |
| | | include unlabeled | 31.81% | 1.4930 | 0.1665 | 0.0628 | 0.7583 |
| | | +BST, include unlabeled | **36.42%** ❶ | **1.2529** ❶ | **0.1397** ❶ | **0.0493** ❶ | **0.7957** ❶ |

Table 14: Performance comparison between whether to exclude unlabeled data, include unlabeled data, or include and pseudo-label unlabeled data (+BST, include unlabelled), evaluated on 10% labeled group 1 WikiHan (top) and 10% labeled group 1 Rom-phon (bottom), averaged across 10 runs. Bold: best performance for the corresponding base strategy; ❶: significantly better than when unlabeled data are not used ($p < 0.01$). For SUPV without BST, including and excluding unlabeled data are equivalent.

| Dataset | Reconstruction System | ACC% ↑ | TED ↓ | TER ↓ | FER ↓ | BCFS ↑ |
|---|---|---|---|---|---|---|
| WikiHan | GRU-SUPV (Meloni et al., 2021) | 55.58% | 0.7360 | 0.1724 | 0.0686 | 0.7426 |
| | Trans-SUPV (Kim et al., 2023) | 54.62% | 0.7453 | 0.1746 | 0.0696 | 0.7393 |
| | GRU-BS + GRU Reranker (Lu et al., 2024) | 57.14% ○ | 0.7045 ○ | 0.1650 ○ | 0.0661 ○ | 0.7515 ○ |
| | GRU-BS + Trans. Reranker (Lu et al., 2024) | 57.26% ○ | 0.7029 ○ | 0.1646 ○ | 0.0658 ○ | 0.7520 ○ |
| | GRU-ΠM (Laine and Aila, 2017) | 57.37% ○ | 0.7109 ○ | 0.1665 ○ | 0.0646 ○ | 0.7505 ○ |
| | GRU-DPD (ours) | 55.22% | 0.7405 | 0.1734 | 0.0680 | 0.7410 |
| | GRU-DPD-ΠM (ours) | 56.63% ○ | 0.7206 | 0.1688 | 0.0645 ○ | 0.7469 |
| | Trans-ΠM (Laine and Aila, 2017) | 56.89% ○ | 0.7144 ○ | 0.1673 ○ | 0.0661 ○ | 0.7487 ○ |
| | Trans-DPD (ours) | 56.36% | 0.7183 ○ | 0.1683 ○ | 0.0670 | 0.7479 ○ |
| | Trans-DPD-ΠM (ours) | **57.63%** ○ | **0.6967** ○ | **0.1632** ○ | **0.0631** ● | **0.7544** ○ |
| Rom-phon | GRU-SUPV (Meloni et al., 2021) | 51.92% | 0.9775 | 0.1244 | 0.0390 | 0.8275 |
| | Trans-SUPV (Kim et al., 2023) | 53.04% | 0.9050 | 0.1148 | 0.0377 | 0.8417 |
| | GRU-BS + GRU Reranker (Lu et al., 2024) | 53.95% ○ | 0.8775 ○ | 0.0979 ○ | 0.0336 ○ | 0.8460 ○ |
| | GRU-BS + Trans. Reranker (Lu et al., 2024) | 53.85% ○ | 0.8765 ○ | 0.0978 ○ | **0.0333** ○ | 0.8461 ○ |
| | GRU-ΠM (Laine and Aila, 2017) | 54.43% ○ | 0.9226 | 0.1029 ○ | 0.0388 | 0.8390 |
| | GRU-DPD (ours) | **56.20%** ● | **0.8658** ○ | **0.0966** ○ | 0.0350 ○ | **0.8462** ○ |
| | GRU-DPD-ΠM (ours) | 55.68% ● | 0.8810 ○ | 0.0983 ○ | 0.0371 | 0.8453 ○ |
| | Trans-ΠM (Laine and Aila, 2017) | 54.95% ○ | 0.8864 ○ | 0.0989 ○ | 0.0379 | 0.8450 |
| | Trans-DPD (ours) | 53.44% | 0.9318 | 0.1039 ○ | 0.0391 | 0.8367 |
| | Trans-DPD-ΠM (ours) | 53.25% | 0.8921 | 0.0995 ○ | 0.0376 | 0.8459 ○ |

Table 15: Supervised reconstruction (100% labeling setting) performance of baseline methods and semisupervised strategies, averaged across 20 or 10 runs (see Section J for detail). Bold: best-performing reconstruction system for the corresponding dataset; ○: significantly better than both GRU-SUPV (Meloni et al., 2021) and Trans-SUPV (Kim et al., 2023) ($p < 0.01$); ●: significantly better than all of GRU-SUPV (Meloni et al., 2021), Trans-SUPV (Kim et al., 2023), GRU-BS + GRU Reranker (Lu et al., 2024), and GRU-BS + Trans. Reranker (Lu et al., 2024) ($p < 0.01$).

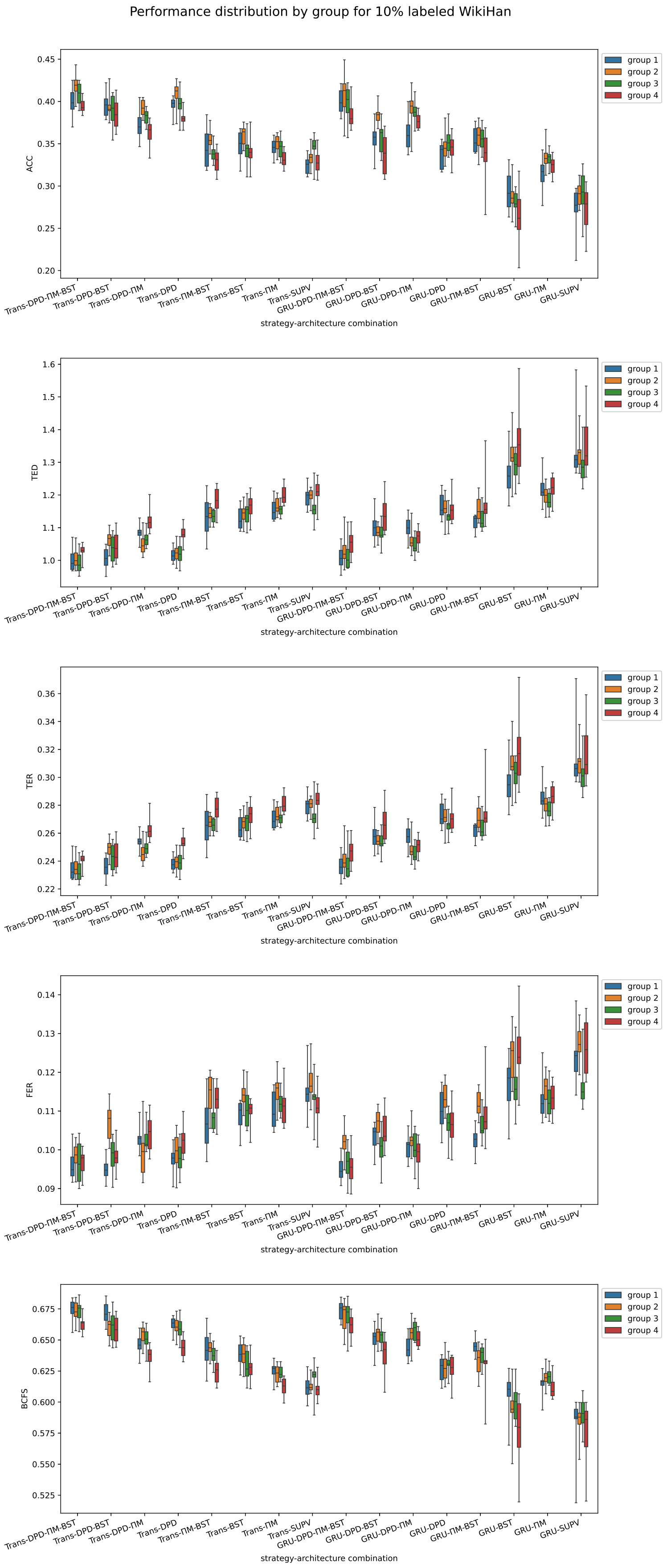

Figure 6: Box plots showing performance distribution for each metric and for each group on 10% labeled WikiHan.

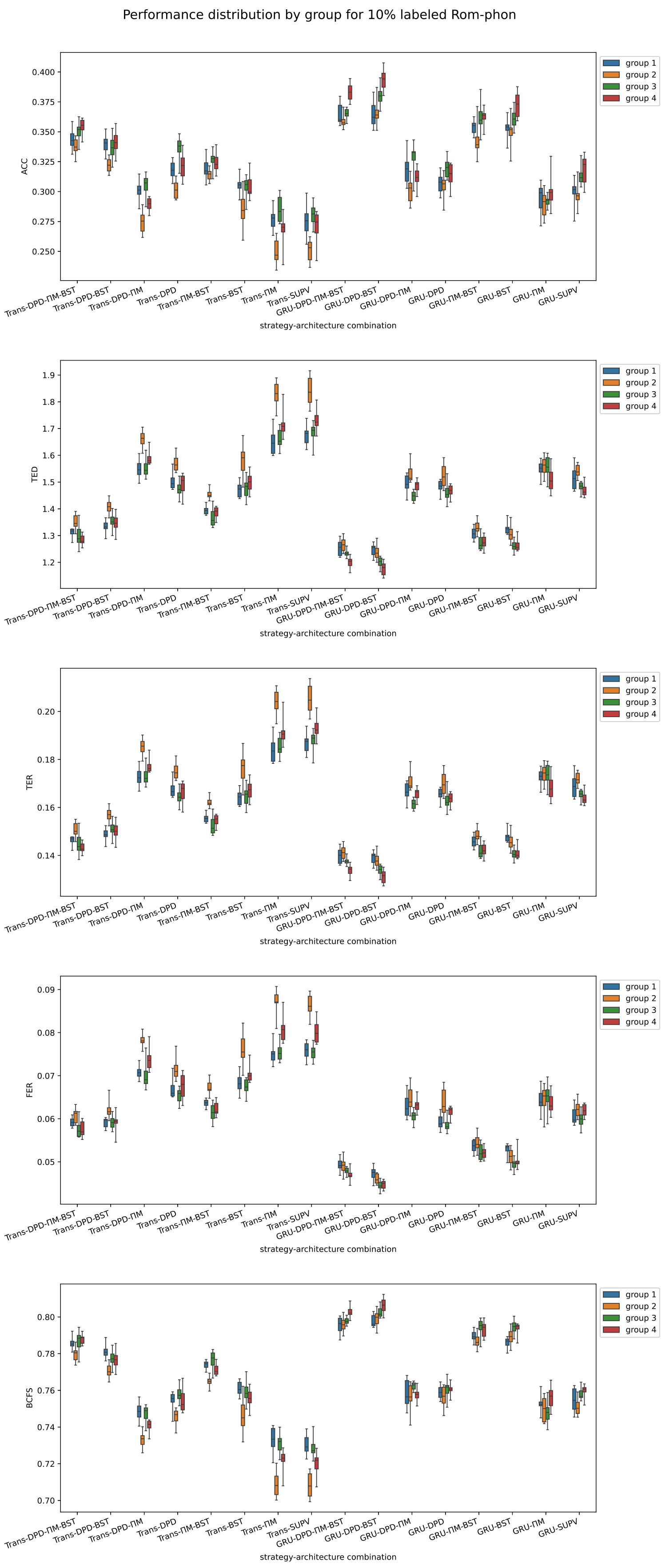

Figure 7: Box plots showing performance distribution for each metric and for each group on 10% labeled Rom-phon.

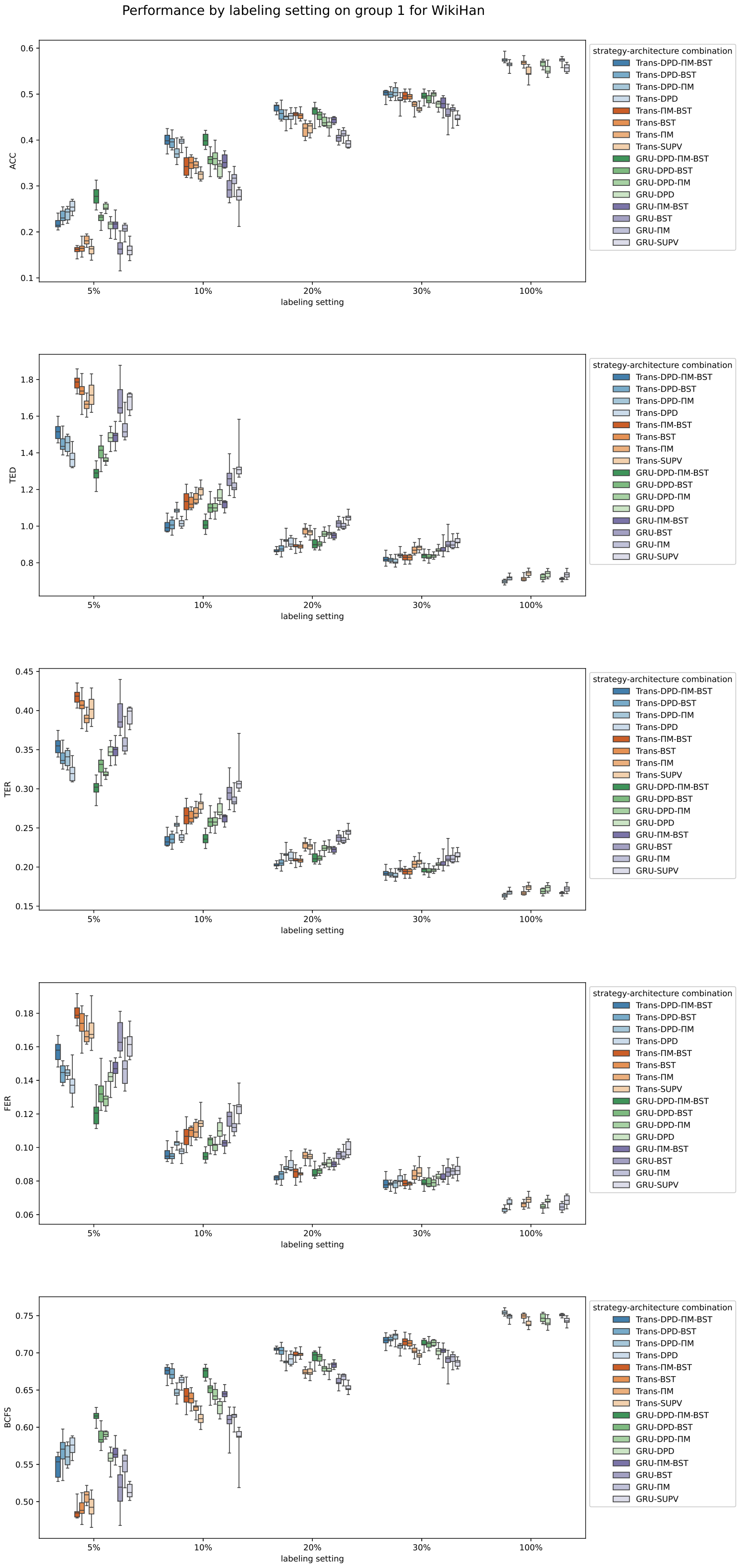

Figure 8: Box plots showing performance distribution for each metric given varied percentages of labels on WikiHan group 1.

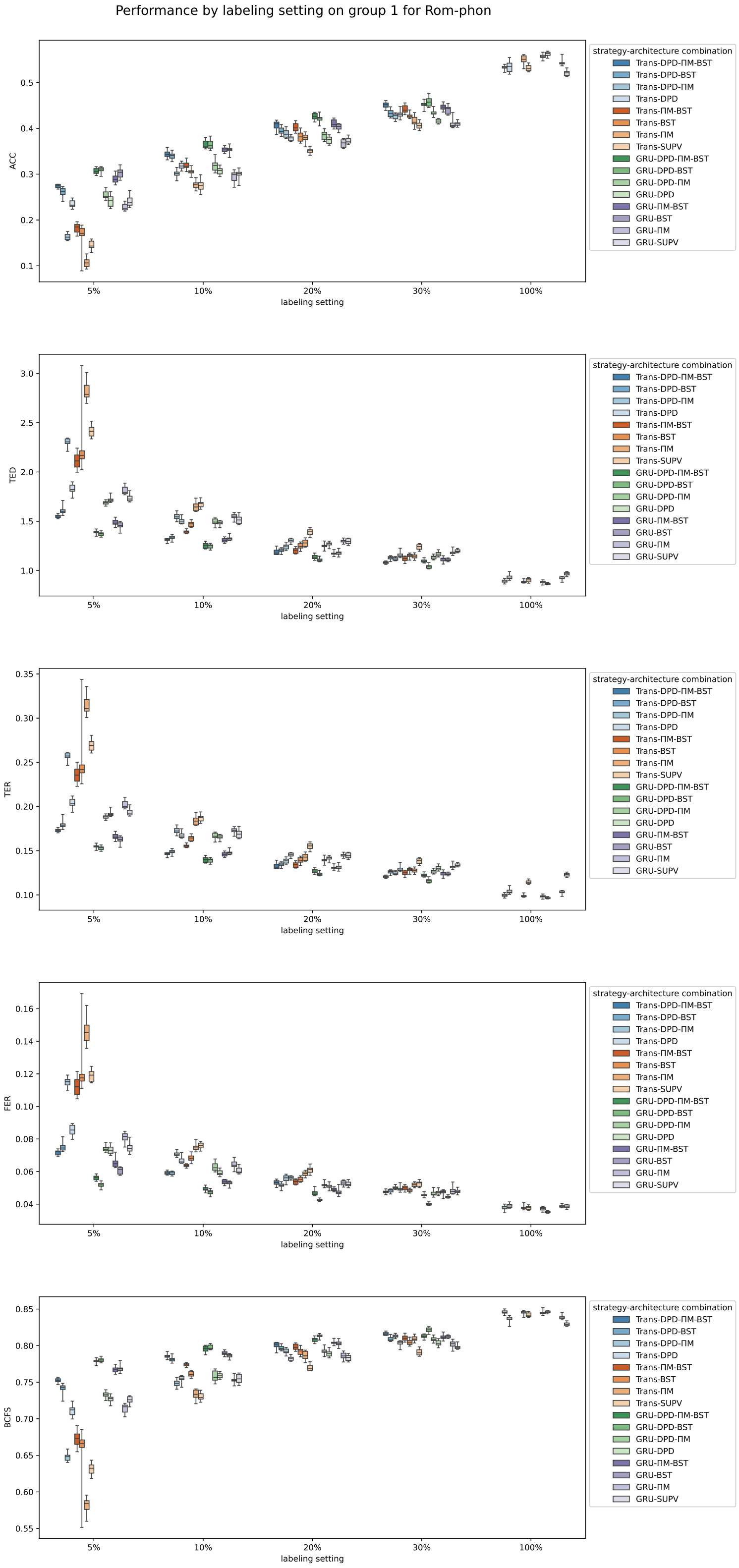

Figure 9: Box plots showing performance distribution for each metric given varied percentages of labels on Rom-phon group 1.

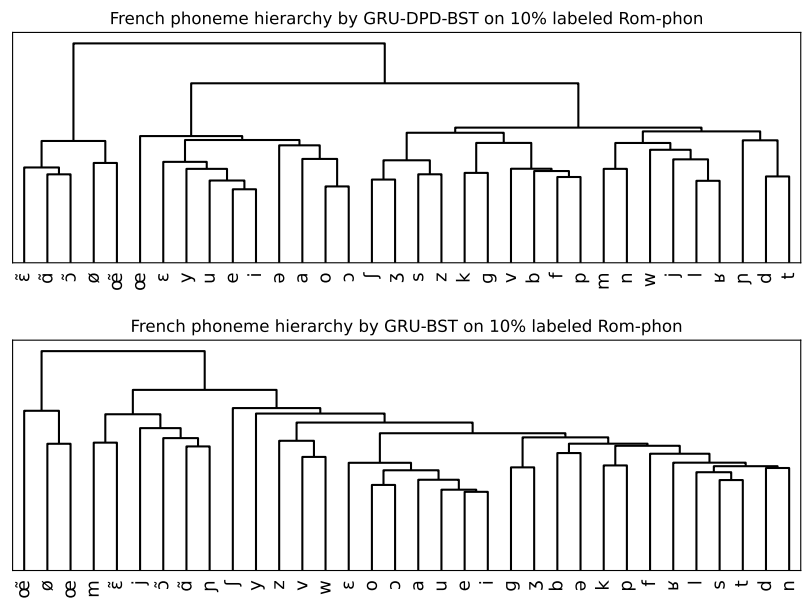

(a) French

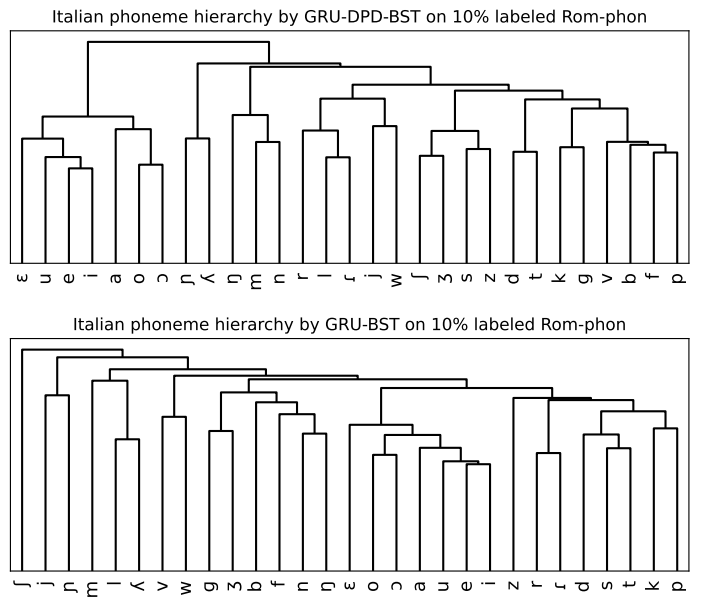

(b) Italian

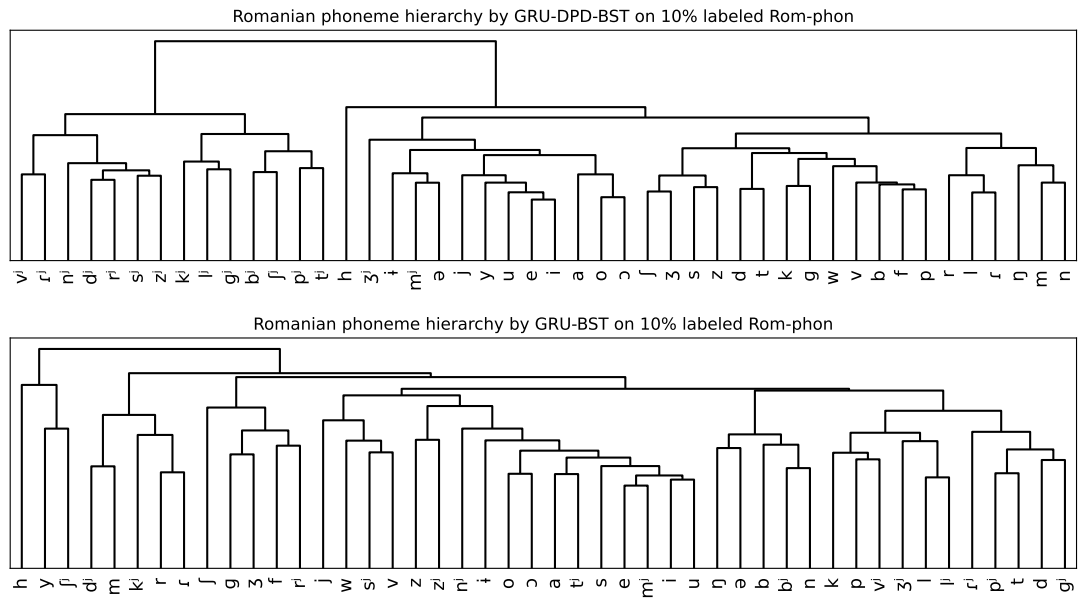

(c) Romanian

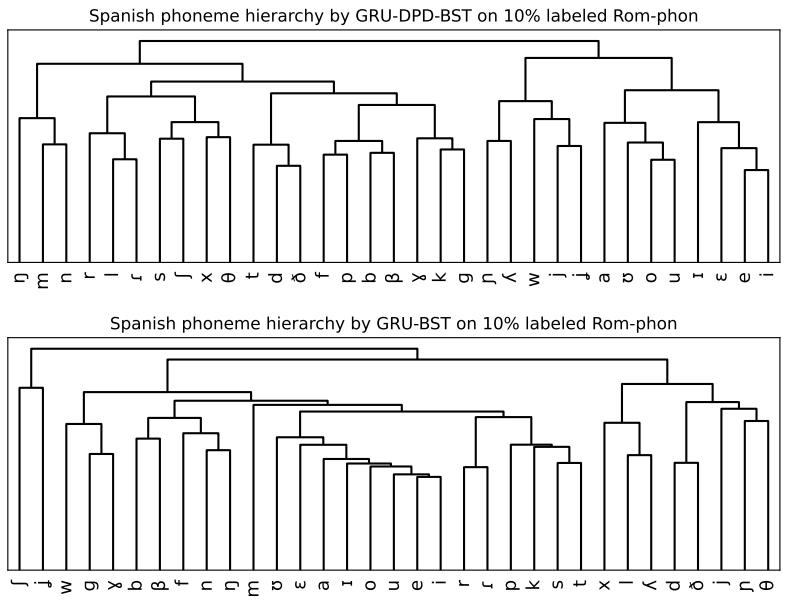

(d) Spanish

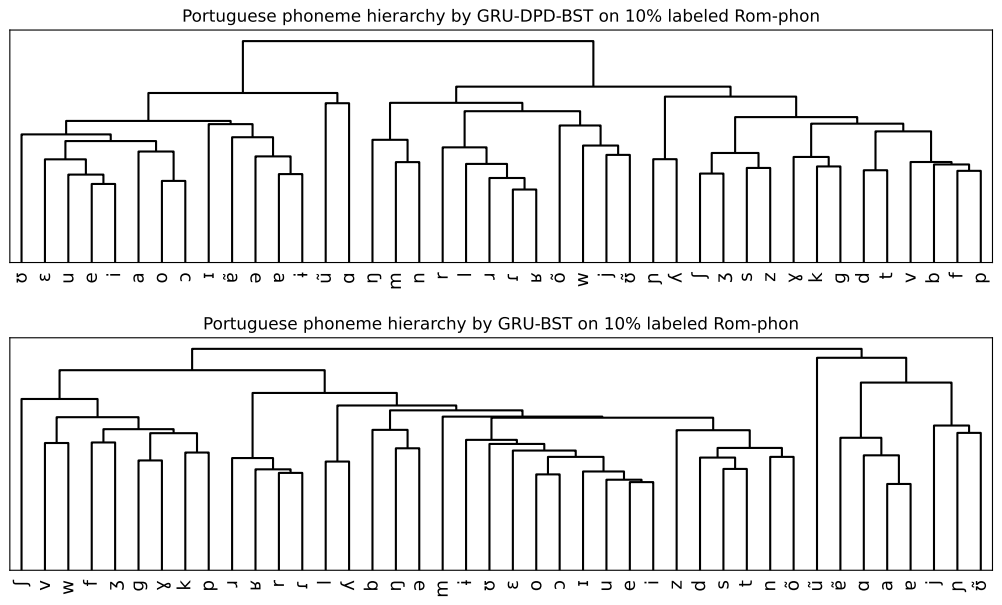

(e) Portuguese

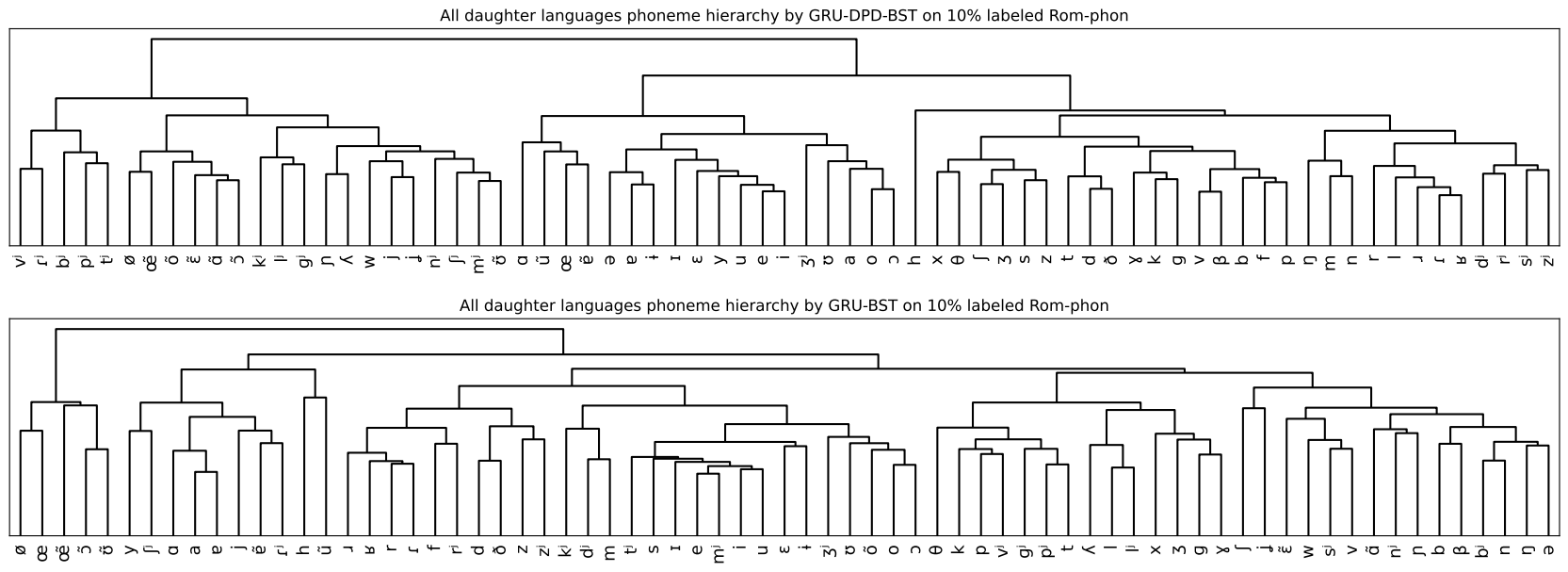

(f) All daughters

Figure 10: Hierarchical clustering revealing phoneme organization learned by the best run in the best DPD-based strategy-architecture combination (within group 1 and on 10% labeled Rom-phon) (top) and the best run from their non-DPD counterpart (bottom). Note that a comparison for Latin is not possible since non-DPD strategies do not learn embeddings for Latin phonemes.

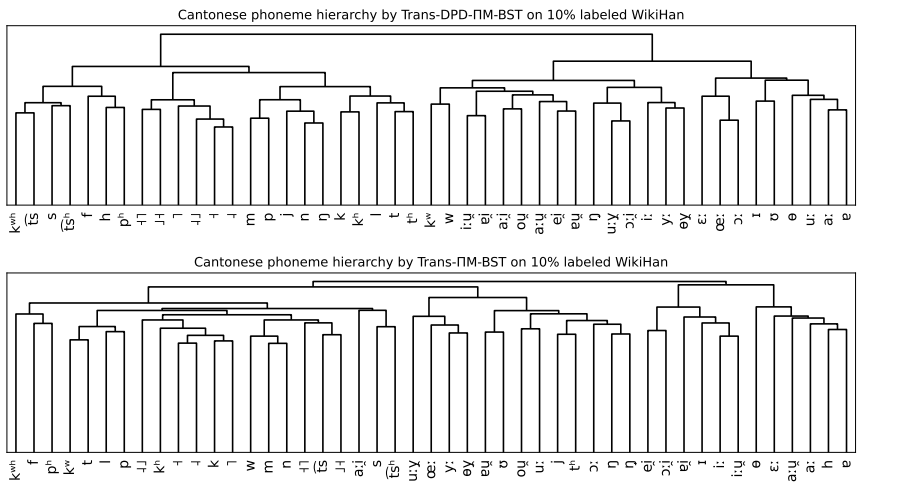

(a) Cantonese

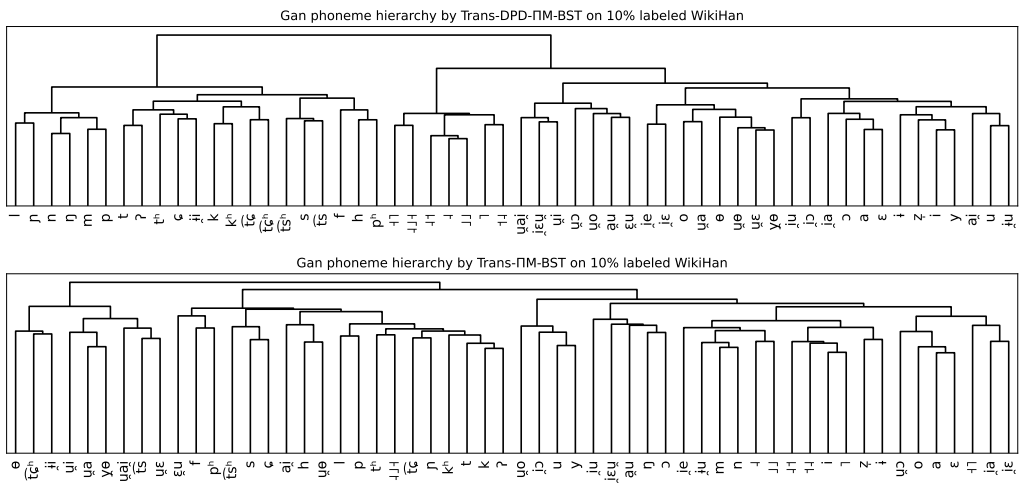

(b) Gan

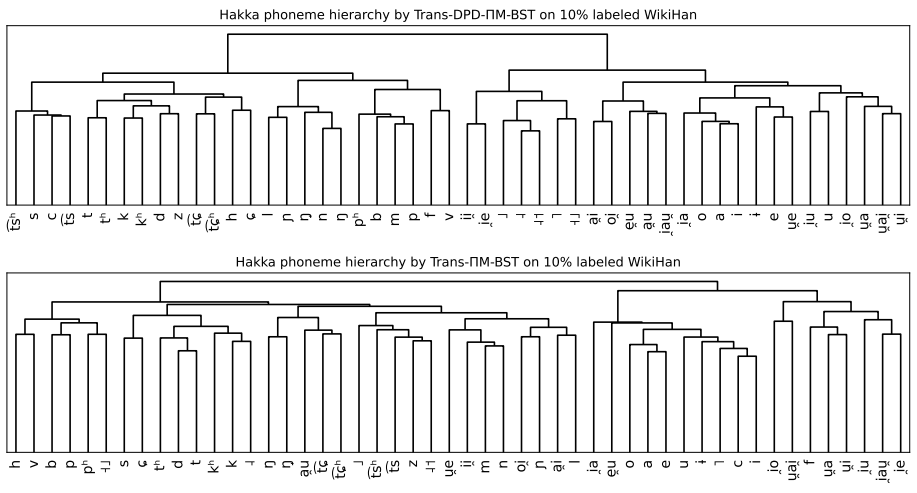

(c) Hakka

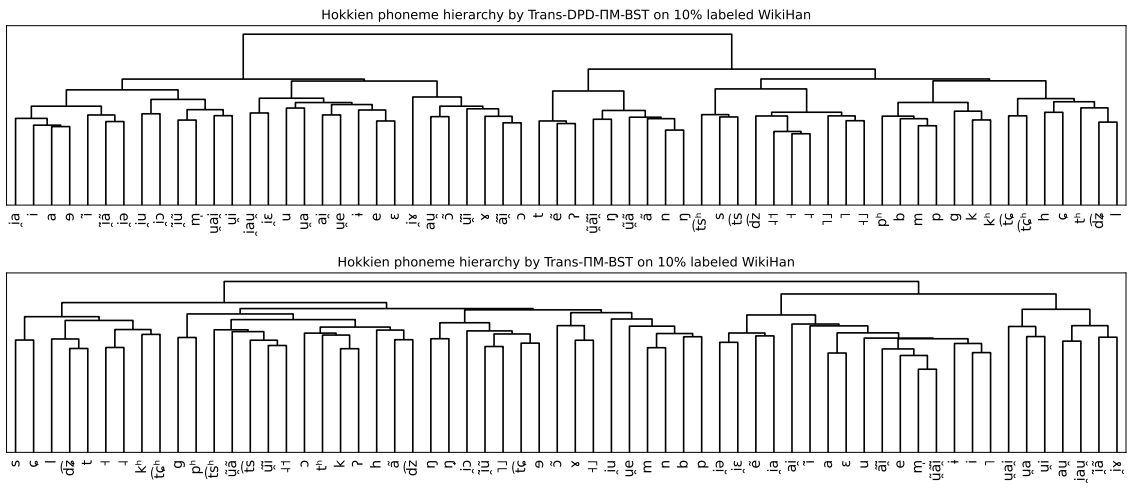

(d) Hokkien

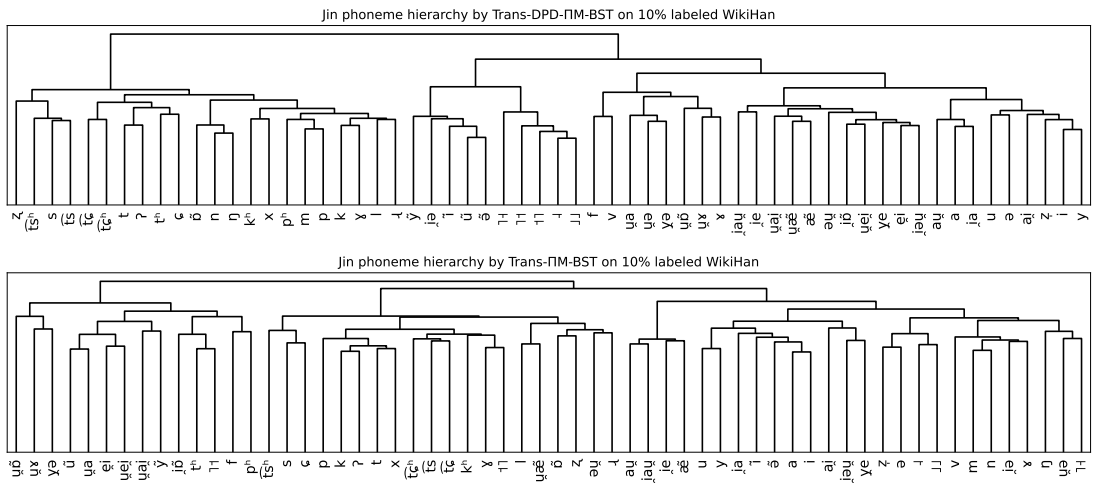

(e) Jin

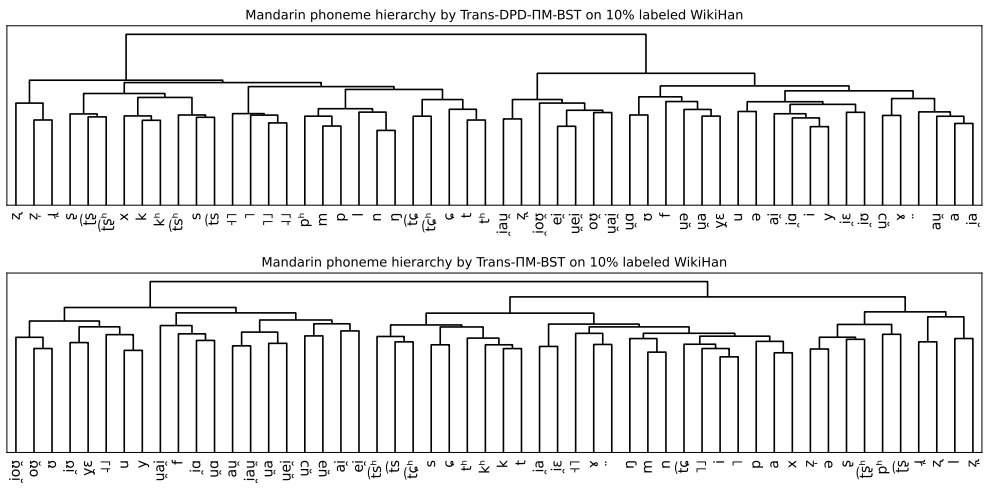

(f) Mandarin

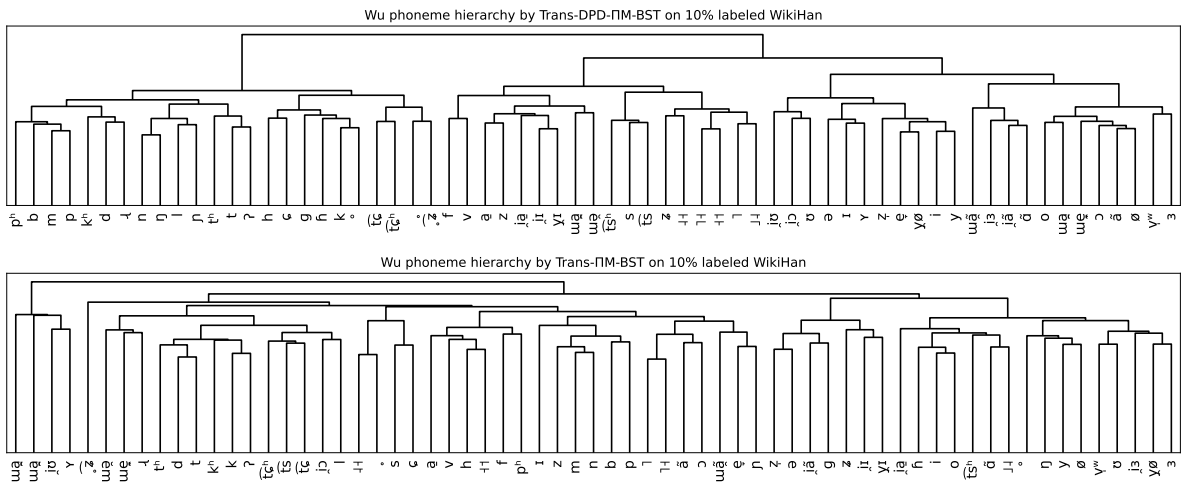

(g) Wu

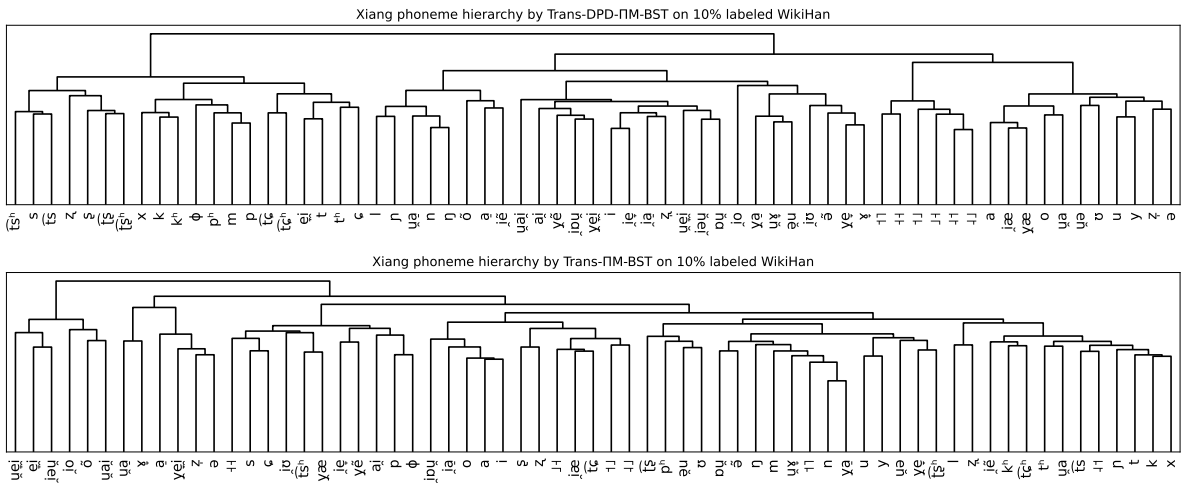

(h) Xiang

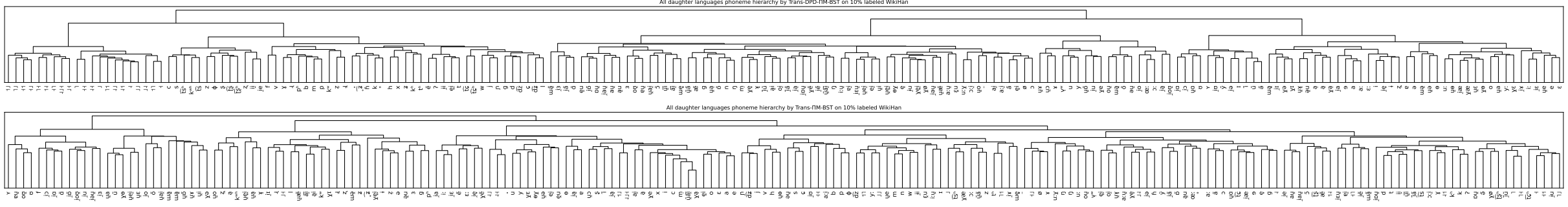

(i) All daughters

Figure 11: Hierarchical clustering revealing phoneme organization learned by the best run in the best DPD-based strategy-architecture combination (within group 1 and on 10% labeled WikiHan) (top) and the best run from their non-DPD counterpart (bottom). Note that a comparison for Middle Chinese is not possible since non-DPD strategies do not learn embeddings for Middle Chinese phonemes.

| | GRU-SUPV | GRU-ΠM | GRU-BST | GRU-ΠM-BST | GRU-DPD | GRU-DPD-ΠM | GRU-DPD-BST | GRU-DPD-ΠM-BST | Trans-SUPV | Trans-ΠM | Trans-BST | Trans-ΠM-BST | Trans-DPD | Trans-DPD-ΠM | Trans-DPD-BST | Trans-DPD-ΠM-BST |
|---|---|---|---|---|---|---|---|---|---|---|---|---|---|---|---|---|
| batch size | 128 | 64 | 128 | 64 | 128 | 128 | 64 | 64 | 128 | 128 | 256 | 256 | 64 | 256 | 64 | 64 |
| max epochs | 221 | 247 | 238 | 253 | 345 | 268 | 384 | 253 | 205 | 288 | 374 | 341 | 261 | 256 | 390 | 259 |
| warmup epochs (learning rate) | 29 | 12 | 22 | 16 | 37 | 10 | 2 | 27 | 26 | 6 | 17 | 26 | 18 | 7 | 31 | 22 |
| $\beta_1$ (Adam) | 0.9000 | 0.9000 | 0.9000 | 0.9000 | 0.9000 | 0.9000 | 0.9000 | 0.9000 | 0.9000 | 0.9000 | 0.9000 | 0.9000 | 0.9000 | 0.9000 | 0.9000 | 0.9000 |
| $\beta_2$ (Adam) | 0.9990 | 0.9990 | 0.9990 | 0.9990 | 0.9990 | 0.9990 | 0.9990 | 0.9990 | 0.9990 | 0.9990 | 0.9990 | 0.9990 | 0.9990 | 0.9990 | 0.9990 | 0.9990 |
| $\epsilon$ (Adam) | 1.000e-08 | 1.000e-08 | 1.000e-08 | 1.000e-08 | 1.000e-08 | 1.000e-08 | 1.000e-08 | 1.000e-08 | 1.000e-08 | 1.000e-08 | 1.000e-08 | 1.000e-08 | 1.000e-08 | 1.000e-08 | 1.000e-08 | 1.000e-08 |
| learning rate | 0.0009315 | 0.0007376 | 0.0008285 | 0.0005646 | 0.0007678 | 0.0009974 | 0.0008380 | 0.0008118 | 0.0007602 | 0.0005026 | 0.0009074 | 0.0008765 | 0.0007032 | 0.0009268 | 0.0006642 | 0.0008475 |
| weight decay (Adam) | 0.000 | 0.000 | 0.000 | 0.000 | 0.000 | 0.000 | 0.000 | 0.000 | 1.000e-07 | 0.000 | 1.000e-07 | 1.000e-07 | 1.000e-07 | 0.000 | 0.000 | 0.000 |
| DPD protoform reconstruction weight | | | | | 1.100 | 0.7723 | 0.7696 | 0.5396 | | | | | 0.6270 | 0.5239 | 1.338 | 0.6952 |
| DPD bridge network weight | | | | | 0.3119 | 0.06223 | 0.6173 | 0.3858 | | | | | 0.5003 | 0.5839 | 0.1296 | 0.2182 |
| DPD reflex prediction from gold protoform weight | | | | | 0.04734 | 0.5011 | 0.09115 | 0.6411 | | | | | 0.6265 | 0.4270 | 0.4811 | 0.2464 |
| DPD reflex prediction from reconstruction weight | | | | | 1.498 | 1.249 | 0.9654 | 1.468 | | | | | 1.196 | 1.475 | 0.8612 | 1.322 |
| DPD CRINGE loss weight $\alpha$ | | | | | 0.7015 | 0.3820 | 0.5720 | 0.3732 | | | | | 0.3769 | 0.2332 | 0.6782 | 0.3056 |
| DPD CRINGE loss top $k$ | | | | | 3 | 4 | 1 | 3 | | | | | 1 | 4 | 3 | 3 |
| DPD shared embedding size | | | | | 128 | 256 | 128 | 256 | | | | | 256 | 128 | 256 | 256 |
| D2P encoder layers count | 2 | 2 | 2 | 2 | 2 | 2 | 2 | 2 | 2 | 2 | 2 | 2 | 2 | 2 | 2 | 2 |
| D2P dropout | 0.3117 | 0.2189 | 0.2336 | 0.3137 | 0.3228 | 0.2997 | 0.1999 | 0.2471 | 0.2539 | 0.1951 | 0.1949 | 0.1877 | 0.1594 | 0.2649 | 0.1574 | 0.1692 |
| D2P inference decode max length | 15 | 15 | 15 | 15 | 15 | 15 | 15 | 15 | 15 | 15 | 15 | 15 | 15 | 15 | 15 | 15 |
| D2P feedforward dimension | 512 | 512 | 512 | 512 | 512 | 512 | 512 | 512 | 512 | 512 | 512 | 512 | 512 | 512 | 512 | 512 |
| D2P embedding size | 384 | 128 | 128 | 128 | | | | | 256 | 384 | 256 | 128 | | | | |
| D2P model size | 128 | 64 | 128 | 128 | 64 | 64 | 64 | 64 | | | | | | | | |
| D2P encoder layers count | | | | | | | | | 2 | 2 | 2 | 2 | 2 | 2 | 2 | 2 |
| D2P number of heads | | | | | | | | | 8 | 8 | 8 | 8 | 8 | 8 | 8 | 8 |
| D2P max input length | | | | | | | | | 128 | 128 | 128 | 128 | 128 | 128 | 128 | 128 |
| P2D encoder layers count | | | | | 2 | 2 | 2 | 2 | | | | | 2 | 2 | 2 | 2 |
| P2D dropout | | | | | 0.1728 | 0.3136 | 0.3145 | 0.1546 | | | | | 0.3374 | 0.2532 | 0.1752 | 0.3339 |
| P2D inference decode max length | | | | | 15 | 15 | 15 | 15 | | | | | 15 | 15 | 15 | 15 |
| P2D feedforward dimension | | | | | 512 | 512 | 512 | 512 | | | | | 512 | 512 | 512 | 512 |
| P2D embedding size | | | | | | | | | | | | | | | | |
| P2D model size | | | | | 128 | 64 | 128 | 128 | | | | | | | | |
| P2D encoder layers count | | | | | | | | | | | | | 2 | 2 | 2 | 2 |
| P2D number of heads | | | | | | | | | | | | | 8 | 8 | 8 | 8 |
| P2D max input length | | | | | | | | | | | | | 128 | 128 | 128 | 128 |
| Bootstrapping starting epoch | | | 38 | 7 | | | 29 | 24 | | | 16 | 4 | | | 25 | 11 |
| Bootstrapping log probability threshold | | | -0.006900 | -0.008731 | | | -0.007431 | -0.005348 | | | -0.002607 | -0.009312 | | | -0.003975 | -0.007042 |
| Bootstrapping max new pseudo-labels per epoch | | | 75 | 87 | | | 41 | 49 | | | 54 | 79 | | | 86 | 35 |
| Π-model consistency ramp-up epochs | | 1 | | 11 | | 8 | | 23 | | 24 | | 8 | | 25 | | 27 |
| Π-model max consistency scaling | | 382.0 | | 222.1 | | 275.1 | | 181.7 | | 223.3 | | 184.5 | | 207.5 | | 88.70 |

Table 16: Hyperparameters for semisupervised reconstruction experiments on WikiHan.

| | GRU-SUPV | GRU-ΠM | GRU-BST | GRU-ΠM-BST | GRU-DPD | GRU-DPD-ΠM | GRU-DPD-BST | GRU-DPD-ΠM-BST | Trans-SUPV | Trans-ΠM | Trans-BST | Trans-ΠM-BST | Trans-DPD | Trans-DPD-ΠM | Trans-DPD-BST | Trans-DPD-ΠM-BST |
|---|---|---|---|---|---|---|---|---|---|---|---|---|---|---|---|---|
| batch size | 256 | 256 | 64 | 64 | 256 | 256 | 128 | 128 | 256 | 128 | 256 | 128 | 256 | 256 | 128 | 128 |
| max epochs | 316 | 283 | 289 | 383 | 224 | 334 | 371 | 373 | 219 | 266 | 382 | 344 | 305 | 206 | 257 | 216 |
| warmup epochs (learning rate) | 39 | 15 | 20 | 29 | 34 | 29 | 26 | 13 | 33 | 4 | 36 | 37 | 37 | 29 | 10 | 32 |
| $\beta_1$ (Adam) | 0.9000 | 0.9000 | 0.9000 | 0.9000 | 0.9000 | 0.9000 | 0.9000 | 0.9000 | 0.9000 | 0.9000 | 0.9000 | 0.9000 | 0.9000 | 0.9000 | 0.9000 | 0.9000 |
| $\beta_2$ (Adam) | 0.9990 | 0.9990 | 0.9990 | 0.9990 | 0.9990 | 0.9990 | 0.9990 | 0.9990 | 0.9990 | 0.9990 | 0.9990 | 0.9990 | 0.9990 | 0.9990 | 0.9990 | 0.9990 |
| $\varepsilon$ (Adam) | 1.000e-08 | 1.000e-08 | 1.000e-08 | 1.000e-08 | 1.000e-08 | 1.000e-08 | 1.000e-08 | 1.000e-08 | 1.000e-08 | 1.000e-08 | 1.000e-08 | 1.000e-08 | 1.000e-08 | 1.000e-08 | 1.000e-08 | 1.000e-08 |
| learning rate | 0.0006089 | 0.0005475 | 0.0007060 | 0.0009183 | 0.0006284 | 0.0005706 | 0.0008698 | 0.0006257 | 0.0007210 | 0.0005128 | 0.0009196 | 0.0006014 | 0.0005052 | 0.0006181 | 0.0005023 | 0.0008986 |
| weight decay (Adam) | 0.000 | 0.000 | 0.000 | 0.000 | 0.000 | 0.000 | 0.000 | 0.000 | 1.000e-07 | 0.000 | 0.000 | 1.000e-07 | 0.000 | 1.000e-07 | 0.000 | 1.000e-07 |
| DPD protoform reconstruction weight | | | | | 1.358 | 0.5163 | 0.6345 | 0.8579 | | | | | 0.6164 | 1.033 | 0.5132 | 0.5674 |
| DPD bridge network weight | | | | | 0.4293 | 0.3316 | 0.5736 | 0.5268 | | | | | 0.7455 | 0.4612 | 0.6321 | 0.5607 |
| DPD reflex prediction from gold protoform weight | | | | | 0.6922 | 0.6284 | 0.1439 | 0.4346 | | | | | 0.2108 | 0.5989 | 0.3991 | 0.3401 |
| DPD reflex prediction from reconstruction weight | | | | | 1.317 | 0.9957 | 1.476 | 0.9865 | | | | | 1.201 | 0.8703 | 1.294 | 1.421 |
| DPD CRINGE loss weight $\alpha$ | | | | | 0.02017 | 0.002749 | 0.5966 | 0.5339 | | | | | 0.03967 | 0.3295 | 0.6362 | 0.02023 |
| DPD CRINGE loss top $k$ | | | | | 5 | 1 | 5 | 5 | | | | | 5 | 2 | 4 | 1 |
| DPD shared embedding size | | | | | 256 | 384 | 384 | 384 | | | | | 384 | 384 | 384 | 256 |
| D2P encoder layers count | 2 | 2 | 2 | 2 | 2 | 2 | 2 | 2 | 2 | 2 | 2 | 2 | 2 | 2 | 2 | 2 |
| D2P dropout | 0.3342 | 0.2129 | 0.2680 | 0.2261 | 0.3300 | 0.1774 | 0.1970 | 0.2402 | 0.2838 | 0.3484 | 0.1963 | 0.2587 | 0.2667 | 0.3453 | 0.2627 | 0.2551 |
| D2P inference decode max length | 30 | 30 | 30 | 30 | 30 | 30 | 30 | 30 | 30 | 30 | 30 | 30 | 30 | 30 | 30 | 30 |
| D2P feedforward dimension | 512 | 512 | 512 | 512 | 512 | 512 | 512 | 512 | 512 | 512 | 512 | 512 | 512 | 512 | 512 | 512 |
| D2P embedding size | 384 | 128 | 256 | 128 | | | | | 256 | 128 | 384 | 384 | | | | |
| D2P model size | 128 | 128 | 64 | 64 | 64 | 64 | 128 | 128 | | | | | | | | |
| D2P encoder layers count | | | | | | | | | 2 | 2 | 2 | 2 | 2 | 2 | 2 | 2 |
| D2P number of heads | | | | | | | | | 8 | 8 | 8 | 8 | 8 | 8 | 8 | 8 |
| D2P max input length | | | | | | | | | 128 | 128 | 128 | 128 | 128 | 128 | 128 | 128 |
| P2D encoder layers count | | | | | 2 | 2 | 2 | 2 | | | | | 2 | 2 | 2 | 2 |
| P2D dropout | | | | | 0.2214 | 0.2429 | 0.3359 | 0.3346 | | | | | 0.3470 | 0.3168 | 0.2199 | 0.2197 |
| P2D inference decode max length | | | | | 30 | 30 | 30 | 30 | | | | | 30 | 30 | 30 | 30 |
| P2D feedforward dimension | | | | | 512 | 512 | 512 | 512 | | | | | 512 | 512 | 512 | 512 |
| P2D embedding size | | | | | | | | | | | | | | | | |
| P2D model size | | | | | 128 | 64 | 64 | 64 | | | | | | | | |
| P2D encoder layers count | | | | | | | | | | | | | 2 | 2 | 2 | 2 |
| P2D number of heads | | | | | | | | | | | | | 8 | 8 | 8 | 8 |
| P2D max input length | | | | | | | | | | | | | 128 | 128 | 128 | 128 |
| Bootstrapping starting epoch | | | 37 | 20 | | | 28 | 16 | | | 4 | 28 | | | 39 | 39 |
| Bootstrapping log probability threshold | | | -0.005144 | -0.009153 | | | -0.003401 | -0.005465 | | | -0.004997 | -0.003632 | | | -0.001733 | -0.007130 |
| Bootstrapping max new pseudo-labels per epoch | | | 51 | 95 | | | 86 | 37 | | | 48 | 69 | | | 73 | 46 |
| Π-model consistency ramp-up epochs | | 18 | | 14 | | 29 | | 12 | | 2 | | 1 | | 23 | | 18 |
| Π-model max consistency scaling | | 166.8 | | 168.4 | | 198.2 | | 211.6 | | 393.6 | | 248.6 | | 301.3 | | 192.2 |

Table 17: Hyperparameters for semisupervised reconstruction experiments on Rom-phon.

| | GRU-SUPV | GRU-ΠM | GRU-DPD | GRU-DPD-ΠM | Trans-SUPV | Trans-ΠM | Trans-DPD | Trans-DPD-ΠM |
|---|---|---|---|---|---|---|---|---|
| batch size | 128 | 128 | 64 | 256 | 128 | 64 | 64 | 64 |
| max epochs | 221 | 206 | 323 | 363 | 205 | 383 | 217 | 363 |
| warmup epochs (learning rate) | 29 | 17 | 5 | 11 | 26 | 26 | 35 | 37 |
| $\beta_1$ (Adam) | 0.9000 | 0.9000 | 0.9000 | 0.9000 | 0.9000 | 0.9000 | 0.9000 | 0.9000 |
| $\beta_2$ (Adam) | 0.9990 | 0.9990 | 0.9990 | 0.9990 | 0.9990 | 0.9990 | 0.9990 | 0.9990 |
| $\varepsilon$ (Adam) | 1.000e-08 | 1.000e-08 | 1.000e-08 | 1.000e-08 | 1.000e-08 | 1.000e-08 | 1.000e-08 | 1.000e-08 |
| learning rate | 0.0009315 | 0.0008794 | 0.0005477 | 0.0009955 | 0.0007602 | 0.0009179 | 0.0008318 | 0.0009033 |
| weight decay (Adam) | 0.000 | 0.000 | 0.000 | 0.000 | 1.000e-07 | 1.000e-07 | 0.000 | 1.000e-07 |
| DPD protoform reconstruction weight | | | 1.195 | 0.6016 | | | 0.6045 | 0.5390 |
| DPD bridge network weight | | | 0.01908 | 0.2512 | | | 0.2167 | 0.3800 |
| DPD reflex prediction from gold protoform weight | | | 0.4224 | 0.1871 | | | 0.1942 | 0.7438 |
| DPD reflex prediction from reconstruction weight | | | 1.140 | 0.7006 | | | 1.326 | 1.089 |
| DPD CRINGE loss weight $\alpha$ | | | 0.4218 | 0.03425 | | | 0.1233 | 0.6634 |
| DPD CRINGE loss top $k$ | | | 2 | 5 | | | 1 | 2 |
| DPD shared embedding size | | | 384 | 256 | | | 384 | 128 |
| D2P encoder layers count | 2 | 2 | 2 | 2 | 2 | 2 | 2 | 2 |
| D2P dropout | 0.3117 | 0.3494 | 0.2859 | 0.2805 | 0.2539 | 0.3113 | 0.2902 | 0.2546 |
| D2P inference decode max length | 15 | 15 | 15 | 15 | 15 | 15 | 15 | 15 |
| D2P feedforward dimension | 512 | 512 | 512 | 512 | 512 | 512 | 512 | 512 |
| D2P embedding size | 384 | 128 | | | 256 | 384 | | |
| D2P model size | 128 | 64 | 64 | 128 | | | | |
| D2P encoder layers count | | | | | 2 | 2 | 2 | 2 |
| D2P number of heads | | | | | 8 | 8 | 8 | 8 |
| D2P max input length | | | | | 128 | 128 | 128 | 128 |
| P2D encoder layers count | | | 2 | 2 | | | 2 | 2 |
| P2D dropout | | | 0.1934 | 0.2122 | | | 0.2663 | 0.2011 |
| P2D inference decode max length | | | 15 | 15 | | | 15 | 15 |
| P2D feedforward dimension | | | 512 | 512 | | | 512 | 512 |
| P2D embedding size | | | | | | | | |
| P2D model size | | | 64 | 128 | | | | |
| P2D encoder layers count | | | | | | | 2 | 2 |
| P2D number of heads | | | | | | | 8 | 8 |
| P2D max input length | | | | | | | 128 | 128 |
| Bootstrapping starting epoch | | | | | | | | |
| Bootstrapping log probability threshold | | | | | | | | |
| Bootstrapping max new pseudo-labels per epoch | | | | | | | | |
| Π-model consistency ramp-up epochs | | 11 | | 24 | | 28 | | 4 |
| Π-model max consistency scaling | | 352.9 | | 257.3 | | 202.9 | | 197.1 |

Table 18: Additional hyperparameters for 10% labeled WikiHan when unlabeled cognate sets are excluded.

| | GRU-SUPV | GRU-ΠM | GRU-DPD | GRU-DPD-ΠM | Trans-SUPV | Trans-ΠM | Trans-DPD | Trans-DPD-ΠM |
|---|---|---|---|---|---|---|---|---|
| batch size | 256 | 256 | 128 | 128 | 256 | 64 | 64 | 256 |
| max epochs | 316 | 210 | 357 | 373 | 219 | 355 | 221 | 310 |
| warmup epochs (learning rate) | 39 | 17 | 29 | 14 | 33 | 31 | 22 | 33 |
| $\beta_1$ (Adam) | 0.9000 | 0.9000 | 0.9000 | 0.9000 | 0.9000 | 0.9000 | 0.9000 | 0.9000 |
| $\beta_2$ (Adam) | 0.9990 | 0.9990 | 0.9990 | 0.9990 | 0.9990 | 0.9990 | 0.9990 | 0.9990 |
| $\varepsilon$ (Adam) | 1.000e-08 | 1.000e-08 | 1.000e-08 | 1.000e-08 | 1.000e-08 | 1.000e-08 | 1.000e-08 | 1.000e-08 |
| learning rate | 0.0006089 | 0.0009945 | 0.0006927 | 0.0005954 | 0.0007210 | 0.0007154 | 0.0007520 | 0.0009266 |
| weight decay (Adam) | 0.000 | 0.000 | 0.000 | 0.000 | 1.000e-07 | 0.000 | 0.000 | 0.000 |
| DPD protoform reconstruction weight | | | 0.9617 | 0.8749 | | | 0.6145 | 0.7346 |
| DPD bridge network weight | | | 0.07922 | 0.5155 | | | 0.2638 | 0.4678 |
| DPD reflex prediction from gold protoform weight | | | 0.2101 | 0.6491 | | | 0.5297 | 0.1037 |
| DPD reflex prediction from reconstruction weight | | | 1.248 | 0.6504 | | | 0.9362 | 1.036 |
| DPD CRINGE loss weight $\alpha$ | | | 0.6470 | 0.2729 | | | 0.6543 | 0.1272 |
| DPD CRINGE loss top $k$ | | | 1 | 2 | | | 3 | 4 |
| DPD shared embedding size | | | 384 | 384 | | | 384 | 384 |
| D2P encoder layers count | 2 | 2 | 2 | 2 | 2 | 2 | 2 | 2 |
| D2P dropout | 0.3342 | 0.2882 | 0.3309 | 0.2913 | 0.2838 | 0.3305 | 0.2617 | 0.3448 |
| D2P inference decode max length | 30 | 30 | 30 | 30 | 30 | 30 | 30 | 30 |
| D2P feedforward dimension | 512 | 512 | 512 | 512 | 512 | 512 | 512 | 512 |
| D2P embedding size | 384 | 128 | | | 256 | 384 | | |
| D2P model size | 128 | 128 | 64 | 64 | | | | |
| D2P encoder layers count | | | | | 2 | 2 | 2 | 2 |
| D2P number of heads | | | | | 8 | 8 | 8 | 8 |
| D2P max input length | | | | | 128 | 128 | 128 | 128 |
| P2D encoder layers count | | | 2 | 2 | | | 2 | 2 |
| P2D dropout | | | 0.3284 | 0.2994 | | | 0.3333 | 0.2712 |
| P2D inference decode max length | | | 30 | 30 | | | 30 | 30 |
| P2D feedforward dimension | | | 512 | 512 | | | 512 | 512 |
| P2D embedding size | | | | | | | | |
| P2D model size | | | 64 | 128 | | | | |
| P2D encoder layers count | | | | | | | 2 | 2 |
| P2D number of heads | | | | | | | 8 | 8 |
| P2D max input length | | | | | | | 128 | 128 |
| Bootstrapping starting epoch | | | | | | | | |
| Bootstrapping log probability threshold | | | | | | | | |
| Bootstrapping max new pseudo-labels per epoch | | | | | | | | |
| Π-model consistency ramp-up epochs | | 21 | | 25 | | 24 | | 6 |
| Π-model max consistency scaling | | 266.8 | | 145.5 | | 370.4 | | 334.1 |

Table 19: Additional hyperparameters for 10% labeled Rom-phon when unlabeled cognate sets are excluded.

| | GRU-ΠM | GRU-DPD | GRU-DPD-ΠM | Trans-ΠM | Trans-DPD | Trans-DPD-ΠM |
|---|---|---|---|---|---|---|
| batch size | 256 | 64 | 256 | 128 | 128 | 64 |
| max epochs | 283 | 269 | 275 | 382 | 232 | 315 |
| warmup epochs (learning rate) | 4 | 7 | 4 | 22 | 40 | 22 |
| $\beta_1$ (Adam) | 0.9000 | 0.9000 | 0.9000 | 0.9000 | 0.9000 | 0.9000 |
| $\beta_2$ (Adam) | 0.9990 | 0.9990 | 0.9990 | 0.9990 | 0.9990 | 0.9990 |
| $\varepsilon$ (Adam) | 1.000e-08 | 1.000e-08 | 1.000e-08 | 1.000e-08 | 1.000e-08 | 1.000e-08 |
| learning rate | 0.0006593 | 0.0005142 | 0.0008696 | 0.0007581 | 0.0007199 | 0.0007997 |
| weight decay (Adam) | 0.000 | 0.000 | 0.000 | 0.000 | 0.000 | 1.000e-07 |
| DPD protoform reconstruction weight | | 0.8200 | 0.9314 | | 0.5602 | 1.427 |
| DPD bridge network weight | | 0.3285 | 0.5967 | | 0.4405 | 0.5442 |
| DPD reflex prediction from gold protoform weight | | 0.5770 | 0.3550 | | 0.5421 | 0.4134 |
| DPD reflex prediction from reconstruction weight | | 0.8717 | 0.6495 | | 1.009 | 0.8214 |
| DPD CRINGE loss weight $\alpha$ | | 0.3174 | 0.2638 | | 0.001917 | 0.2851 |
| DPD CRINGE loss top $k$ | | 5 | 3 | | 4 | 4 |
| DPD shared embedding size | | 128 | 128 | | 128 | 128 |
| D2P encoder layers count | 2 | 2 | 2 | 2 | 2 | 2 |
| D2P dropout | 0.2404 | 0.2346 | 0.1761 | 0.1685 | 0.2935 | 0.3055 |
| D2P inference decode max length | 15 | 15 | 15 | 15 | 15 | 15 |
| D2P feedforward dimension | 512 | 512 | 512 | 512 | 512 | 512 |
| D2P embedding size | 128 | | | 256 | | |
| D2P model size | 128 | 64 | 128 | | | |
| D2P encoder layers count | | | | 2 | 2 | 2 |
| D2P number of heads | | | | 8 | 8 | 8 |
| D2P max input length | | | | 128 | 128 | 128 |
| P2D encoder layers count | | 2 | 2 | | 2 | 2 |
| P2D dropout | | 0.1692 | 0.3488 | | 0.3083 | 0.1804 |
| P2D inference decode max length | | 15 | 15 | | 15 | 15 |
| P2D feedforward dimension | | 512 | 512 | | 512 | 512 |
| P2D embedding size | | | | | | |
| P2D model size | | 128 | 64 | | | |
| P2D encoder layers count | | | | | 2 | 2 |
| P2D number of heads | | | | | 8 | 8 |
| P2D max input length | | | | | 128 | 128 |
| Bootstrapping starting epoch | | | | | | |
| Bootstrapping log probability threshold | | | | | | |
| Bootstrapping max new pseudo-labels per epoch | | | | | | |
| Π-model consistency ramp-up epochs | 4 | | 22 | 14 | | 14 |
| Π-model max consistency scaling | 257.4 | | 156.4 | 200.2 | | 142.0 |

Table 20: Additional hyperparameters for 100% labeled WikiHan.

| | GRU-ΠM | GRU-DPD | GRU-DPD-ΠM | Trans-ΠM | Trans-DPD | Trans-DPD-ΠM |
|---|---|---|---|---|---|---|
| batch size | 256 | 128 | 64 | 64 | 64 | 128 |
| max epochs | 379 | 309 | 307 | 295 | 218 | 365 |
| warmup epochs (learning rate) | 39 | 37 | 7 | 33 | 13 | 3 |
| $\beta_1$ (Adam) | 0.9000 | 0.9000 | 0.9000 | 0.9000 | 0.9000 | 0.9000 |
| $\beta_2$ (Adam) | 0.9990 | 0.9990 | 0.9990 | 0.9990 | 0.9990 | 0.9990 |
| $\varepsilon$ (Adam) | 1.000e-08 | 1.000e-08 | 1.000e-08 | 1.000e-08 | 1.000e-08 | 1.000e-08 |
| learning rate | 0.0008681 | 0.0008735 | 0.0005797 | 0.0006855 | 0.0006586 | 0.0006777 |
| weight decay (Adam) | 0.000 | 0.000 | 0.000 | 1.000e-07 | 0.000 | 1.000e-07 |
| DPD protoform reconstruction weight | | 0.5814 | 1.184 | | 0.5771 | 1.473 |
| DPD bridge network weight | | 0.7379 | 0.5496 | | 0.5159 | 0.1906 |
| DPD reflex prediction from gold protoform weight | | 0.3473 | 0.2647 | | 0.3375 | 0.5105 |
| DPD reflex prediction from reconstruction weight | | 1.147 | 0.5560 | | 0.5103 | 1.405 |
| DPD CRINGE loss weight $\alpha$ | | 0.1322 | 0.5651 | | 0.4757 | 0.09542 |
| DPD CRINGE loss top $k$ | | 4 | 4 | | 5 | 1 |
| DPD shared embedding size | | 256 | 384 | | 384 | 256 |
| D2P encoder layers count | 2 | 2 | 2 | 2 | 2 | 2 |
| D2P dropout | 0.2888 | 0.2923 | 0.3016 | 0.1566 | 0.2144 | 0.2418 |
| D2P inference decode max length | 30 | 30 | 30 | 30 | 30 | 30 |
| D2P feedforward dimension | 512 | 512 | 512 | 512 | 512 | 512 |
| D2P embedding size | 256 | | | 128 | | |
| D2P model size | 128 | 128 | 128 | | | |
| D2P encoder layers count | | | | 2 | 2 | 2 |
| D2P number of heads | | | | 8 | 8 | 8 |
| D2P max input length | | | | 128 | 128 | 128 |
| P2D encoder layers count | | 2 | 2 | | 2 | 2 |
| P2D dropout | | 0.1589 | 0.2117 | | 0.1996 | 0.2921 |
| P2D inference decode max length | | 30 | 30 | | 30 | 30 |
| P2D feedforward dimension | | 512 | 512 | | 512 | 512 |
| P2D embedding size | | | | | | |
| P2D model size | | 64 | 64 | | | |
| P2D encoder layers count | | | | | 2 | 2 |
| P2D number of heads | | | | | 8 | 8 |
| P2D max input length | | | | | 128 | 128 |
| Bootstrapping starting epoch | | | | | | |
| Bootstrapping log probability threshold | | | | | | |
| Bootstrapping max new pseudo-labels per epoch | | | | | | |
| Π-model consistency ramp-up epochs | 9 | | 15 | 30 | | 29 |
| Π-model max consistency scaling | 83.25 | | 92.97 | 67.85 | | 64.89 |

Table 21: Additional hyperparameters for 100% labeled Rom-phon.

| | Semisupervised | | 10%, exclude unlabeled | | 100% | |
|---|---|---|---|---|---|---|
| | WikiHan | Rom-phon | WikiHan | Rom-phon | WikiHan | Rom-phon |
| GRU-SUPV | 1,724,288 | 1,605,184 | 1,724,288 | 1,605,184 | | |
| GRU-ΠM | 1,231,232 | 1,605,184 | 1,231,232 | 1,605,184 | 1,724,288 | 1,605,184 |
| GRU-DPD | 2,851,584 | 3,036,416 | 3,130,112 | 2,929,024 | 2,851,584 | 3,101,952 |
| GRU-DPD-ΠM | 2,754,560 | 2,929,024 | 3,748,864 | 3,450,752 | 2,917,120 | 3,516,288 |
| Trans-SUPV | 3,938,917 | 3,846,935 | 3,938,917 | 3,846,935 | | |
| Trans-ΠM | 3,938,917 | 3,846,935 | 3,938,917 | 3,846,935 | 3,938,917 | 3,846,935 |
| Trans-DPD | 6,373,066 | 11,488,686 | 11,653,706 | 11,488,686 | 2,665,290 | 11,488,686 |
| Trans-DPD-ΠM | 2,665,290 | 11,488,686 | 2,665,290 | 11,488,686 | 2,665,290 | 6,262,958 |
| GRU-BST | 1,724,288 | 1,112,128 | | | | |
| GRU-ΠM-BST | 1,724,288 | 1,112,128 | | | | |
| GRU-DPD-BST | 2,851,584 | 3,516,288 | | | | |
| GRU-DPD-ΠM-BST | 3,218,944 | 3,516,288 | | | | |
| Trans-BST | 3,938,917 | 3,846,935 | | | | |
| Trans-ΠM-BST | 3,938,917 | 3,846,935 | | | | |
| Trans-DPD-BST | 6,373,066 | 11,488,686 | | | | |
| Trans-DPD-ΠM-BST | 6,373,066 | 6,262,958 | | | | |

Table 22: Trainable parameter count for each architecture-strategy combination in each experiment setup.

| **Dataset** | **Category** | **Count** | **Percentage** |
|---|---|---|---|
| WikiHan | Correct at all % | 140 | 13.55% |
| | Correct only above % threshold | 359 | 34.75% |
| | Correct only below % threshold | 37 | 3.58% |
| | Incorrect at all % | 278 | 26.91% |
| | Other pattern | 219 | 21.20% |
| Rom-phon | Correct at all % | 359 | 20.47% |
| | Correct only above % threshold | 451 | 25.71% |
| | Correct only below % threshold | 65 | 3.71% |
| | Incorrect at all % | 586 | 33.41% |
| | Other pattern | 293 | 16.70% |

Table 23: Distribution of reconstruction correctness patterns across labeling setting for Trans-DPD-ΠM-BST on group 1 WikiHan and GRU-DPD-BST on group 1 Rom-phon.

| Category | $\hat{y}$ @ 5% | $\hat{y}$ @ 10% | $\hat{y}$ @ 20% | $\hat{y}$ @ 30% | $\hat{y}$ @ 100% | $y$ (Reference) |
|---|---|---|---|---|---|---|
| Correct at all % | **bjwot入** | **bjwot入** | **bjwot入** | **bjwot入** | **bjwot入** | **bjwot入** |
| | **d͡zoj平** | **d͡zoj平** | **d͡zoj平** | **d͡zoj平** | **d͡zoj平** | **d͡zoj平** |
| | **jen平** | **jen平** | **jen平** | **jen平** | **jen平** | **jen平** |
| | **kʰwa平** | **kʰwa平** | **kʰwa平** | **kʰwa平** | **kʰwa平** | **kʰwa平** |
| | **lukʷ入** | **lukʷ入** | **lukʷ入** | **lukʷ入** | **lukʷ入** | **lukʷ入** |
| | **mjun平** | **mjun平** | **mjun平** | **mjun平** | **mjun平** | **mjun平** |
| | **t͡ɕuw平** | **t͡ɕuw平** | **t͡ɕuw平** | **t͡ɕuw平** | **t͡ɕuw平** | **t͡ɕuw平** |
| | **xjuw去** | **xjuw去** | **xjuw去** | **xjuw去** | **xjuw去** | **xjuw去** |
| Correct only above % threshold | bjuw去 | **bju去** | **bju去** | **bju去** | **bju去** | **bju去** |
| | bjwɨn平 | **bjwon平** | **bjwon平** | **bjwon平** | **bjwon平** | **bjwon平** |
| | laŋ去 | **ljaŋ去** | **ljaŋ去** | **ljaŋ去** | **ljaŋ去** | **ljaŋ去** |
| | maw上 | **mæw上** | **mæw上** | **mæw上** | **mæw上** | **mæw上** |
| | jew上 | **new上** | **new上** | **new上** | **new上** | **new上** |
| | bjwɨn平 | **pjun平** | **pjun平** | **pjun平** | **pjun平** | **pjun平** |
| | ʂjuw去 | **su去** | **su去** | **su去** | **su去** | **su去** |
| | d͡zaj去 | **t͡soj去** | **t͡soj去** | **t͡soj去** | **t͡soj去** | **t͡soj去** |
| | sen去 | **zjen去** | **zjen去** | **zjen去** | **zjen去** | **zjen去** |
| | ɲuŋʷ平 | ɲoŋʷ平 | **joŋʷ平** | **joŋʷ平** | **joŋʷ平** | **joŋʷ平** |
| | kwak入 | kwak入 | **kækʷ入** | **kækʷ入** | **kækʷ入** | **kækʷ入** |
| | mat入 | mat入 | **mwat入** | **mwat入** | **mwat入** | **mwat入** |
| | ɣæj平 | ɣoj平 | **ɣɛj平** | **ɣɛj平** | **ɣɛj平** | **ɣɛj平** |
| | ɖjuŋʷ平 | ɖjuŋʷ平 | t͡ʂʰuŋʷ平 | **t͡sʰuŋʷ平** | **t͡sʰuŋʷ平** | **t͡sʰuŋʷ平** |
| | t͡ʂʰjak入 | ɖjwak入 | t͡ɕak入 | **t͡ɕʰak入** | **t͡ɕʰak入** | **t͡ɕʰak入** |
| | gij平 | gij平 | gij平 | gij平 | **gi平** | **gi平** |
| | kwan去 | kwan去 | kwen去 | kwen去 | **kjwen去** | **kjwen去** |
| | kʰiŋ平 | kʰoŋ平 | kʰoŋ平 | kʰæŋ平 | **kʰɛŋ平** | **kʰɛŋ平** |
| | mam平 | kom平 | ɣam平 | ɣam平 | **ɣom平** | **ɣom平** |
| | t͡ɕaŋ平 | ʈiŋ平 | ʈɛŋ平 | ɖ͡ʐæŋ平 | **t͡ʂɛŋ平** | **t͡ʂɛŋ平** |
| | juŋʷ平 | joŋʷ平 | joŋʷ平 | joŋʷ平 | **ʔjoŋʷ平** | **ʔjoŋʷ平** |
| Correct only below % threshold | **jej去** | **jej去** | ʔej去 | ʔej去 | ʔjej去 | **jej去** |
| | **kij去** | **kij去** | kjɨj去 | kjɨj去 | kjɨj去 | **kij去** |
| Incorrect at all % | daw上 | daw上 | daw上 | daw上 | daw上 | **daw去** |
| | t͡sʰjen平 | d͡ʑem平 | d͡ʑem平 | d͡ʑem平 | d͡ʑem平 | **d͡ʑen平** |
| | gjo平 | gjo平 | gjo平 | gjo平 | gjo平 | **gju平** |
| | kwaŋ去 | kwaŋ去 | ɣwæŋ去 | gjwaŋ去 | kwaŋ去 | **kjwaŋ上** |
| | kʰwaŋ平 | kʰwaŋ平 | kʰwaŋ平 | kʰwaŋ平 | kʰjaŋ平 | **kʰjwaŋ平** |
| | pap入 | pæ上 | pæ上 | pæ上 | pæ上 | **pæ去** |
| | kep入 | kjep入 | kjep入 | t͡sjep入 | kjæp入 | **t͡sep入** |
| | ʈjwan上 | t͡san上 | t͡ʂɛn上 | t͡san上 | t͡san上 | **t͡swan上** |
| | tʰe去 | ɖi去 | t͡sʰi去 | t͡sʰi去 | t͡sʰje去 | **t͡sʰij去** |
| | ɣwɨj去 | ɣwej去 | ɣwej去 | ɣwej去 | ɣwej去 | **zwij去** |
| | jen去 | ŋjen去 | ŋjen去 | ŋjen去 | ŋjen去 | **ŋen去** |
| | ɲæn去 | ɲep入 | d͡ʑep入 | ɲep入 | ɲep入 | **ɕep入** |
| | t͡ɕi去 | t͡ɕe去 | ɖi去 | ɖi平 | ɖij去 | **ɖi上** |
| | ʂjwa上 | swa上 | swa上 | swa上 | swa上 | **ʂjo上** |
| | t͡ɕe平 | t͡sʰi平 | t͡sʰi平 | t͡ɕʰi平 | t͡sʰi平 | **t͡ʂʰje平** |
| | ʔaŋ去 | ʔaŋ去 | ʔaŋ去 | ʔaŋ去 | ʔaŋ去 | **ʔaŋ上** |
| Other pattern | **dan平** | don平 | **dan平** | **dan平** | **dan平** | **dan平** |
| | **mjew平** | mew平 | **mjew平** | **mjew平** | **mjew平** | **mjew平** |
| | **ɣwæ平** | ɣu平 | **ɣwæ平** | **ɣwæ平** | **ɣwæ平** | **ɣwæ平** |
| | **d͡zjem平** | zjem平 | **d͡zjem平** | zjem平 | **d͡zjem平** | **d͡zjem平** |
| | pjej去 | **pej去** | pjej去 | **pej去** | **pej去** | **pej去** |
| | **ŋæ平** | ŋeŋ平 | bæ平 | **ŋæ平** | **ŋæ平** | **ŋæ平** |
| | ɣjeŋ平 | **ɣeŋ平** | **ɣeŋ平** | **ɣeŋ平** | ɣæŋ平 | **ɣeŋ平** |
| | lukʷ入 | **jokʷ入** | jukʷ入 | lok入 | **jokʷ入** | **jokʷ入** |
| | kew上 | kʰæw上 | **kʰaw上** | kʰæw上 | **kʰaw上** | **kʰaw上** |
| | pjeŋ上 | pjeŋ上 | **pjæŋ上** | pjieŋ上 | **pjæŋ上** | **pjæŋ上** |
| | kwam平 | kʰom平 | **kʰam平** | kæm平 | kam平 | **kʰam平** |
| | min上 | min上 | min上 | **min平** | min上 | **min平** |
| | ɣjwæn去 | ɣjen去 | ɣjwen去 | **ɣwen去** | zjwen去 | **ɣwen去** |

Table 24: Sample protoform predictions by Trans-DPD-ΠM-BST for different labeling settings on group 1 WikiHan. Bold: correct protoform; $\hat{y}$ @ {5%, 10%, 20%, 30%, 100%}: protoform prediction when trained with a {5%, 10%, 20%, 30%, 100%} labeling setting; $y$: gold protoform.

| Category | $\hat{y}$ @ 5% | $\hat{y}$ @ 10% | $\hat{y}$ @ 20% | $\hat{y}$ @ 30% | $\hat{y}$ @ 100% | $y$ (Reference) |
|---|---|---|---|---|---|---|
| Correct at all % | **baltɛʊm** | **baltɛʊm** | **baltɛʊm** | **baltɛʊm** | **baltɛʊm** | **baltɛʊm** |
| | **dɪstɪllatɪonɛm** | **dɪstɪllatɪonɛm** | **dɪstɪllatɪonɛm** | **dɪstɪllatɪonɛm** | **dɪstɪllatɪonɛm** | **dɪstɪllatɪonɛm** |
| | **kakarɛ** | **kakarɛ** | **kakarɛ** | **kakarɛ** | **kakarɛ** | **kakarɛ** |
| | **mɔrsʊm** | **mɔrsʊm** | **mɔrsʊm** | **mɔrsʊm** | **mɔrsʊm** | **mɔrsʊm** |
| | **mɛdɪkamɛntʊm** | **mɛdɪkamɛntʊm** | **mɛdɪkamɛntʊm** | **mɛdɪkamɛntʊm** | **mɛdɪkamɛntʊm** | **mɛdɪkamɛntʊm** |
| | **purɪtatɛm** | **purɪtatɛm** | **purɪtatɛm** | **purɪtatɛm** | **purɪtatɛm** | **purɪtatɛm** |
| | **sanarɛ** | **sanarɛ** | **sanarɛ** | **sanarɛ** | **sanarɛ** | **sanarɛ** |
| | **skapʊlam** | **skapʊlam** | **skapʊlam** | **skapʊlam** | **skapʊlam** | **skapʊlam** |
| | **taktɪlɛm** | **taktɪlɛm** | **taktɪlɛm** | **taktɪlɛm** | **taktɪlɛm** | **taktɪlɛm** |
| | **warɪabɪlɛm** | **warɪabɪlɛm** | **warɪabɪlɛm** | **warɪabɪlɛm** | **warɪabɪlɛm** | **warɪabɪlɛm** |
| | **wɛnʊstʊm** | **wɛnʊstʊm** | **wɛnʊstʊm** | **wɛnʊstʊm** | **wɛnʊstʊm** | **wɛnʊstʊm** |
| | **wɛrmɛm** | **wɛrmɛm** | **wɛrmɛm** | **wɛrmɛm** | **wɛrmɛm** | **wɛrmɛm** |
| Correct only above % threshold | armɪpotɛntɛm | **armɪpɔtɛntɛm** | **armɪpɔtɛntɛm** | **armɪpɔtɛntɛm** | **armɪpɔtɛntɛm** | **armɪpɔtɛntɛm** |
| | enʊnkɪarɛ | **enʊntɪarɛ** | **enʊntɪarɛ** | **enʊntɪarɛ** | **enʊntɪarɛ** | **enʊntɪarɛ** |
| | ɪnfantɪam | **infantɪam** | **infantɪam** | **infantɪam** | **infantɪam** | **infantɪam** |
| | sʊbdʊm | **sʊrdʊm** | **sʊrdʊm** | **sʊrdʊm** | **sʊrdʊm** | **sʊrdʊm** |
| | ɪndɪkatɪwʊm | **ɪndɪkatiwʊm** | **ɪndɪkatiwʊm** | **ɪndɪkatiwʊm** | **ɪndɪkatiwʊm** | **ɪndɪkatiwʊm** |
| | ɛkwabɪlɪtatɛm | ɛkwabɪlɪtatɛm | **aɪkwabɪlɪtatɛm** | **aɪkwabɪlɪtatɛm** | **aɪkwabɪlɪtatɛm** | **aɪkwabɪlɪtatɛm** |
| | dɪskrɛtɪonɛm | dɪskrɛtɪonɛm | **dɪskretɪonɛm** | **dɪskretɪonɛm** | **dɪskretɪonɛm** | **dɪskretɪonɛm** |
| | plɛnarɪʊm | plɛnarɪʊm | **plenarɪʊm** | **plenarɪʊm** | **plenarɪʊm** | **plenarɪʊm** |
| | prɛrogatɪwam | prarɔgatɪwam | **praɪrɔgatiwam** | **praɪrɔgatiwam** | **praɪrɔgatiwam** | **praɪrɔgatiwam** |
| | prɛskribɛrɛ | prɛskribɛrɛ | **praɪskribɛrɛ** | **praɪskribɛrɛ** | **praɪskribɛrɛ** | **praɪskribɛrɛ** |
| | dɪlatorɪʊm | dɪlatɔrɪʊm | dɪlatorɪʊm | **dilatorɪʊm** | **dilatorɪʊm** | **dilatorɪʊm** |
| | fɛrɪtam | fɛrɪtam | fɛrɪtam | **fɛrɪtatɛm** | **fɛrɪtatɛm** | **fɛrɪtatɛm** |
| | ɛkstrarrɛ | ɛkstrarrɛ | ɛkstrarɛ | **ɛkstrahɛrɛ** | **ɛkstrahɛrɛ** | **ɛkstrahɛrɛ** |
| | ɛkwɪwokarɛ | ɛkwiwɔkarɛ | ɛkwɪwɔkarɛ | ɛkwɪwɔkarɛ | **aɪkwɪwɔkarɛ** | **aɪkwɪwɔkarɛ** |
| | kɔnsɛnsʊm | kɔnsɛnsʊm | konsɛnsʊm | konsɛnsʊm | **konsensʊm** | **konsensʊm** |
| | ɪŋkʊrabɪlɛm | ɪŋkʊrabɪlɛm | ɪŋkʊrabɪlɛm | ɪŋkʊrabɪlɛm | **ɪŋkurabɪlɛm** | **ɪŋkurabɪlɛm** |
| Correct only below % threshold | **ɪllustrarɛ** | **ɪllustrarɛ** | **ɪllustrarɛ** | ɪllʊstrarɛ | ɪllʊstrarɛ | **ɪllustrarɛ** |
| | **kɛrnɛrɛ** | **kɛrnɛrɛ** | **kɛrnɛrɛ** | **kɛrnɛrɛ** | kɛrnerɛ | **kɛrnɛrɛ** |
| Incorrect at all % | abɪssʊm | abɪssʊm | abɪssʊm | abɪssʊm | abɪssʊm | **abyssʊm** |
| | adesɪonɛm | adɛsɪonɛm | adesɪonɛm | adesɪonɛm | adesɪonɛm | **adhaɪsɪonɛm** |
| | allegɛrɛ | allegarɛ | allegɛrɛ | allegerɛ | allɪgɛrɛ | **allɛgarɛ** |
| | frotɪsɛtʊm | frotɪsɛtʊm | frotɪstɛtʊm | frotɪtɛtʊm | frʊtɪkɛtʊm | **frʊtɪketʊm** |
| | ɪpnotɪkʊm | ɪpnɔtɪkʊm | hymnɔtɪkʊm | hybnɔtɪkʊm | hymnɔtɪkʊm | **hypnotɪkʊm** |
| | kyrkʊndurrɛ | kɪrkondʊrrɛm | kɪrkɔndurrɛ | kɪrkondurrɛm | kɪrkʊmdurɛm | **kɪrkʊmdukɛrɛ** |
| | kʊwɪlɛm | kʊbɪlɛm | kʊwɪlɛm | kʊwɪlɛm | kʊbɪlɛm | **kʊbilɛ** |
| | marɪtalɛm | marɪtalɛm | marɪtalɛm | marɪtalɛm | marɪtalɛm | **maritalɛm** |
| | nuklɛarɛ | nʊklɛarɛ | nʊklɛarɛ | nuklɛarɛ | nuklɛarɛ | **nuklɛarɛm** |
| | parirɛ | parirɛ | parirɛ | parirɛ | parirɛ | **parɛrɛ** |
| | proklamatɪonɛm | proklamatɪonɛm | proklamatɪonɛm | proklamatɪonɛm | proklamatɪonɛm | **prɔklamatɪonɛm** |
| | pɛlɛgrinʊm | pɛlɛgrinʊm | pɛlɛrrinʊm | pɛlɛgrinʊm | pɛlɛgrinʊm | **pɛrɛgrinʊm** |
| | ɛkswɛrkʊm | ɛkswɛrkʊm | ɛkswɛrtʊm | skʊɛrtʊm | ɛkskɔdɛrɛ | **skɔrtɛʊm** |
| | skʊltorɛm | skʊltorɛm | skʊltorɛm | skʊltorɛm | skʊltorɛm | **skʊlptorɛm** |
| | sɔllatɪʊm | sʊllakɪʊm | sɔllaktɪʊm | sɔllakɪʊm | sollakɪʊm | **solakɪʊm** |
| | stanɪʊm | stanɪʊm | staŋnatɛm | stanɛʊm | stannʊm | **staŋnʊm** |
| | sɛttɛnnɛm | settɛnnɛm | sɛttɛnnɛm | sɛttɛnnɛm | sɛktɛnnɛm | **sɛptɛnnɛm** |
| | tɛrmɪtɛm | tɛrmɪtɛm | tɛrmɪtɛm | tɛrmɪtɛm | tɛrmɪtɛm | **tarmɪtɛm** |
| | trɪstɛtɪam | trɪstɛtɪam | trɪstɛtɪam | trɪstetɪam | trystɛktɪam | **tristɪtɪam** |
| | gɔwɛrnaɪɛm | gʊbɛrnam | guwɛrnatɛm | gʊwɛrnatɪʊm | gʊbɛrnatɪkʊm | **gʊbɛrnakʊlʊm** |
| Other pattern | **tɛporɛm** | tɛpɔrɛm | **tɛporɛm** | **tɛporɛm** | **tɛporɛm** | **tɛporɛm** |
| | **wadʊm** | waʊdʊm | **wadʊm** | **wadʊm** | **wadʊm** | **wadʊm** |
| | **ɪmmatɛrɪalɛm** | **ɪmmatɛrɪalɛm** | ɪmatɛrɪalɛm | ɪmatɛrɪalɛm | **ɪmmatɛrɪalɛm** | **ɪmmatɛrɪalɛm** |
| | **bustʊm** | bʊstʊm | **bustʊm** | bʊstʊm | bʊstʊm | **bustʊm** |
| | ɪŋkwɪlɪnʊm | ɪŋkwɪlɪnʊm | **ɪŋkwɪlinʊm** | ɪŋkwilinʊm | **ɪŋkwɪlinʊm** | **ɪŋkwɪlinʊm** |
| | ɛram | ɛram | ɛram | **aɪram** | heram | **aɪram** |
| | kɔsstɪtutɪonɛm | **kɔnstɪtutɪonɛm** | konstɪtutɪonɛm | konstɪtutɪonɛm | konstɪtutɪonɛm | **kɔnstɪtutɪonɛm** |
| | mɔlkɛrɛ | mɔlkɛrɛ | **mʊlkerɛ** | mʊlkɛrɛ | mʊlkɛrɛ | **mʊlkerɛ** |
| | pɛtatɛm | **pɛtasʊm** | pɛtasɛm | pɛtatɛm | pɛstas | **pɛtasʊm** |
| | ɛmpɪrɪkʊm | ɪmpɪrɪkʊm | **ɛmpirɪkʊm** | ɪmpirɪkʊm | ɛmpyrɪkʊm | **ɛmpirɪkʊm** |

Table 25: Sample protoform predictions by GRU-DPD-BST for different labeling settings on group 1 Rom-phon. Bold: correct protoform; $\hat{y}$ @ {5%, 10%, 20%, 30%, 100%}: protoform prediction when trained with a {5%, 10%, 20%, 30%, 100%} labeling setting; $y$: gold protoform.

| | | Trans-ΠM-BST | |
|---|---|---|---|
| | | correct | incorrect |
| Trans-DPD-ΠM-BST | correct | 334 (32.33%) | 112 (10.84%) |
| | incorrect | 69 (6.68%) | 518 (50.15%) |

| | | GRU-BST | |
|---|---|---|---|
| | | correct | incorrect |
| GRU-DPD-BST | correct | 546 (31.13%) | 126 (7.18%) |
| | incorrect | 96 (5.47%) | 986 (56.21%) |

Table 26: Confusion matrix of protoform prediction correctness for Trans-DPD-ΠM-BST vs. Trans-ΠM-BST on WikiHan (top) and GRU-DPD-BST vs. GRU-BST on Rom-phon (bottom)

| Cantonese | Gan | Hakka | Hokkien | Jin | Mandarin | Wu | Xiang | Middle Chinese | Middle Chinese$_{non\text{-}DPD}$ | Middle Chinese$_{DPD}$ | Cantonese$_{DPD}$ | Gan$_{DPD}$ | Hakka$_{DPD}$ | Hokkien$_{DPD}$ | Jin$_{DPD}$ | Mandarin$_{DPD}$ | Wu$_{DPD}$ | Xiang$_{DPD}$ |
|---|---|---|---|---|---|---|---|---|---|---|---|---|---|---|---|---|---|---|
| pɐi̯˧ | pʰi˩ | pi˥ | pe˧˩ | pi˥ | pi˥˩ | b̥i˩˧ | pe̯i̯˥ | bjej去 | **bjej去** | **bjej去** | **pɐi̯˧** | pʰi˧˥ | **pi˥** | pe˧ | **pi˥** | **pi˥˩** | **b̥i˩˧** | pi˥ |
| fɔːŋ˨˩ | - | - | haŋ˥ | - | faŋ˧˥ | - | - | bjwaŋ平 | **bjwaŋ平** | **bjwaŋ平** | **fɔːŋ˨˩** | - | - | hɔŋ˩˧ | - | **faŋ˧˥** | - | - |
| pʊk˧ | - | pʰuk˧ | pɔk˥ | pʰxaʔ˥˧ | pʰu˧˥ | b̥ʊʔ˨˩ | pʰu˧˥ | bukʷ入 | **bukʷ入** | bu平 | **pʊk˧** | - | **pʰuk˧** | pʰɔk˥ | pxu̯əʔ˥˧ | **pʰu˧˥** | **b̥ʊʔ˨˩** | **pʰu˧˥** |
| paː˥ | - | pa˩˧ | pe˧ | - | pa˥˩ | pa̠˥˧ | pa̠˩˧ | bwa上 | pa平 | pæ平 | **paː˥** | - | **pa˩˧** | pe˥ | - | pa˥ | **pa̠˥˧** | pa̠˧ |
| pʰuːy̯˨˩ | - | pʰi˨˩ | pu̯e˩˧ | - | pʰei̯˧˥ | b̥e̯˩˧ | - | bwoj平 | **bwoj平** | **bwoj平** | **pʰuːy̯˨˩** | - | **pʰi˨˩** | **pu̯e˩˧** | - | **pʰei̯˧˥** | **b̥e̯˩˧** | - |
| pʰaːi̯˨˩ | - | - | pai̯˥ | - | pʰai̯˧˥ | - | - | bɛj平 | boj平 | **bɛj平** | **pʰaːi̯˨˩** | - | - | pe˩˧ | - | pʰei̯˧˥ | - | - |
| taːn˧ | - | tan˥ | tã˧ | - | tan˥˩ | - | - | dan上 | tan去 | tan去 | **taːn˧** | - | **tan˥** | ta˧˩ | - | **tan˥˩** | - | - |
| t͡sʰɔːi̯˨˩ | - | t͡sʰoi̯˩ | t͡sai̯˩˧ | - | t͡sʰai̯˧˥ | z̥e̯˩˧ | - | d͡zoj平 | **d͡zoj平** | **d͡zoj平** | **t͡sʰɔːi̯˨˩** | - | t͡sʰa̯i˩ | **t͡sai̯˩˧** | - | **t͡sʰai̯˧˥** | **z̥e̯˩˧** | - |
| t͡sɔː˧ | - | t͡sʰo˥ | t͡sɤ˧ | - | t͡su̯ɔ˥˩ | z̥ʏʷ˩˧ | - | d͡zwa去 | **d͡zwa去** | **d͡zwa去** | t͡sʰɔː˧ | - | **t͡sʰo˥** | **t͡sɤ˧** | - | **t͡su̯ɔ˥˩** | **z̥ʏʷ˩˧** | - |
| søy˨˩ | - | - | su̯i˩˧ | - | ʈ͡ʂʰu̯ei̯˧˥ | - | - | d͡ʑwe平 | d͡ʑwij平 | d͡ʑwij平 | **søy˨˩** | - | - | **su̯i˩˧** | - | ʂu̯ei̯˧˥ | - | - |
| jiː˨˩ | - | - | i˩˧ | - | i˧˥ | - | - | ji平 | jij平 | je平 | **jiː˨˩** | - | - | **i˩˧** | - | **i˧˥** | - | - |
| kʰɐi̯˥ | - | - | ke˥ | - | t͡ɕi˥ | - | - | kej平 | **kej平** | **kej平** | kʷɐi̯˥ | - | - | **ke˥** | - | **t͡ɕi˥** | - | - |
| kɐm˥ | t͡ɕin˥˧ | kim˩˧ | kin˥ | t͡ɕĩŋ˩ | t͡ɕin˥ | t͡ɕɪŋ˥˧ | t͡ɕin˧ | kim平 | **kim平** | **kim平** | **kɐm˥** | **t͡ɕin˥˧** | **kim˩˧** | kim˥ | **t͡ɕĩŋ˩** | **t͡ɕin˥** | **t͡ɕɪŋ˥˧** | **t͡ɕin˧** |
| kei̯˧˥ | - | ki˧˩ | ku̯i˥˩ | - | t͡ɕi˨˩ | t͡ɕi˧˥ | - | kjɨj上 | kij上 | **kjɨj上** | kʷei̯˧˥ | - | **ki˧˩** | ki˥˩ | - | **t͡ɕi˨˩** | **t͡ɕi˧˥** | - |
| kuː˧˥ | - | - | kɔ˥ | - | ku˨˩ | - | - | ku上 | **ku上** | ku去 | **kuː˧˥** | - | - | **kɔ˥** | - | **ku˨˩** | - | - |
| liːu̯˧ | - | - | li̯au̯˩˧ | - | li̯oʊ̯˥˩ | - | - | lew平 | lew去 | ljew平 | **liːu̯˧** | - | - | li̯au̯˧ | - | **li̯oʊ̯˥˩** | - | - |
| liːp˧ | - | - | li̯ap˥ | - | li̯ɛ˥˩ | - | - | ljep入 | lep入 | **ljep入** | **liːp˧** | - | - | **li̯ap˥** | - | **li̯ɛ˥˩** | - | - |
| lou̯˨˩ | - | - | - | - | lu˧˥ | lʏʷ˩˧ | - | lu平 | **lu平** | **lu平** | **lou̯˨˩** | - | - | - | - | **lu˧˥** | **lʏʷ˩˧** | - |
| mei̯˧ | - | - | bi˧ | - | mei̯˥˩ | - | - | mij去 | **mij去** | mje去 | **mei̯˧** | - | - | **bi˧** | - | **mei̯˥˩** | - | - |
| miːn˩˧ | - | mi̯en˧˩ | bi̯ɛn˥˩ | - | mi̯ɛn˨˩ | mi˩˧ | - | mjen上 | **mjen上** | **mjen上** | **miːn˩˧** | - | **mi̯en˧˩** | **bi̯ɛn˥˩** | - | **mi̯ɛn˨˩** | **mi˩˧** | - |
| mɐn˧ | - | - | bun˧ | - | u̯ən˨˩ | - | - | mjun去 | mjun上 | mjun上 | mɐn˩˧ | - | - | bun˥˩ | - | **u̯ən˨˩** | - | - |
| mou̯˨˩ | u˥˧ | mu˩ | bu˩˧ | vu˩ | u˥ | ʏ̥ʏʷ˩˧ | u˧ | mju平 | mu平 | mu平 | **mou̯˨˩** | u˧˥ | **mu˩** | bɔ˩˧ | **vu˩** | u˧˥ | **ʏ̥ʏʷ˩˧** | ɸ˩˧ |
| mʊŋ˨˩ | - | - | bɔ˩˧ | - | mɤŋ˧˥ | - | - | muŋʷ平 | **muŋʷ平** | **muŋʷ平** | **mʊŋ˨˩** | - | - | bɔŋ | - | **mɤŋ˧˥** | - | - |
| maː˧ | - | ma˥ | me˧ | - | ma˥˩ | mo˩˧ | - | mæ上 | mæ去 | mæ去 | **maː˧** | - | **ma˥** | be˧ | - | **ma˥˩** | **mo˩˧** | - |
| pɪŋ˥ | - | pen˧˥ | pi̯əŋ˥ | - | piŋ˥ | - | - | piŋ平 | pjeŋ平 | pjeŋ平 | **pɪŋ˥** | - | pin˧˥ | **pi̯əŋ˥** | - | **piŋ˥** | - | - |
| piːt˧ | pʰi̯ɛt˥ | pi̯et˧ | piʔ˧ | pi̯əʔ˧ | pi̯ɛ˥ | pi̯ɪʔ˥ | pʰi̯e̯˩˧ | pjet入 | pʰet入 | pʰjet入 | **piːt˧** | pi̯ɛt˥ | **pi̯et˧** | **piʔ˧** | **pi̯əʔ˧** | pi˥ | **pi̯ɪʔ˥** | pi̯e̯˩˧ |
| piːu̯˥ | - | - | pi̯au̯˥ | - | pi̯au̯˥ | - | - | pjew平 | **pjew平** | **pjew平** | **piːu̯˥** | - | - | **pi̯au̯˥** | - | **pi̯au̯˥** | - | - |
| paː˧˥ | - | pa˧˩ | pe˥˩ | - | pa˨˩ | - | - | pæ去 | pæ上 | pæ上 | **paː˧˥** | - | **pa˧˩** | pa˥˩ | - | **pa˨˩** | - | - |
| pʰiːu̯˥ | - | - | - | - | pʰi̯au̯˥ | pʰi̯ɔ˥˧ | - | pʰjew平 | **pʰjew平** | **pʰjew平** | **pʰiːu̯˥** | - | - | - | - | **pʰi̯au̯˥** | **pʰi̯ɔ˥˧** | - |
| pʰʊk˥ | - | pʰu˥ | pʰɔk˧ | - | pʰu˥ | pʰʊʔ˥ | - | pʰukʷ入 | **pʰukʷ入** | **pʰukʷ入** | **pʰʊk˥** | - | pʰuk | **pʰɔk˧** | - | **pʰu˥** | **pʰʊʔ˥** | - |
| pɔːk˥ | - | pʰuk˧ | pʰɔk˧ | - | pʰu˥ | - | - | pʰukʷ入 | bukʷ入 | pʰu平 | pʰɔːk˧ | - | **pʰuk˧** | **pʰɔk˧** | - | pʰu˧˥ | - | - |
| tʰaːi̯˧ | - | - | tʰai̯˧˩ | - | tʰai̯˥˩ | - | - | tʰaj去 | tʰoj去 | tʰoj去 | tʰɔːi̯˧ | - | - | **tʰai̯˧˩** | - | **tʰai̯˥˩** | - | - |
| t͡sʰɔːi̯˧˥ | - | t͡sʰa̯i˧˩ | t͡sʰai̯˥˩ | - | t͡sʰai̯˨˩ | t͡sʰe̯˧˥ | - | t͡sʰoj上 | ʈ͡ʂʰæj上 | ʈ͡ʂʰoj上 | **t͡sʰɔːi̯˧˥** | - | **t͡sʰa̯i˧˩** | **t͡sʰai̯˥˩** | - | **t͡sʰai̯˨˩** | **t͡sʰe̯˧˥** | - |
| t͡siː˧ | - | - | - | - | ʈ͡ʂʐ̩˥˩ | t͡sz̩˧˥ | - | t͡ɕi去 | t͡ɕe去 | t͡ɕe去 | **t͡siː˧** | - | - | - | - | t͡sz̩˥˩ | **t͡sz̩˧˥** | - |
| t͡sɐu̯˥ | t͡si̯u˥˧ | t͡su˧˥ | t͡ɕi̯u˥ | t͡səu̯˩ | ʈ͡ʂoʊ̯˥ | t͡sɜ˥˧ | ʈ͡ʂə̯u˧ | t͡ɕuw平 | **t͡ɕuw平** | **t͡ɕuw平** | **t͡sɐu̯˥** | **t͡si̯u˥˧** | **t͡su˧˥** | **t͡ɕi̯u˥** | **t͡səu̯˩** | **ʈ͡ʂoʊ̯˥** | **t͡sɜ˥˧** | **ʈ͡ʂə̯u˧** |
| hei̯˥ | - | - | hi˥ | - | ɕi˥ | - | - | xi平 | xij平 | xje平 | høy̯˥ | - | - | **hi˥** | - | ɕy˥ | - | - |
| hœː˥ | ɕy̯e˥˧ | hi̯o˧˥ | hi̯a˥ | ɕy̯e˩ | ɕyɤ˥ | ɕi̯o˥˧ | ɕy̯a˧ | xjwa平 | xja平 | xja平 | **hœː˥** | ɕi̯a˥˧ | **hi̯o˧˥** | **hi̯a˥** | ɕi̯õ˩ | ɕi̯aɤ˥ | ɕi̯a˥˧ | ɕi̯a˧ |
| syːn˧ | - | - | su̯an˧ | - | ɕyan˥˩ | - | - | zjwen平 | d͡ʑwan去 | swen去 | **syːn˧** | - | - | su̯an˧˩ | - | **ɕyan˥˩** | - | - |
| sɪk˥ | sz̩it˥ | sit˧ | ɕi̯ək˧ | səʔ˧ | ʂʐ̩˥˩ | səʔ˥ | ʂʐ̩˧˥ | ɕik入 | ɕek入 | ɕek入 | **sɪk˥** | **sz̩it˥** | **sit˧** | **ɕi̯ək˧** | **səʔ˧** | **ʂʐ̩˥˩** | **səʔ˥** | ʂʐ̩˩˧ |
| sɐu̯˧˥ | - | su˧˩ | ɕi̯u˥˩ | - | ʂoʊ̯˨˩ | - | - | ɕuw上 | **ɕuw上** | **ɕuw上** | **sɐu̯˧˥** | - | se̯u˧˩ | **ɕi̯u˥˩** | - | **ʂoʊ̯˨˩** | - | - |
| t͡sɪk˧ | - | tat˥ | ti̯ək˥ | - | ʈ͡ʂʐ̩˧˥ | z̥əʔ˨˩ | - | ɖi去 | ʈik入 | ʈik入 | **t͡sɪk˧** | - | t͡sɨk˧ | **ti̯ək˥** | - | ʈ͡ʂʐ̩˥˩ | **z̥əʔ˨˩** | - |
| t͡sʊk˧ | - | t͡sʰuk˥ | ti̯ɔk˥ | - | ʈ͡ʂu˧˥ | - | - | ɖjukʷ入 | ʈjukʷ入 | **ɖjukʷ入** | **t͡sʊk˧** | - | t͡sʰuk˧ | ti̯ɔk˧ | - | **ʈ͡ʂu˧˥** | - | - |
| hɔːn˩˧ | - | hon˧˥ | ũ̯ã˧ | xæ̃˥ | xan˥˩ | ɦø˩˧ | - | ɣan上 | **ɣan上** | **ɣan上** | hɔːn˧ | - | hon˥ | h˧ | **xæ̃˥** | **xan˥˩** | **ɦø˩˧** | - |
| siː˥ | sz̩˥˧ | si˧˥ | sai̯˥ | sz̩˩ | ʂʐ̩˥ | sz̩˥˧ | sz̩˧ | ɕij平 | si平 | ɕe平 | **siː˥** | **sz̩˥˧** | **si˧˥** | ɕu˥ | **sz̩˩** | **ʂʐ̩˥** | **sz̩˥˧** | ʂʐ̩˧ |
| siː˥ | - | - | su˥ | - | ʂʐ̩˥ | - | - | ɕje平 | si平 | ɕe平 | **siː˥** | - | - | ɕu˥ | - | **ʂʐ̩˥** | - | - |
| t͡siː˥ | t͡sz̩˥˧ | ti˧˥ | t͡sai̯˥ | t͡sz̩˩ | ʈ͡ʂʐ̩˥ | t͡sz̩˥˧ | ʈ͡ʂʐ̩˧ | ʈje平 | t͡ɕe平 | t͡ɕe平 | **t͡siː˥** | **t͡sz̩˥˧** | t͡si˧˥ | t͡ɕe˥ | **t͡sz̩˩** | **ʈ͡ʂʐ̩˥** | **t͡sz̩˥˧** | **ʈ͡ʂʐ̩˧** |
| t͡syːt˧ | - | t͡sʰot˧ | t͡sʰeʔ˧ | - | ʈ͡ʂʰu̯ɔ˥˩ | t͡sʰʊʔ˥ | - | ʈjwej去 | t͡sʰwat入 | ɖjwet入 | t͡sʰyːt˧ | - | t͡sʰut˧ | t͡sʰu̯aʔ˧ | - | **ʈ͡ʂʰu̯ɔ˥˩** | t͡sʰəʔ˥ | - |
| t͡sʰaː˥ | - | t͡sʰa˧˥ | t͡sʰa˥ | - | ʈ͡ʂʰa˥ | t͡sʰo˥˧ | - | ʈ͡ʂʰje平 | ʈ͡ʂʰæ平 | ʈ͡ʂʰæ平 | **t͡sʰaː˥** | - | **t͡sʰa˧˥** | **t͡sʰa˥** | - | **ʈ͡ʂʰa˥** | **t͡sʰo˥˧** | - |
| ɔːi̯˥ | ŋai̯˥˧ | a̯i˧˥ | ai̯˥ | - | ai̯˥ | e̯˧˥ | ŋai̯˩˧ | ʔoj平 | ʔaj平 | **ʔoj平** | **ɔːi̯˥** | **ŋai̯˥˧** | oi̯˧˥ | **ai̯˥** | - | **ai̯˥** | e̯˥˧ | ŋai̯˧ |

Table 27: Sample reconstruction predictions by Trans-ΠM-BST (denoted DPD) and Trans-DPD-ΠM-BST (denoted non-DPD) and sample reflex prediction predictions by Trans-DPD-ΠM-BST (with latent reconstruction as input) for WikiHan (proportionate sampling according to the confusion matrix in Table 26). **Languaĝe**: prediction for **Language**; bold: correct prediction; '-': the dataset does not contain this reflex.

| Romanian | French | Italian | Spanish | Portuguese | Latin | $\widehat{\text{Latin}}_{\text{non-DPD}}$ | $\widehat{\text{Latin}}_{\text{DPD}}$ | $\widehat{\text{Romanian}}_{\text{DPD}}$ | $\widehat{\text{French}}_{\text{DPD}}$ | $\widehat{\text{Italian}}_{\text{DPD}}$ | $\widehat{\text{Spanish}}_{\text{DPD}}$ | $\widehat{\text{Portuguese}}_{\text{DPD}}$ |
|---|---|---|---|---|---|---|---|---|---|---|---|---|
| - | adikt | addetto | aðikto | ɐditʊ | addɪktʊm | **addɪktʊm** | **addɪktʊm** | - | **adikt** | **addetto** | **aðikto** | **ɐditʊ** |
| - | apɑ̃dʁ | appendere | apender | - | appɛndɛrɛ | **appɛndɛrɛ** | **appɛndɛrɛ** | - | apɑ̃de | **appɛndere** | **apender** | - |
| ardoare | aʁdœʁ | ardore | arðor | ɐrədoɹ | ardorɛm | **ardorɛm** | ardɔrɛm | ardoare | aʁdoʁ | **ardore** | **arðor** | **ɐrədoɹ** |
| - | - | arpinate | - | - | arpinatɛm | arpɪnatɛm | arpɪnatɛm | - | - | **arpinate** | - | - |
| - | bɛstjɛʁ | bestjario | bestjarjo | - | bestɪarɪʊm | bɛstɪarɪʊm | bɛstɪarɪʊm | - | **bɛstjɛʁ** | **bestjario** | **bestjarjo** | - |
| bovin | bovɛ̃ | bovino | boβino | buvinʊ | bɔwinʊm | bɔwɪnʊm | bɔwɪnʊm | **bovin** | bonɛ̃ | **bovino** | **boβino** | **buvinʊ** |
| - | - | - | - | dibɐkaɹ | debakkʰari | debakarɛ | debakarɛ | - | - | - | - | **dibɐkaɹ** |
| debil | - | dɛbile | deβil | dɛbil | debɪlɛm | **debɪlɛm** | **debɪlɛm** | debile | - | **dɛbile** | deβile | dɛbili |
| - | delɛ̃kɑ̃s | delinkwɛntsa | delinkwenθja | diliŋkweiŋsjɐ | delɪŋkwentɪam | **delɪŋkwentɪam** | **delɪŋkwentɪam** | - | **delɛ̃kɑ̃s** | **delinkwɛntsa** | **delinkwɛnθja** | **diliŋkweiŋsjɐ** |
| - | - | dilidʒere | - | - | dɪlɪgɛrɛ | dɪlɪdɛrɛ | dɪlɪgɛrɛ | - | - | **dilidʒere** | - | - |
| flwid | flyid | fluido | fluiðo | fluɪdʊ | flʊɪdʊm | fludʊm | fluɪdʊm | fluid | **flyid** | **fluido** | **fluiðo** | **fluɪdʊ** |
| - | fɛʁmɔte | - | - | - | fɪrmɪtatɛm | fermatɛrɛ | fermatɛm | - | **fɛʁmɔte** | - | - | - |
| tʃeapə | siv | - | - | - | kepam | kɪwam | kipam | tʃeamə | siva | - | - | - |
| - | kwadʁilatɛʁ | kwadrilatero | kwaðrilatero | - | kwadrɪlatɛrʊm | kwadrilatɛrʊm | **kwadrɪlatɛrʊm** | - | **kwadʁilatɛʁ** | **kwadrilatero** | **kwaðrilatero** | - |
| - | - | - | - | kɛrulʊ | kwerʊlʊm | kerʊlʊm | kerʊlʊm | - | - | - | - | **kɛrulʊ** |
| - | - | kommizerare | - | - | kɔmmɪsɛrari | kɔmɪserarɛ | kɔmmiserarɛ | - | - | **kommizerare** | - | - |
| konsubstantsjal | kɔ̃sybstɑ̃sjɛl | konsustantsjale | konsustanθjal | kuŋsubstɛ̃ŋsiɑl | kɔnsʊbstantɪalɛm | kɔnsʊstantɪalɛm | kɔnsʊbstantɪalɛm | **konsubstantsjal** | kɔ̃sybstɑ̃tjɛl | **konsustantsjale** | **konsustanθjal** | **kuŋsubstɛ̃ŋsiɑl** |
| kumnat | koɲa | koɲato | kuɲaðo | - | kɔɲatʊm | kʊɪatʊm | kɔnatʊm | kunnat | kuɲa | **koɲato** | kogato | - |
| labil | labil | labile | laβil | labil | labɪlɛm | **labɪlɛm** | **labɪlɛm** | labile | **labil** | **labile** | laβile | labili |
| - | mitiʒe | mitigare | mitiɣar | - | mitɪgarɛ | **mitɪgarɛ** | mɪtɪgarɛ | - | mitige | **mitigare** | **mitiɣar** | - |
| - | - | mendatʃe | - | meŋdasi | mendakem | **mendakem** | **mendakem** | - | - | **mendatʃe** | - | meŋdaʃi |
| nodos | - | nodozo | nuðoso | nudozʊ | nodosʊm | nɔdosʊm | nɔdɔsʊm | nodus | - | **nodozo** | noðoso | nuduzʊ |
| notə | nɔt | nɔta | nota | nɔtɐ | notam | nɔtam | nɔtam | **notə** | nom | **nɔta** | **nota** | **nɔtɐ** |
| oratsje | oʁezɔ̃ | oratsione | oraθjon | ɔrɐsɐ̃ʊ̃ | oratɪonɛm | ɔratɪonɛm | ɔratɪonɛm | **oratsje** | oʁasj | **oratsione** | **oraθjon** | **ɔrɐsɐ̃ʊ̃** |
| prefekturə | pʁefɛktyʁ | prefettura | prefektura | pɹifeituɾɐ | praɪfekturam | prɛfɛktʊram | prɛfɛktʊram | **prefekturə** | **pʁefɛktyʁ** | prefftura | **prefektura** | **pɹifeituɾɐ** |
| - | pʁesede | pretʃedere | preθeðer | - | praɪkedɛrɛ | praɪkedɛrɛ | praɪkɛdɛrɛ | - | **pʁesede** | **pretʃedere** | **preθeðer** | - |
| protʃes | pʁosɛ | protʃesso | proθeso | pɹusɛsʊ | prokessʊm | **prokɛssʊm** | **prokɛssʊm** | **protʃes** | pʁosɛs | **protʃesso** | **proθeso** | **pɹusɛsʊ** |
| purifika | pyʁifje | purifikare | purifikar | puɾifikaɹ | purɪfɪkarɛ | **purɪfɪkarɛ** | **purɪfɪkarɛ** | **purifika** | pyʁifie | **purifikare** | **purifikar** | **puɾifikaɹ** |
| pekuliw | pekyl | pekulio | pekuljo | pekuliʊ | pekulɪʊm | pekʊlɪʊm | pekʊlɪʊm | **pekuliw** | **pekyl** | **pekulio** | **pekuljo** | pekuljʊ |
| - | - | - | puðoroso | - | pʊdorosʊm | pʊdɔrosʊm | pudorosʊm | - | - | - | **puðoroso** | - |
| - | - | - | pultaθeo | - | pʊltɛm | pʊltateʊm | pʊltatɛʊm | - | - | - | pultad | - |
| - | sanasjɔ̃ | sanatsione | sanaθjon | - | sanatɪonɛm | **sanatɪonɛm** | **sanatɪonɛm** | - | **sanasjɔ̃** | **sanatsione** | **sanaθjon** | - |
| sartʃinə | - | sartʃina | sarθina | - | sarkɪnam | **sarkɪnam** | sarkinam | **sartʃinə** | - | **sartʃina** | **sarθina** | - |
| - | - | speme | - | - | spɛm | spɛmɛm | spɛmɛm | - | - | **speme** | - | - |
| - | sɛʁvityd | - | serβiðumbre | serəvidɐ̃ʊ̃ | serwɪtudɪnɛm | serwɪtʊm | serwɪtʊmbrɛm | - | sɛʁvitym | - | serβiðunbre | serəvituɐ̃ʊ̃ |
| - | tʁɑ̃spɔʁte | trasportare | transportar | tɹɐ̃ŋspurətaɹ | transpɔrtarɛ | transportarɛ | **transpɔrtarɛ** | - | **tʁɑ̃spɔʁte** | **trasportare** | **transportar** | **tɹɐ̃ŋspurətaɹ** |
| tristetsə | tʁistɛs | tristetssa | tristeθa | tɹiʃtezɐ | tristɪtɪam | trɪstetam | trɪstɛtɪam | **tristetsə** | tʁistɛt | **tristetssa** | tristeθj | tɹiʃtisɐ |
| tub | tœb | tubo | tuβo | tubʊ | tʊbʊm | **tʊbʊm** | **tʊbʊm** | **tub** | tym | **tubo** | **tuβo** | **tubʊ** |
| venjal | venjɛl | venjale | benjal | veniɑl | wenɪalɛm | **wenɪalɛm** | **wenɪalɛm** | **venjal** | venjal | **venjale** | **benjal** | **veniɑl** |
| - | vɛʁtiʒinø | vertidʒinozo | bertixinoso | verətiʒinozʊ | wertigɪnosʊm | wertɪgɪnosʊm | wertɪgɪnosʊm | - | **vɛʁtiʒinø** | **vertidʒinozo** | **bertixinoso** | **verətiʒinozʊ** |
| vəduvə | vøv | vedova | bjuða | viuvɐ | wɪdʊam | wediwam | wedʊwam | veduwə | **vøv** | vuduvo | bjuβo | veuvʊ |
| ivoriw | ivwaʁ | avɔrio | eβore | - | ɛbʊr | ɪwwarɪʊm | aworɪʊm | evɔrek | evwaʁ | avɔrea | eβor | - |
| - | ɛkspyʁʒe | spurgare | ekspurɣar | - | ɛkspʊrgarɛ | spʊrgarɛ | **ɛkspʊrgarɛ** | - | ɛkspyʁge | **spurgare** | **ekspurɣar** | - |
| - | ɛksikasjɔ̃ | essikkatsione | - | - | ɛkssɪkkatɪonɛm | ɛkssɪkatɪonɛm | **ɛkssɪkkatɪonɛm** | - | ɛksiikasjɔ̃ | **essikkatsione** | - | - |
| episkopat | episkopa | episkopato | episkopaðo | epiʃkupadʊ | ɛpɪskɔpatʊm | ɛpɪskɔpatʊm | ɛpɪskɔpatʊm | **episkopat** | **episkopa** | **episkopato** | **episkopaðo** | **epiʃkupadʊ** |
| gladjum | glɛv | gjado | glaðjo | gladjʊ | gladɪʊm | **gladɪʊm** | **gladɪʊm** | gladwuə | glad | gjadu | glaðwo | glɐduʊ |
| gratsjos | gʁasjø | gratsiozo | graθjoso | gɹɐsiozʊ | gratɪosʊm | **gratɪosʊm** | **gratɪosʊm** | **gratsjos** | **gʁasjø** | **gratsiozo** | **graθjoso** | **gɹɐsiozʊ** |
| - | - | dʒeometria | xeometria | - | gɛɔmetrɪam | jɛɔmetrɪam | gɛɔmetrɪam | - | - | **dʒeometria** | **xeometria** | - |
| - | - | inalbare | - | - | ɪnalbarɛ | **ɪnalbarɛ** | **ɪnalbarɛ** | - | - | **inalbare** | - | - |
| - | - | indagare | indaɣar | - | ɪndagarɛ | **ɪndagarɛ** | **ɪndagarɛ** | - | - | **indagare** | **indaɣar** | - |

Table 28: Sample reconstruction predictions by GRU-DPD-BST (denoted DPD) and GRU-BST (denoted non-DPD) and sample reflex prediction predictions by GRU-DPD-BST (with latent reconstruction as input) for Rom-phon (proportionate sampling according to the confusion matrix in Table 26). $\widehat{\textbf{Language}}$: prediction for **Language**; bold: correct prediction; '-': the dataset does not contain this reflex.

| | Cantonese | Gan | Hakka | Hokkien | Jin | Mandarin | Wu | Xiang | Middle Chinese |
|---|---|---|---|---|---|---|---|---|---|
| Reference | **t͡sʰaːm˨˩** | - | - | **t͡sʰam˨˦** | - | **t͡sʰan˧˥** | - | - | **d͡zam平** |
| GRU-SUPV | | | | | | | | | d͡zæm平 |
| GRU-ΠM | | | | | | | | | d͡zjem平 |
| GRU-BST | | | | | | | | | t͡sʰæm平 |
| GRU-ΠM-BST | | | | | | | | | d͡ʑom平 |
| GRU-DPD | **t͡sʰaːm˨˩** | - | - | **t͡sʰam˨˦** | - | t͡sʰan˥ | - | - | d͡zæm平 |
| GRU-DPD-ΠM | **t͡sʰaːm˨˩** | - | - | t͡sam˨˦ | - | **t͡sʰan˧˥** | - | - | ʈ͡ʂʰæm平 |
| GRU-DPD-BST | **t͡sʰaːm˨˩** | - | - | **t͡sʰam˨˦** | - | **t͡sʰan˧˥** | - | - | ɖæm平 |
| GRU-DPD-ΠM-BST | **t͡sʰaːm˨˩** | - | - | **t͡sʰam˨˦** | - | **t͡sʰan˧˥** | - | - | **d͡zam平** |
| Trans-SUPV | | | | | | | | | ʈ͡ʂʰæm平 |
| Trans-ΠM | | | | | | | | | d͡zæm平 |
| Trans-BST | | | | | | | | | ʈ͡ʂʰæm平 |
| Trans-ΠM-BST | | | | | | | | | **d͡zam平** |
| Trans-DPD | **t͡sʰaːm˨˩** | - | - | **t͡sʰam˨˦** | - | **t͡sʰan˧˥** | - | - | ʈ͡ʂʰæm平 |
| Trans-DPD-ΠM | **t͡sʰaːm˨˩** | - | - | **t͡sʰam˨˦** | - | **t͡sʰan˧˥** | - | - | t͡sʰam平 |
| Trans-DPD-BST | **t͡sʰaːm˨˩** | - | - | **t͡sʰam˨˦** | - | **t͡sʰan˧˥** | - | - | **d͡zam平** |
| Trans-DPD-ΠM-BST | **t͡sʰaːm˨˩** | - | - | **t͡sʰam˨˦** | - | **t͡sʰan˧˥** | - | - | **d͡zam平** |
| Reference | **jœːk˨** | **i̯ɔʔ˥** | **i̯ok˥** | **i̯ɤʔ˥** | **i̯əʔ˨** | **i̯au̯˥˩** | **ɦi̯a̯ʔ˩˨** | **i̯o˨˦** | **jak入** |
| GRU-SUPV | | | | | | | | | **jak入** |
| GRU-ΠM | | | | | | | | | **jak入** |
| GRU-BST | | | | | | | | | **jak入** |
| GRU-ΠM-BST | | | | | | | | | **jak入** |
| GRU-DPD | **jœːk˨** | **i̯ɔʔ˥** | **i̯ok˥** | i̯aʔ˥ | **i̯əʔ˨** | y̑ɛ˧˥ | **ɦi̯a̯ʔ˩˨** | i̯ę˨˦ | jæk入 |
| GRU-DPD-ΠM | **jœːk˨** | **i̯ɔʔ˥** | **i̯ok˥** | i̯ɔʔ˥ | y̑əʔ˨ | y̑ɛ˥˩ | **ɦi̯a̯ʔ˩˨** | **i̯o˨˦** | **jak入** |
| GRU-DPD-BST | **jœːk˨** | **i̯ɔʔ˥** | i̯ok˨ | i̯ɔʔ˥ | **i̯əʔ˨** | y̑ɛ˥˩ | **ɦi̯a̯ʔ˩˨** | **i̯o˨˦** | **jak入** |
| GRU-DPD-ΠM-BST | **jœːk˨** | **i̯ɔʔ˥** | i̯ok˨ | i̯ɔʔ˥ | **i̯əʔ˨** | y̯ɛ˥˩ | **ɦi̯a̯ʔ˩˨** | **i̯o˨˦** | **jak入** |
| Trans-SUPV | | | | | | | | | jekʷ入 |
| Trans-ΠM | | | | | | | | | **jak入** |
| Trans-BST | | | | | | | | | jek入 |
| Trans-ΠM-BST | | | | | | | | | **jak入** |
| Trans-DPD | **jœːk˨** | **i̯ɔʔ˥** | **i̯ok˥** | i̯ɔʔ˥ | y̑əʔ˨ | **i̯au̯˥˩** | ɦi̯ʊʔ˩˨ | **i̯o˨˦** | jew去 |
| Trans-DPD-ΠM | **jœːk˨** | **i̯ɔʔ˥** | i̯ok˨ | i̯ɔʔ˥ | **i̯əʔ˨** | i̯ɛ˥˩ | **ɦi̯a̯ʔ˩˨** | i̯ę˨˦ | **jak入** |
| Trans-DPD-BST | **jœːk˨** | **i̯ɔʔ˥** | i̯ok˨ | i̯ɔʔ˥ | **i̯əʔ˨** | y̑ɛ˨˩ | **ɦi̯a̯ʔ˩˨** | **i̯o˨˦** | **jak入** |
| Trans-DPD-ΠM-BST | **jœːk˨** | **i̯ɔʔ˥** | i̯ok˨ | i̯aʔ˥ | **i̯əʔ˨** | y̑ɛ˥˩ | **ɦi̯a̯ʔ˩˨** | **i̯o˨˦** | jaw去 |
| Reference | **fɐt˥** | - | **fut˨** | **hut˧** | **xu̯aʔ˨** | **xu˥** | - | - | **xwot入** |
| GRU-SUPV | | | | | | | | | ɣwot入 |
| GRU-ΠM | | | | | | | | | ɣuw去 |
| GRU-BST | | | | | | | | | ɣwot入 |
| GRU-ΠM-BST | | | | | | | | | ɣjut入 |
| GRU-DPD | **fɐt˥** | - | fat˨ | **hut˧** | fu̯aʔ˨ | fu˧˥ | - | - | kʰwot入 |
| GRU-DPD-ΠM | **fɐt˥** | - | **fut˨** | **hut˧** | xu̯əʔ˨ | **xu˥** | - | - | **xwot入** |
| GRU-DPD-BST | faːt˥ | - | vat˨ | hu̯at˧ | **xu̯aʔ˨** | xua | - | - | ɣwot入 |
| GRU-DPD-ΠM-BST | **fɐt˥** | - | fit˨ | **hut˧** | xu̯əʔ˨ | **xu˥** | - | - | xjut入 |
| Trans-SUPV | | | | | | | | | xjut入 |
| Trans-ΠM | | | | | | | | | **xwot入** |
| Trans-BST | | | | | | | | | xjwot入 |
| Trans-ΠM-BST | | | | | | | | | **xwot入** |
| Trans-DPD | **fɐt˥** | - | fat˨ | hu̯at˧ | **xu̯aʔ˨** | xua | - | - | xwat入 |
| Trans-DPD-ΠM | **fɐt˥** | - | **fut˨** | **hut˧** | fu̯əʔ˨ | **xu˥** | - | - | **xwot入** |
| Trans-DPD-BST | **fɐt˥** | - | fit˨ | **hut˧** | xu̯əʔ˨ | **xu˥** | - | - | **xwot入** |
| Trans-DPD-ΠM-BST | **fɐt˥** | - | fit˨ | **hut˧** | ɕu̯əʔ˨ | ɕu˥ | - | - | xut入 |
| Reference | **saːn˨˩** | - | - | **t͡sʰan˥** | - | **ʈ͡ʂʰan˧˥** | - | - | **ɖ͡ʐɛn平** |
| GRU-SUPV | | | | | | | | | t͡sʰæn平 |
| GRU-ΠM | | | | | | | | | d͡ʑæn平 |
| GRU-BST | | | | | | | | | d͡ʑæn平 |
| GRU-ΠM-BST | | | | | | | | | d͡ʑon平 |
| GRU-DPD | **saːn˨˩** | - | - | san˨˦ | - | ʈ͡ʂʰan˥ | - | - | d͡ʑon平 |
| GRU-DPD-ΠM | t͡sʰaːn˥ | - | - | **t͡sʰan˥** | - | ʈ͡ʂʰan˥ | - | - | t͡ɕʰan平 |
| GRU-DPD-BST | t͡sʰaːn˨˩ | - | - | **t͡sʰan˥** | - | ʈ͡ʂʰan˥ | - | - | ʂæn平 |
| GRU-DPD-ΠM-BST | t͡sʰaːn˥ | - | - | **t͡sʰan˥** | - | ʈ͡ʂʰan˥ | - | - | d͡ʑan平 |
| Trans-SUPV | | | | | | | | | ʈ͡ʂʰæn平 |
| Trans-ΠM | | | | | | | | | ʈ͡ʂʰæn平 |
| Trans-BST | | | | | | | | | ʈ͡ʂʰæn平 |
| Trans-ΠM-BST | | | | | | | | | d͡ʑan平 |
| Trans-DPD | t͡sʰaːn˥ | - | - | **t͡sʰan˥** | - | ʈ͡ʂʰan˥ | - | - | ʈ͡ʂʰæn平 |
| Trans-DPD-ΠM | t͡sʰaːn˥ | - | - | **t͡sʰan˥** | - | ʈ͡ʂʰan˥ | - | - | ʈ͡ʂʰæn平 |
| Trans-DPD-BST | saːn˥ | - | - | **t͡sʰan˥** | - | **ʈ͡ʂʰan˧˥** | - | - | ʈ͡ʂʰæn平 |
| Trans-DPD-ΠM-BST | t͡sʰaːn˥ | - | - | **t͡sʰan˥** | - | ʈ͡ʂʰan˥ | - | - | d͡ʑæn平 |

Table 29: Example outputs on WikiHan for all strategy-architecture combinations. Bold: correct protoform or reflex; '-': the dataset does not contain this reflex; blank: reflex prediction is not applicable for the strategy.

| | Romanian | French | Italian | Spanish | Portuguese | Latin |
|---|---|---|---|---|---|---|
| Reference | - | **kɔ̃tʁibye** | **kontribuire** | **kontɾiβwiɾ** | **kuŋtɹibuiɹ** | **kɔntrɪbʊɛrɛ** |
| GRU-SUPV | | | | | | kɔntrɪbʊbrɛm |
| GRU-ΠM | | | | | | kɔntrɪbutorɛm |
| GRU-BST | | | | | | kɔntriburɛ |
| GRU-ΠM-BST | | | | | | kɔntrɪbuɪrɛm |
| GRU-DPD | - | **kɔ̃tʁibye** | **kontribuire** | **kontɾiβwiɾ** | **kuŋtɹibuiɹ** | kɔntribʊrɛ |
| GRU-DPD-ΠM | - | kɔ̃tʁibyl | **kontribuire** | kontɾiβueɾ | kuŋtɹibuvɹ | kɔntribʊrɛ |
| GRU-DPD-BST | - | **kɔ̃tʁibye** | **kontribuire** | **kontɾiβwiɾ** | **kuŋtɹibuiɹ** | kɔntrɪbuɪrɛm |
| GRU-DPD-ΠM-BST | - | **kɔ̃tʁibye** | **kontribuire** | **kontɾiβwiɾ** | **kuŋtɹibuiɹ** | kɔntrɪbuɪrɛm |
| Trans-SUPV | | | | | | kɔntribɪrɛ |
| Trans-ΠM | | | | | | kɔntrɪbʊrɛ |
| Trans-BST | | | | | | kɔntribɪrɛ |
| Trans-ΠM-BST | | | | | | kɔntribʊrɛ |
| Trans-DPD | - | **kɔ̃tʁibye** | **kontribuire** | **kontɾiβwiɾ** | **kuŋtɹibuiɹ** | kɔntrɪbʊrɛ |
| Trans-DPD-ΠM | - | kɔ̃tʁibyʁ | kontriburre | kontɾiβuiɾ | **kuŋtɹibuiɹ** | kɔntrɪbʊrɛ |
| Trans-DPD-BST | - | **kɔ̃tʁibye** | **kontribuire** | **kontɾiβwiɾ** | **kuŋtɹibuiɹ** | kɔntribʊorɛ |
| Trans-DPD-ΠM-BST | - | **kɔ̃tʁibye** | **kontribuire** | kontɾiβuiɾ | **kuŋtɹibuiɹ** | kɔntribʊrɛ |
| Reference | **tʃenzor** | **sɑ̃sœʁ** | **tʃensore** | **θensoɾ** | **seɪŋsoɹ** | **kensorɛm** |
| GRU-SUPV | | | | | | kɪnsɔres |
| GRU-ΠM | | | | | | kɪnsɔrsʊm |
| GRU-BST | | | | | | kɛnsɔrɛm |
| GRU-ΠM-BST | | | | | | kɛnsɔrɛm |
| GRU-DPD | tʃensor | **sɑ̃sœʁ** | tʃɛnsore | **θensoɾ** | **seɪŋsoɹ** | kɛnsɔrɛ |
| GRU-DPD-ΠM | tʃensor | sɑ̃sɔʁ | **tʃensore** | **θensoɾ** | seɪŋsuɹ | kɪnsɔrɛm |
| GRU-DPD-BST | tʃensoɾ | **sɑ̃sœʁ** | **tʃensore** | **θensoɾ** | **seɪŋsoɹ** | kɛnsorɛm |
| GRU-DPD-ΠM-BST | tʃensoɾ | **sɑ̃sœʁ** | **tʃensore** | **θensoɾ** | seɪŋsuɹ | kɛnsorɛm |
| Trans-SUPV | | | | | | kɛnssorɛm |
| Trans-ΠM | | | | | | kɪŋkorɛm |
| Trans-BST | | | | | | kɛnsorɛm |
| Trans-ΠM-BST | | | | | | kɛnsorɛm |
| Trans-DPD | tʃensor | **sɑ̃sœʁ** | tʃensɔre | **θensoɾ** | **seɪŋsoɹ** | kɪŋsɛrorɛm |
| Trans-DPD-ΠM | tʃensor | sɑ̃sɔʁ | **tʃensore** | **θensoɾ** | seɪŋsɔɹ | kɛnsɔrɛm |
| Trans-DPD-BST | tʃensor | **sɑ̃sœʁ** | **tʃensore** | **θensoɾ** | **seɪŋsoɹ** | kɛnsorɛm |
| Trans-DPD-ΠM-BST | tʃensor | **sɑ̃sœʁ** | **tʃensore** | **θensoɾ** | **seɪŋsoɹ** | kɛnsorɛm |
| Reference | **insuflatsje** | **ɛ̃syflasjɔ̃** | **insufflatsione** | **insuflaθjon** | **iŋsuflɐsɐ̃ʊ̃** | **insʊfflatɪonɛm** |
| GRU-SUPV | | | | | | insʊflatɪonɛm |
| GRU-ΠM | | | | | | ɪnsʊpʰlatɪonɛm |
| GRU-BST | | | | | | ɪnsʊflatɪonɛm |
| GRU-ΠM-BST | | | | | | ɪnsʊfflatɪonɛm |
| GRU-DPD | **insuflatsje** | **ɛ̃syflasjɔ̃** | insufllatsione | **insuflaθjon** | iŋsuflɐtɐ̃ʊ̃ | insʊflatɪonɛm |
| GRU-DPD-ΠM | **insuflatsje** | ɛ̃syflsjɔ̃ | **insufflatsione** | **insuflaθjon** | **iŋsuflɐsɐ̃ʊ̃** | ɪnsʊpʰlatɪonɛm |
| GRU-DPD-BST | **insuflatsje** | **ɛ̃syflasjɔ̃** | insufllatsione | **insuflaθjon** | **iŋsuflɐsɐ̃ʊ̃** | ɪnsʊflatɪonɛm |
| GRU-DPD-ΠM-BST | insuflaksje | ɛ̃syflakjɔ̃ | **insufflatsione** | **insuflaθjon** | **iŋsuflɐsɐ̃ʊ̃** | insʊflatɪonɛm |
| Trans-SUPV | | | | | | insʊflatɪonɛm |
| Trans-ΠM | | | | | | insʊpʰlatɪonɛm |
| Trans-BST | | | | | | insʊlfatɪonɛm |
| Trans-ΠM-BST | | | | | | insʊflatɪonɛm |
| Trans-DPD | **insuflatsje** | **ɛ̃syflasjɔ̃** | insufllatsione | **insuflaθjon** | **iŋsuflɐsɐ̃ʊ̃** | insʊlatɪonɛm |
| Trans-DPD-ΠM | insullatsje | ɛ̃syllasjɔ̃ | insulllatsione | insullaθjon | iŋsullɐsɐ̃ʊ̃ | insʊlatɪonɛm |
| Trans-DPD-BST | **insuflatsje** | **ɛ̃syflasjɔ̃** | insufllatsione | **insuflaθjon** | **iŋsuflɐsɐ̃ʊ̃** | insʊflatɪonɛm |
| Trans-DPD-ΠM-BST | **insuflatsje** | **ɛ̃syflasjɔ̃** | insufllatsione | **insuflaθjon** | **iŋsuflɐsɐ̃ʊ̃** | insʊflatɪonɛm |
| Reference | - | - | **perdʒurio** | - | **peɾəʒuɾjʊ** | **pɛrɪurɪʊm** |
| GRU-SUPV | | | | | | pɛrdɪʊrɪʊm |
| GRU-ΠM | | | | | | pɛrdʊrerɪʊm |
| GRU-BST | | | | | | pɛrgurɪʊm |
| GRU-ΠM-BST | | | | | | pɛrdɛrɪʊm |
| GRU-DPD | - | - | periʒorio | - | peɾiʒuɾiʊ | pɛrgurɪʊm |
| GRU-DPD-ΠM | - | - | periʒurio | - | peɾiʒuɾiʊ | **pɛrɪurɪʊm** |
| GRU-DPD-BST | - | - | **perdʒurio** | - | **peɾəʒuɾjʊ** | pɛrgurɪʊm |
| GRU-DPD-ΠM-BST | - | - | pergʒurio | - | peɾəguɾjʊ | pɛrgʊrɪʊm |
| Trans-SUPV | | | | | | pɛrgʊrirɪʊm |
| Trans-ΠM | | | | | | pɛrgurɪʊm |
| Trans-BST | | | | | | pɛrgʊrɪʊm |
| Trans-ΠM-BST | | | | | | pɛrgurɪʊm |
| Trans-DPD | - | - | **perdʒurio** | - | **peɾəʒuɾjʊ** | pɛrgurɪʊm |
| Trans-DPD-ΠM | - | - | pɛrdʒʒrio | - | pɛɾəguɾjʊ | pɛrgɪʊrɪʊm |
| Trans-DPD-BST | - | - | **perdʒurio** | - | **peɾəʒuɾjʊ** | pɛrgurɪʊm |
| Trans-DPD-ΠM-BST | - | - | **perdʒurio** | - | **peɾəʒuɾjʊ** | pɛrgurɪʊm |

Table 30: Example outputs on Rom-phon for all strategy-architecture combinations. Bold: correct protoform or reflex; '-': the dataset does not contain this reflex; blank: reflex prediction is not applicable for the strategy.